\documentclass[11pt]{article}
\PassOptionsToPackage{hyperfootnotes=false}{hyperref}
\usepackage[final]{acl}
\usepackage{times}
\usepackage{graphicx}
\usepackage{booktabs}
\usepackage{array}
\usepackage{adjustbox}
\usepackage{amsmath}
\usepackage{ragged2e}
\usepackage[table,dvipsnames]{xcolor}
\usepackage{enumerate}
\usepackage{multirow}
\usepackage{tabularray}
\usepackage{siunitx}
\usepackage{colortbl}
\usepackage{tabularx}
\usepackage{supertabular} 
\usepackage{subcaption}
\usepackage{longtable}
\usepackage{enumitem}
\usepackage{xurl}
\definecolor{ailens_purple}{HTML}{b637fb}

\title{PRISM: Prompt-Refined In-Context System Modeling \\for Financial Retrieval}

\author{
  Chun Chet Ng\textsuperscript{1}\thanks{\ Both authors contributed equally to this work.}, \;
  Jia Yu Lim\textsuperscript{1}\footnotemark[1], \;
  Zhen Hao Chu\textsuperscript{1}, \;
  Yixi Zhou\textsuperscript{2}, \;
  Wei Zeng Low\textsuperscript{1} \\
  \textsuperscript{1}AI Lens, Kuala Lumpur, Malaysia \\
  \textsuperscript{2}Hong Kong Baptist University, Hong Kong \\
  \texttt{chunchet.ng@ailensgroup.com}
}

\begin{document}
\maketitle
\begin{abstract}
Deploying LLMs for financial information retrieval demands accuracy, cost efficiency, and reproducibility, yet most approaches require expensive fine-tuning. PRISM is a training-free framework that composes three deployable modules (prompt engineering, in-context learning, and a candidate refinement module) with a boundary-condition study of multi-agent coordination for financial document retrieval. On FiQA-2018, stacked reranking and $k$-reciprocal encoding raise Recall@100 from 0.7783 to 0.8326. On FinAgentBench, a Cohere-refined top-50 pool yields a higher observed end-to-end NDCG@5, below this evaluation's minimum detectable effect, while cutting per-query cost by about 35\%. Added complexity does not always pay off: $k$-reciprocal encoding stops helping once the upstream stage is already high-recall, and multi-agent coordination is not a default for fine-grained chunk ranking. Matched pairs that hold prompt, model, and exemplars fixed lose 0.0702 and 0.0681 combined NDCG@5 to their single-path controls, and the whole gap sits at the chunk stage. We release an expert-annotated FinAgentBench evaluation set, together with a label-robustness sweep over 16 alternative ground-truth constructions, restoring a reproducible evaluation path after the original evaluator went offline. Without fine-tuning, PRISM placed third on the official FinAgentBench private leaderboard (leaderboard basis, NDCG@5 0.71181), reaches NDCG@10 0.6197 on FiQA-2018, and 98\% accuracy on FinanceBench. We report per-query latency, token, and dollar accounting across roughly 40 configurations, present the quality--cost--latency Pareto frontier, and introduce FACET, a transparent SLA-specific scalarization for choosing among non-dominated configurations. Code and dataset are released at \href{https://bit.ly/prism-ailens}{\texttt{bit.ly/prism-ailens}}.
\end{abstract}

\section{Introduction}
\begin{figure*}[!ht]
    \centering
    \includegraphics[width=\linewidth, keepaspectratio]{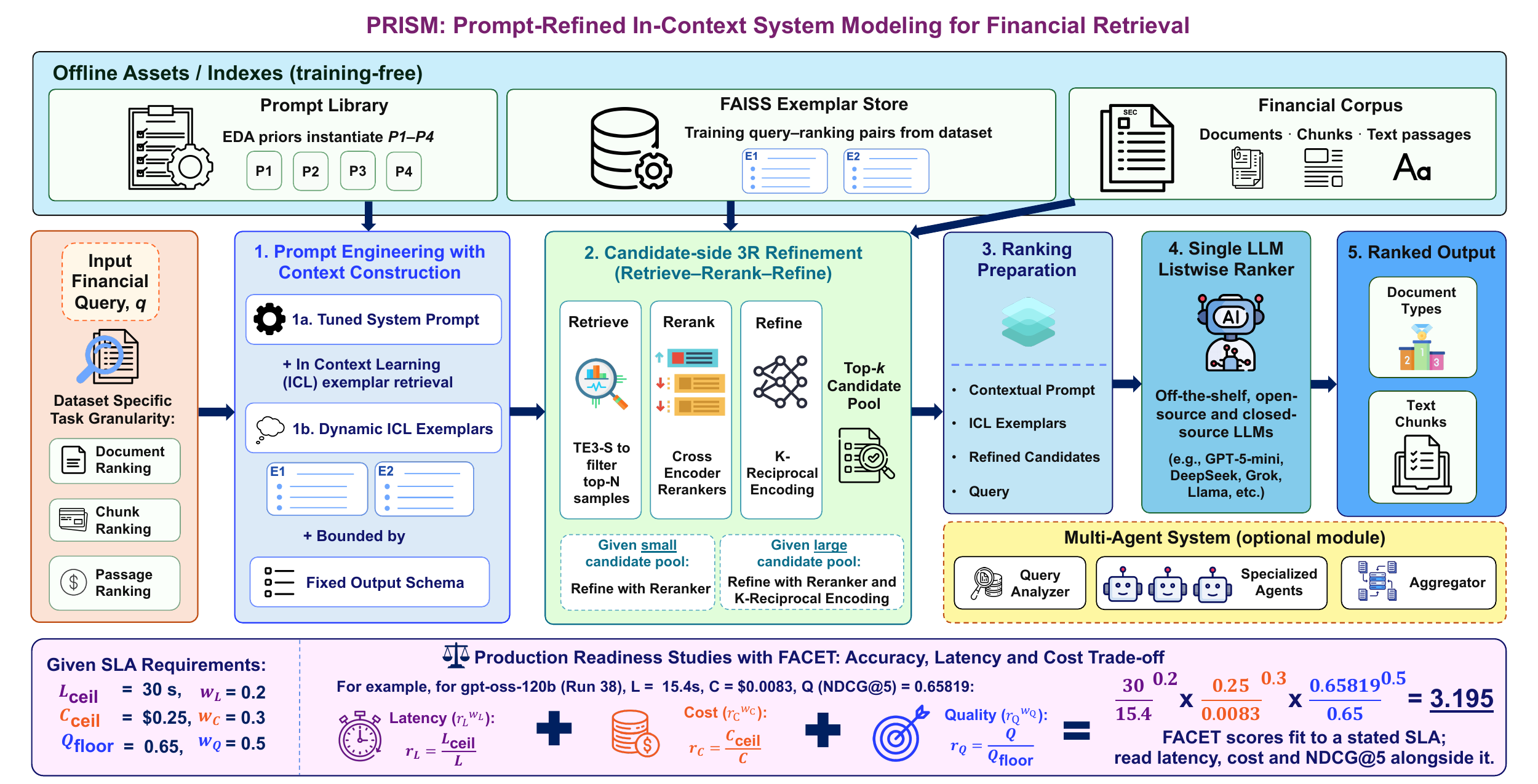}
    \caption{Overview of the PRISM framework. A query passes left to right through five stages. (1)~\textbf{Prompt engineering} encodes financial-domain priors and a relevance scaffold into the ranking instruction (1a), and \textbf{ICL exemplar retrieval} \emph{prepends} demonstrations from a separate training-split store (1b), never changing the candidate set. (2)~\textbf{Candidate refinement} queries the corpus for the current query and \emph{shrinks} the candidate set through dense retrieval, reranking, and optional $k$-reciprocal encoding. The two retrieval paths are drawn separately because they query different indexes and move context in opposite directions (Section~\ref{sec:two_retrievers}). (3)~Prompt, exemplars, refined candidates, and query are assembled and (4)~ranked in a single LLM call, giving (5)~the ranked output. \textbf{Multi-agent coordination} replaces stage~4 when enabled; it is optional and is studied here as a boundary condition, not as the recommended production path for chunk ranking (Section~\ref{sec:mas_matched}).}
    \label{fig:overview}
    \vspace{-0.5em}
\end{figure*}

Financial information retrieval (IR) systems must retrieve verifiable evidence from long documents under cost and latency constraints~\cite{zhu2023large,zhao2024optimizing}. This is difficult in production because relevance must be judged across granularities, from full filings to passages, while task-specific labels for fine-tuning are often unavailable. Large language model (LLM)-based retrieval and Retrieval-Augmented Generation (RAG) offer training-free alternatives, but their performance depends on interacting choices: prompt design, in-context examples, candidate refinement, context length, and model capacity~\cite{choi2025finder,chatgptLLMRerank}. Existing work provides limited deployment guidance on which intervention to use, when it improves ranking, and when it adds cost or complexity without benefit. We address this gap with Prompt-Refined In-Context System Modeling (PRISM), a training-free financial IR framework and deployment-oriented empirical study that evaluates prompt engineering, in-context learning (ICL), candidate refinement, and multi-agent coordination across document, chunk, and passage ranking. Figure~\ref{fig:overview} summarizes the PRISM workflow. We also re-annotate and release FinAgentBench labels to restore reproducible evaluation after the original evaluator became unavailable~\cite{finagentbench}. The main contributions of this work are:

\begin{enumerate}[itemsep=0pt,leftmargin=*]
    \item We present PRISM, a training-free financial IR framework evaluated on FinAgentBench, FiQA-2018, and FinanceBench. It combines three deployable modules, prompt engineering, in-context learning, and a candidate refinement module, with a boundary-condition study of multi-agent coordination. The refinement module stacks dense retrieval, commercial and late-interaction rerankers, and $k$-reciprocal encoding, improving FiQA-2018 Recall@100 and yielding a higher observed FinAgentBench chunk NDCG@5, below this evaluation's minimum detectable effect, while reducing per-query cost by about 35\%.
    \item We release a systematically re-annotated FinAgentBench evaluation set with annotator-agreement analysis (Kendall's $\tau_b$, RBO@5, Cohen's $\kappa$) and a label-robustness sweep that rebuilds the ground truth 16 different ways and re-scores all 40 configurations against each, restoring a reproducible evaluation path after the original evaluator became unavailable. Every score we report is labeled with its evaluation basis, re-annotated or leaderboard.
    \item We attribute the module-level findings with controlled comparisons rather than argument: architecture-matched pairs holding prompt, model, and exemplars fixed, query-level paired bootstrap intervals, and minimum detectable effects. Configurations whose intervals overlap are not described as ordered.
    \item We report the first per-query latency, token, and dollar accounting on this benchmark, present the quality--cost--latency Pareto frontier, and summarize the deployment choice in a practitioner decision table; FACET follows as a transparent SLA-specific scalarization over the non-dominated set, with raw axes alongside.
\end{enumerate}

\section{Related Work}
\noindent\textbf{LLMs for Ranking.}
LLMs are strong zero-shot rankers, typically formulated pointwise, pairwise, or listwise~\cite{LLMZeroShotRankers}. Listwise methods such as RankGPT balance effectiveness and efficiency but are limited by input length and sensitive to document order~\cite{chatgptLLMRerank}; TourRank adds multi-stage grouping and a tournament strategy for large candidate sets~\cite{tourrankLLM}, and pairwise ranking compares candidates directly at linear complexity~\cite{LLMTextRerankPairwiseRanking}. We treat these as related formulations and instead target deployment-component ablations for long, dense financial texts~\cite{zhao2024docmath}.

\noindent\textbf{In-Context Learning (ICL).}
ICL adapts LLMs through in-prompt examples without gradient updates~\cite{ICLBayesianInference,WhatMakesICLWork}. Embedding-based ICL retrieval for financial IR is underexplored when latency-sensitive applications must allocate a fixed context budget between examples and long-document evidence.

\noindent\textbf{Retrieval and Reranking.}
Neural retrieval pipelines pair a first-stage retriever with a reranker; dense, sparse, late-interaction, cross-encoder, and commercial rerankers are strong alternatives to lexical baselines~\cite{splade2021,colbertv2,nogueira2019passage,bgem3}. Prior financial IR work has largely benchmarked against BM25-style baselines, leaving the deployment value of stronger modules unclear in production settings.

\noindent\textbf{Rank Fusion and $k$-Reciprocal Encoding.}
Reciprocal Rank Fusion combines ranked lists through reciprocal-rank weights~\cite{reciprocalrankfusion}, while $k$-reciprocal encoding, first proposed for person re-identification, refines candidates through mutual nearest-neighbor structure~\cite{zhong2017kreciprocal,qin2011hello}, both without supervision. Further background is in Appendix~\ref{sec:additional_related_work}.

\section{Datasets}
We use three financial datasets covering complementary ranking granularities (document, chunk, passage), task types (ranking, QA), and reasoning demands (lexical matching, numerical analysis).

\noindent\textbf{FiQA-2018}~\cite{fiqa2018} is a financial opinion QA dataset from BEIR~\cite{beir}, requiring passage reranking over short financial texts. \textbf{FinanceBench}~\cite{islam2023financebench} tests numerical-reasoning QA over SEC filings rather than ranking. \textbf{FinAgentBench}~\cite{finagentbench} comprises 2023--2024 SEC filings from EDGAR and supports two tasks: ranking five document types (10-K, 10-Q, 8-K, DEF 14A, Earnings Transcripts) by relevance, and ranking text chunks within documents. It has 4,986 document-ranking and 18,855 chunk-ranking training samples, and a closed-source 200-example validation set covering both document and chunk ranking. 

\vspace{-0.2em}
\subsection{FinAgentBench Re-Annotation}
\label{sec:annotation}
\vspace{-0.2em}
The original evaluation dataset became unavailable after the competition~\cite{kaggle}, leaving leaderboard-era results no longer reproducible by the community. We address this gap by releasing an open validation set in which three domain experts label the document-ranking samples and four label the chunk-ranking samples.

\noindent\textbf{Inter-Annotator Agreement.}
For each task, we measure pairwise agreement using Kendall's $\tau_b$~\cite{kendall1938tau}, RBO@5~\cite{webber2010rbo}, and Cohen's $\kappa$~\cite{cohen1960kappa}. Agreement is substantially higher for document ranking than for chunk ranking (Appendix~\ref{sec:annotation_details}). Table~\ref{tab:annotation_summary} summarizes the selected pair per task; full pairwise agreement is reported in Appendix Table~\ref{tab:iaa_pairs}.

\noindent\textbf{Ground Truth and Release.}
We select the most concordant annotator pair per task to reduce label noise from low-agreement annotators, merge their rankings by Borda count, and validate the resulting labels by re-scoring all configurations in Appendix Table~\ref{tab:ablation} against the new ground truth. The re-scored results preserve the broad performance pattern of the original private leaderboard (Table~\ref{tab:annotation_summary}); per-configuration deltas for every eligible run are in Appendix Table~\ref{tab:annotation_validation}.

\begin{table}[t]
\footnotesize
\centering
\caption{Summary of the released FinAgentBench evaluation labels. Agreement is for the selected annotator pair; validation compares all eligible configurations with original private-leaderboard scores.}
\label{tab:annotation_summary}
\resizebox{\columnwidth}{!}{%
\begin{tabular}{lccccc}
\toprule
\textbf{Task} & \textbf{Experts} & \textbf{Pair} & $\boldsymbol{\tau_b}$ & \textbf{RBO@5} & $\boldsymbol{\kappa}$ \\
\midrule
Document ranking & 3 & 1, 2 & 0.754 & 0.957 & 0.827 \\
Chunk ranking & 4 & 1, 4 & 0.582 & 0.730 & 0.649 \\
\midrule
\multicolumn{5}{r}{\textbf{Mean absolute NDCG@5 difference from original scores}} & \textbf{0.0702} \\
\multicolumn{5}{r}{\textbf{Ordering vs.\ leaderboard,} $\boldsymbol{\rho}$ \textbf{/} $\boldsymbol{\tau_b}$ ($n=34$)} & \textbf{0.7800 / 0.6613} \\
\bottomrule
\end{tabular}}
\end{table}

\noindent\textbf{Evaluation Bases.}
\label{sec:bases}
Two bases appear in this paper and no column mixes them. The \emph{leaderboard basis} is the score returned by the original evaluator on its held-out private subset, and exists only for configurations submitted before it went offline (Appendix Tables~\ref{tab:kaggle_lb} and~\ref{tab:annotation_validation}). The \emph{re-annotated basis} re-scores released prediction files against the released expert labels over the 200 document-ranking and 199 chunk-ranking scoring queries. The two differ in both label source and query sample, so their absolute values are not expected to coincide: across the 34 configurations carrying both, the mean level shift is 0.070 while the ordering is broadly preserved (Spearman $\rho=0.7800$, $p=5.4\times10^{-8}$; Kendall $\tau_b=0.6613$, $p=3.8\times10^{-8}$). We use that correlation to validate broad ordering only, and do not read close configurations as ordered. Unless labeled otherwise, every FinAgentBench number here is on the re-annotated basis.

\noindent\textbf{Dataset Diagnostics.}
We conduct EDA on FinAgentBench, FiQA-2018, and FinanceBench to characterize document-type heterogeneity, label sparsity, and chunk-length variability across tasks. These diagnostics motivate type-aware prompting, selective ICL, and candidate refinement; full diagnostics are provided in Appendix~\ref{sec:eda_finagentbench}--\ref{sec:financebench_analysis}.

\section{PRISM}
\vspace{-0.2em}
\subsection{Prompt Engineering}
\vspace{-0.2em}
Prompt engineering encodes financial-domain priors and output constraints into the ranking instruction. \(P_1\) supplies domain keyword hints and statistical cues only; \(P_2\)--\(P_4\) add a structured relevance scaffold combining chain-of-thought decomposition, ReAct-style alternation, and tree-of-thought multi-expert simulation, and \(P_4\) is \(P_3\) with streamlined instructions and terminology unified across stages, so what separates them is output discipline rather than length. Matched on model and exemplars (GPT-5-mini, no ICL), the step that carries the effect is adding the scaffold: \(P_1\) is the weakest of the four on both evaluation bases (Runs~2, 9, 12, 13; re-annotated 0.75758, 0.76648, 0.76287, 0.77668). Within the scaffolded set \(P_4\) is the strongest on the re-annotated basis and within 0.0012 of the strongest on the leaderboard basis, so we standardize on it as the safest default. This is a reliability claim rather than a per-prompt gap, consistent with the best prompt form differing by benchmark. Chunk scores also depend on the document stage, since the chunk ranker only sees chunks from the documents the first stage selected, so prompt effects are read at the pipeline level. Further analyses are in Appendix~\ref{sec:prompt_version} and~\ref{sec:prompt_template}.

\vspace{-0.2em}
\subsection{In-Context Learning Sample Retrieval}
\vspace{-0.2em}
The ICL module retrieves semantically similar training examples from a separate FAISS~\cite{faiss} exemplar store and prepends them as few-shot demonstrations. Exemplars are drawn only from the training split, so no test label or test candidate is ever used as a demonstration and an exemplar cannot enter the candidate set for the query it conditions. Implementation details and shot-count sensitivity are in Appendix~\ref{sec:icl_retrieval_details}.

\noindent\textbf{Two retrievers, opposite roles.}
\label{sec:two_retrievers}
The ICL retriever and the candidate-refinement retriever of Section~\ref{sec:refinement} share TE3-S as a backbone but are not redundant: the first queries the training-split exemplar store and prepends demonstrations, leaving the candidate set untouched, while the second queries the corpus for the current query and changes the candidate set. One adds context for task adaptation, the other removes it for pool construction, and the effects follow that difference: ICL helps document ranking but not chunk ranking, where exemplars consume context that candidate evidence needs, whereas refinement improves the chunk path precisely by shrinking what the LLM sees, so the two are complementary.

\vspace{-0.2em}
\subsection{Candidate Refinement Module}
\label{sec:refinement}
\vspace{-0.2em}
The third module is a model-agnostic \emph{retrieve--rerank--refine} (3R) stage before the LLM ranker. It compresses noisy candidate pools into smaller, denser sets via an embedding retriever, a reranker, and optional $k$-reciprocal encoding, reducing context length while preserving relevance.

\noindent\textbf{Embedding Retrieval.}
The first stage retrieves candidates with text-embedding-3-small (TE3-S), selected empirically in Appendix~\ref{sec:embedding_selection}. On FiQA-2018, every refinement method receives the same TE3-S top-1000 pool from the full 57.6K-passage corpus. On FinAgentBench, each query already has a few hundred associated chunks, which are refined only for chunk ranking and passed to the LLM as a top-$k$ pool.

\noindent\textbf{Reranking.}
The second stage rescores candidates with cross-encoder, late-interaction, or commercial rerankers. We compare Cohere rerank-v4.0-fast (hereafter Cohere)~\cite{cohere2026rerank}, ColBERTv2~\cite{colbertv2}, BGE-M3 ColBERT, BGE reranker V2 M3~\cite{bgem3}, and Jina reranker v3~\cite{jinav3} under the same candidate budget in Appendix~\ref{sec:filtering_fiqa}.

\noindent\textbf{K-Reciprocal Encoding.}
The final stage applies training-free $k$-reciprocal encoding~\cite{zhong2017kreciprocal}. For L2-normalized embeddings, $d(q,c_i)=2-2\hat e_q^{\top}\hat e_{c_i}$ and the reciprocal set is $R(q,k_1)=\{c_i\in N(q,k_1)\mid q\in N(c_i,k_1)\}$. We encode these mutual neighborhoods and blend their Jaccard distance $d_J$ with direct distance,
\[
d^{*}(q,c_i)=\lambda d(q,c_i)+(1-\lambda)d_J(q,c_i),
\]
then rerank candidates by ascending $d^*$. This stage exploits complementary neighborhood structure in large, sparse pools; our experiments (Appendix~\ref{sec:filtering_appendix}) show it is not beneficial on small, homogeneous chunk sets; reranker-only pipelines are better defaults for compact financial retrieval under these observed conditions.

\vspace{-0.2em}
\subsection{Multi-Agent System (MAS) Modeling}
\vspace{-0.2em}
The fourth module studies MAS as a boundary-condition analysis rather than a default production recommendation. It belongs in the paper because the negative result is only interpretable against the rest of the framework: without matched non-agentic controls there is nothing to attribute the loss to. We test when role-specialized agents help coarse document routing and when agent handoffs cost quality at chunk granularity; architectures \(A_1\)--\(A_4\) are detailed in Appendix~\ref{sec:mas_architecture_details}. The controlled comparison in Section~\ref{sec:mas_matched} is run under \(A_4\) with \(P_4\), so the resulting statement is scoped to the tested \(A_4\) workflow and is not a claim about multi-agent design in general.

\section{Experiments}
We evaluate four closed-source models (GPT-4o-mini, GPT-4.1, GPT-5-mini, and GPT-5) and four open-source models: DeepSeek-V4-Flash~\cite{deepseekai2026deepseekv4}, gpt-oss-120b~\cite{gptoss120b}, grok-4-20-reasoning~\cite{grok4}, and Llama-4-Maverick-17B-128E-Instruct-FP8~\cite{llama4}, using OpenAI's text-embedding-3-small (TE3-S) and text-embedding-3-large (TE3-L)~\cite{openai2024embedding} for ICL retrieval. Our evaluation spans three datasets chosen to cover complementary task characteristics: FinAgentBench (document and chunk ranking with SEC filings), FiQA-2018 (passage reranking for financial QA), and FinanceBench (factual QA requiring numerical reasoning). PRISM's training-free design allows adaptation to other LLM providers and deployment settings with minimal pipeline changes. Implementation details are in Appendix~\ref{sec:implementation_details}.

\vspace{-0.2em}
\subsection{Evaluation Metrics}
\vspace{-0.2em}
On FinAgentBench we use combined document-and-chunk NDCG@5 as the primary metric, matching the competition objective, and also report MAP@5 and MRR@5. On FiQA-2018 we report NDCG@10 and Recall@100 under the BEIR protocol~\cite{beir}, measuring top-ranked quality and reranking coverage respectively. On FinanceBench we report factual QA accuracy under oracle context.

\vspace{-0.2em}
\subsection{Quantitative Results}
\subsubsection{FinAgentBench}
\label{sec:finagent_results}
\vspace{-0.2em}
All FinAgentBench scores in this section are on the re-annotated basis (Section~\ref{sec:bases}) unless labeled otherwise; the full 40-configuration grid is in Appendix Table~\ref{tab:ablation} and the document/chunk splits in Appendix Table~\ref{tab:ablation_fine}. Table~\ref{tab:headline} reports the configurations the paper's claims rest on, with query-level paired bootstrap confidence intervals (10{,}000 resamples, document and chunk resampled independently because their query sets are disjoint).

\noindent\textbf{Leaderboard basis.}
On the original private leaderboard, PRISM placed 3rd of all teams with the competition-selected Run~19 submission (leaderboard basis, 0.71181 NDCG@5), and repeated submissions of the same configuration reached 0.71818 on the private subset (Appendix~\ref{sec:quanti_finagent_appendix}). This made PRISM the highest-ranked non-fine-tuned submission, trailing the top fine-tuned system by 0.007 NDCG@5 on our best observed private-subset run (0.71818 vs.\ 0.72497, both leaderboard basis). Runs~37--40 could not have been submitted: those models were not available to us during the competition. They are post-competition runs, evaluated only on the re-annotated set and never on the original private leaderboard, so they carry no original score to compare against, and leaderboard and re-annotated numbers are never compared directly.

\noindent\textbf{Re-annotated basis.}
Run~19 scores 0.77468 here; several configurations score higher under the new labels, and we track it only because it was the submitted one. Under the Run~19 recipe, grok-4-20-reasoning (Run~39) reaches the best combined NDCG@5 of 0.83721 and exceeds the strongest GPT-5-family configuration (Run~18, 0.78421) by 0.0529 (95\% CI [0.0331, 0.0727]; chunk Wilcoxon $p=9.7\times10^{-10}$), holding under 15 of 16 label constructions. DeepSeek-V4-Flash (Run~37) reaches 0.80157, but its difference from Run~18 is positive with a 95\% CI including zero ($+0.0175$, $[-0.0084, 0.0424]$) and holds under only 9 of 16 constructions, so we place it in the same quality band as the strongest GPT-5-family run rather than above it, at roughly a fifth of the cost of the GPT-5 runs with measured cost (\$0.0284 against \$0.1448). Against Run~18 both models trade a little document accuracy for a large chunk gain ($-0.023$ and $+0.129$ for grok-4, $-0.015$ and $+0.050$ for DeepSeek; Appendix Table~\ref{tab:ablation_fine}), consistent with a long-context discrimination advantage over the roughly 95.8k-token chunk prompt; we treat it as a descriptive diagnostic rather than causal evidence.

\begin{table}[t]
\footnotesize
\centering
\caption{Configurations the claims rest on, using bootstrap re-scores on the re-annotated basis with query-level 95\% CIs (10{,}000 resamples) over the 200 document-ranking and 199 chunk-ranking scoring queries; the point estimates elsewhere use the 200-query convention, so values differ in the fourth decimal. Run~19 is the competition submission, not the best row. Full grid in Appendix Table~\ref{tab:ablation}.}
\label{tab:headline}
\resizebox{\columnwidth}{!}{%
\begin{tabular}{clcc}
\toprule
\textbf{ID} & \textbf{Configuration} & \textbf{Combined NDCG@5} & \textbf{95\% CI} \\
\midrule
39 & grok-4-20-reasoning & 0.8370 & [0.8190, 0.8541] \\
37 & DeepSeek-V4-Flash & 0.8016 & [0.7787, 0.8226] \\
18 & strongest GPT-5 family & 0.7841 & [0.7637, 0.8044] \\
19 & GPT-5, competition submission & 0.7738 & [0.7528, 0.7944] \\
\midrule
\multicolumn{4}{l}{\emph{Architecture-matched pairs} ($P_4$, GPT-5-mini, ICL held fixed)} \\
16 vs.\ 33 & single-path vs.\ $A_4$ agentic & $+0.0702$ & [$+0.0474$, $+0.0937$] \\
17 vs.\ 34 & single-path vs.\ $A_4$ agentic & $+0.0681$ & [$+0.0436$, $+0.0920$] \\
\bottomrule
\end{tabular}}
% \vspace{-1em}
\end{table}

\noindent\textbf{Chunk ranking} is the axis that separates configurations. Document ranking is effectively saturated, reaching NDCG@5 up to 0.95573 (Run~40), so the candidate-refinement module pays off on the chunk task alone. Reranking alone is insufficient, as the best reranker-only setup with Cohere reaches 0.5464 chunk NDCG@5, below PRISM's LLM ranker (Run~19, 0.61278). Feeding the reranked pool to the LLM yields a higher observed chunk NDCG@5 while reducing cost. Hence, reranking is most useful as a compression stage for the LLM rather than as a standalone ranker. We quantify this end to end in Section~\ref{sec:filtering_results}.

\vspace{-0.2em}
\subsubsection{FiQA-2018}
\vspace{-0.2em}
On FiQA-2018, the strongest evidence comes from controlled stage-wise comparisons under the same candidate pool. TE3-S reaches NDCG@10 0.5499 and Recall@100 0.7783; stacking reranking and $k$-reciprocal encoding raises Recall@100 to 0.8326 (Section~\ref{sec:filtering_results}). Against published BEIR baselines~\cite{beir}, GPT-5 PRISM reaches NDCG@10 0.6197 (Appendix Table~\ref{tab:fiqa_results}), a 78.59\% gain over BM25+Cross-Encoder (CE), and all four open-source models with $P_4$ outperform BM25+CE (full results in Appendix~\ref{sec:quanti_fiqa}). We note that the retrieval stages alone are already competitive at the top of the ranking on this dataset: Cohere reranking reaches NDCG@10 0.6258 with no LLM in the loop (Appendix~\ref{sec:filtering_fiqa}), so on FiQA the LLM ranker contributes candidate coverage rather than top-10 precision. Full results are in Appendix Table~\ref{tab:full_fiqa_results}.

\vspace{-0.2em}
\subsubsection{FinanceBench}
\vspace{-0.2em}
As a secondary QA generalization check, we evaluate PRISM under oracle context (relevant passages provided). PRISM reaches 98\% accuracy (147/150) with \(P_1\), surpassing the GPT-5 oracle baseline of 94\% (141/150). This suggests that prompt engineering transfers to numerical reasoning tasks, though it does not constitute a full retrieval evaluation, since passages are given, not retrieved. Full oracle, zero-shot, and open-source results are in Appendix~\ref{sec:quanti_financebench}; full FinAgentBench run-level ablations are in Appendix Table~\ref{tab:ablation}.

% \vspace{-0.2em}

% \vspace{-0.2em}
\subsection{Multi-Agent Coordination: A Matched Comparison}
\label{sec:mas_matched}
\vspace{-0.2em}
Comparing an agentic run against Run~19 confounds architecture with prompt and model. We therefore use two pairs matched on prompt, model, and exemplars, leaving architecture as the only difference: Runs~16 and~33 are both \(P_4\)/GPT-5-mini with ICL-5 on document and none on chunk, and Runs~17 and~34 are both \(P_4\)/GPT-5-mini with ICL-5 on both stages. Run~32 is excluded as prompt-mismatched. With those held fixed, the single-path pipeline exceeds the \(A_4\) agentic one by 0.0702 and 0.0681 combined NDCG@5 (Table~\ref{tab:headline}). Both differences are significant under a paired query-level bootstrap at 10{,}000 resamples and under a Wilcoxon signed-rank test on chunk scores ($p=9.4\times10^{-9}$ and $2.8\times10^{-7}$), and the two estimates agree to 0.002. Both differences stay positive under all 16 alternative label constructions and the smaller never falls below $+0.035$, so the result is not an artifact of how the ground truth was built. Table~\ref{tab:mas_stage} splits each pair by stage: the whole gap is at chunk granularity, about 0.14 NDCG@5; documents favor the agentic run.

\begin{table}[t]
\footnotesize
\centering
\caption{Where the coordination penalty sits. Matched on prompt, model, and ICL; re-annotated basis. Document differences are not distinguishable from zero, and the whole gap is at the chunk stage.}
\label{tab:mas_stage}
\resizebox{\columnwidth}{!}{%
\begin{tabular}{lcc}
\toprule
\textbf{Matched pair} & \textbf{Document NDCG@5} & \textbf{Chunk NDCG@5} \\
\midrule
Run~16 vs.\ Run~33 & 0.94594 vs.\ 0.95359 ($p=0.32$) & 0.59099 vs.\ 0.44364 \\
Run~17 vs.\ Run~34 & 0.94950 vs.\ 0.95259 ($p=0.20$) & 0.55340 vs.\ 0.41476 \\
\bottomrule
\end{tabular}}
% \vspace{-0.5em}
\end{table}

\vspace{-0.2em}
\subsection{Candidate Refinement Results}
\label{sec:filtering_results}
\vspace{-0.2em}
The embedding backbone that defines the candidate pool and the reranker that rescores it are selected in Appendix~\ref{sec:filtering_appendix}; here we report the end-to-end effect once the refined pool reaches the LLM ranker, which is what directly determines ranking quality, cost, and latency in deployment. Stacking-order ablations are in Appendix~\ref{sec:filtering_finagent}.

\vspace{-0.2em}
We now check whether the retrieval-stage gains survive when the refined pool reaches the LLM ranker. On FiQA-2018 (Appendix Table~\ref{tab:fiqa_refine}) every pipeline sees the identical TE3-S top-1000 pool per query at fixed cutoffs. Stacking the retrieval stages raises Recall@100 from 0.7783 to 0.8326 and NDCG@10 from 0.5499 to 0.6333. Passing the refined pool to GPT-5 adds coverage (Recall@100 0.8115 to 0.8307) but not top-of-ranking precision: NDCG@10 is 0.6243 against 0.6258 for Cohere alone and 0.6333 for Cohere$\to$$k$-reciprocal. On a corpus like FiQA the LLM ranker therefore earns its cost only when coverage matters more than the order of the first ten results. The trade-off inverts on FinAgentBench chunk ranking (Appendix Table~\ref{tab:finagent_e2e}, \(P_4\)/GPT-5, ICL-5). Refining to the top~50 with TE3-S$\to$Cohere gives the best chunk NDCG@5 (0.63408 against a no-refinement baseline of 0.61278) while cutting per-query cost by about 35\% (\$0.181 against \$0.278). Latency does not improve (146.3\,s against 136.7\,s), because the added stages are not offset by the shorter LLM context. We rest this result on the cost reduction rather than the quality change: the numbers come from a single run per configuration, and $+0.021$ chunk NDCG@5 is below the 0.057 chunk-level minimum detectable effect of this evaluation (Appendix~\ref{sec:adjacent_significance}).

\section{Feasibility \& Reproducibility Analysis}

\subsection{Choosing a Configuration}
\label{sec:practitioner}
\begin{figure}[t]
\centering
\includegraphics[width=\columnwidth]{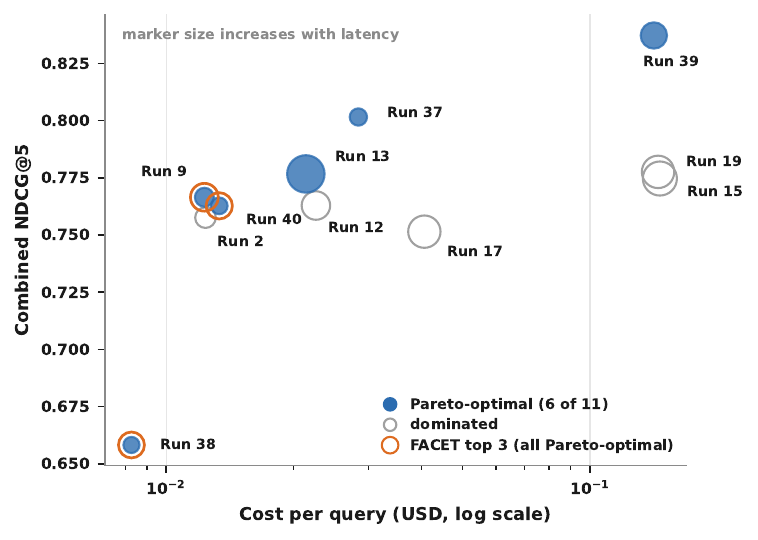}
\caption{Quality--cost--latency dominance over the configurations with measured cost and latency (combined NDCG@5, re-annotated basis, against USD per query on a log axis; marker size grows with latency). Five of 11 are non-dominated in this projection and six on all three axes; the six filled markers are that three-axis set, and the three highest FACET scores are all Pareto-optimal.}
\label{fig:pareto}
\end{figure}

Quality, latency, and cost do not reduce to one number without a stated preference, so we present the dominance structure first. Figure~\ref{fig:pareto} plots the frontier over the 11 configurations with measured cost and latency: 6 are non-dominated, and that is the set to choose from (Appendix~\ref{sec:dominance_appendix}). Table~\ref{tab:practitioner} names one operating point per constraint; the rows are distinct operating points rather than an ordered recommendation. One reading needs stating explicitly: FACET measures fit to a stated SLA rather than quality, so the cheapest configuration scores highest, and a practitioner with a hard constraint on one axis should read that raw column.

\begin{table}[t]
\footnotesize
\centering
\caption{Practitioner decision table. Distinct operating points under different constraints, not a ranking. Run IDs match Appendix Table~\ref{tab:ablation}; models are grok-4-20-reasoning, DeepSeek-V4-Flash, Llama-4-Maverick-17B-128E-Instruct-FP8, gpt-oss-120b, and GPT-5. NDCG@5 is combined, re-annotated basis; latency (s) and cost are per query. Run~19 uses combined document--chunk per-query accounting; post-annotation Runs~37--40 retain their originally reported latency and cost measurements.}
\label{tab:practitioner}
\setlength{\tabcolsep}{3.5pt}
\resizebox{\columnwidth}{!}{%
\begin{tabular}{@{}llrrrr@{}}
\toprule
\textbf{Constraint} & \textbf{Choose} & \textbf{Lat.} & \textbf{\$/q} & \textbf{NDCG} & \textbf{FACET} \\
\midrule
Quality first & 39, grok-4 & 48.9 & 0.1415 & 0.8372 & 1.221 \\
Quality per \$ & 37, DeepSeek & 18.5 & 0.0284 & 0.8016 & 2.349 \\
Latency, $Q{>}0.75$ & 40, Llama-4 & 17.0 & 0.0133 & 0.7627 & 2.923 \\
Lowest \$/query & 38, gpt-oss & 15.4 & 0.0083 & 0.6582 & 3.195 \\
Submitted run & 19, GPT-5 & 84.6 & 0.1462 & 0.7747 & 1.042 \\
\bottomrule
\end{tabular}}
\end{table}

\noindent\textbf{Module-level rules.}
Use the relevance scaffold (\(P_2\)--\(P_4\)), not the keyword-hint baseline (\(P_1\)). Apply ICL to document ranking, not by default to chunk ranking. Treat reranking as an LLM-context compression stage rather than a standalone ranker, and add $k$-reciprocal encoding only where upstream coverage is the bottleneck (Appendix~\ref{sec:krecip_rule}). Treat multi-agent coordination as a boundary condition, not the production default for chunk ranking, where matched pairs cost about 0.07 NDCG@5 against a single LLM call.

\section{Conclusion}
\vspace{-0.4em}
PRISM composes prompt engineering, in-context learning, and candidate refinement without training, and studies multi-agent coordination as a boundary condition. Matched on prompt, model, and exemplars, the single-path pipeline beats the tested \(A_4\) agentic workflow by 0.0702 and 0.0681 combined NDCG@5, entirely at the chunk stage, so the simpler path is the default for fine-grained ranking, and ICL pays off only at document granularity. The scaffold, not length, matters. Refinement improves FiQA-2018 Recall@100 and yields a higher observed FinAgentBench chunk NDCG@5, below this evaluation's minimum detectable effect, at about 35\% lower cost. The released evaluation set and robustness sweep restore a reproducible path, and the cost accounting and frontier make the trade-off readable. In Table~\ref{tab:practitioner}, the anchors are Run~39 (grok-4-20-reasoning, 0.8372 combined NDCG@5 at USD 0.1415 and 48.9\,s) and Run~37 (DeepSeek-V4-Flash, 0.8016 at USD 0.0284 and 18.5\,s). Their 95\% intervals overlap (Table~\ref{tab:headline}), so we do not order them. PRISM is the top non-fine-tuned leaderboard entry, within 0.007 NDCG@5 of the best fine-tuned one, and the cheaper anchor, at one fifth the cost and 40\% of the latency.

\clearpage
\section{Limitations}
\label{sec:limitations}
This section highlights the key limitations of PRISM; Appendix~\ref{sec:full_limitations} discusses them in full.

\noindent\textbf{Benchmark Reproducibility.} The FinAgentBench evaluation environment became unavailable following the competition's conclusion, preventing new submissions and independent replication of private-test scores. The organizers did not disclose ground-truth labels for the public or private test sets. As a result, the private-test competition scores (Appendix Table~\ref{tab:kaggle_lb}) cannot be reproduced using the original evaluator. Our ablation experiments on the validation split remain reproducible with the code we release upon acceptance, and our expert re-annotated evaluation set (Section~\ref{sec:annotation}) provides a community-accessible substitute for the offline evaluator, enabling independent re-evaluation of all 40 configurations in Appendix Table~\ref{tab:ablation} against a fixed public label set.

\noindent\textbf{Scope of the MAS Result.}
The coordination penalty is established for the tested \(A_4\) workflow under \(P_4\), with two matched pairs (Section~\ref{sec:mas_matched}), not for the full \(A_1\)--\(A_4\) grid. The remaining architectures were not run under \(P_4\), so we do not claim that every multi-agent design loses at chunk granularity. The earlier prompt-architecture asymmetry, in which agentic runs used \(P_1\) while the strongest non-agentic run used \(P_4\), no longer carries the conclusion: Runs~33 and~34 are \(P_4\) under \(A_4\) and matched to non-agentic controls on prompt, model, and exemplars, so the comparison isolates architecture. Run~32 remains prompt-mismatched (\(P_4\) on document, \(P_1\) on chunk) and is excluded from the matched analysis; its low score (0.54229) is reported in Appendix Table~\ref{tab:ablation} but is not evidence about coordination.

\noindent\textbf{Domain Specificity.}
PRISM's financial specialization lies in the SEC-filing prompt priors with their keyword-frequency ratios, the financial exemplar pool, and document-type routing. These are inputs rather than architecture, so porting the pipeline to legal or medical retrieval means re-deriving priors and re-indexing exemplars with no weights to retrain. We have not tested that and do not claim it; every module-level finding in this paper is scoped to financial IR, and cross-vertical evaluation is the next step. The evidence is financial and we do not claim cross-domain generalization, but the retrieval-design questions addressed also arise beyond finance.

\noindent\textbf{Chunk-Level Resolution and API Dependence.}
The chunk task has a minimum detectable effect of 0.057 at 80\% power, so differences smaller than that are not resolvable by this evaluation and we do not interpret them. Separately, training-free operation is not infrastructure independence: exact reproduction of API-served results depends on endpoint availability and versioning, and commercial endpoints are not stationary. We report exact endpoint identifiers and per-run call timestamps, and make the evaluation rather than the endpoints reproducible by releasing the expert annotations, source code, prompts, and configuration details.

\section*{Ethical Considerations}
All datasets used in this work are publicly available and were obtained through official distribution channels. FinAgentBench~\cite{finagentbench} was released as part of an open Kaggle competition and contains excerpts of SEC filings, which are public regulatory disclosures rather than personal data. FiQA-2018 is distributed through the BEIR benchmark~\cite{beir}, and FinanceBench~\cite{islam2023financebench} is publicly released for non-commercial research. No proprietary data were used. We release our expert annotations, that is, the labels we produced over the public FinAgentBench validation split, rather than a re-release of the underlying dataset, so the release is consistent with the original distribution terms. The annotations, together with the scoring, bootstrap, and label-robustness scripts, are released under CC BY 4.0; the underlying SEC excerpts remain under their original public-domain status and are not redistributed by us. The annotators are domain experts who consented to the use of their judgments for this release; the labeled material contains no personal or client data. Since financial retrieval may influence high-stakes analysis, PRISM is intended as an evidence-ranking aid rather than an autonomous decision system, with human review, access control, audit logging, and checks that answers remain grounded in retrieved evidence.

\section*{Acknowledgement}
We thank the Cradle Fund and Microsoft for Startups for their funding support, which made this research possible. We also thank the DocLab, DocSuite, and MLSuite teams at AI Lens for their contributions that collectively shaped the PRISM pipeline presented in this work.
\bibliography{ref}

\clearpage
\appendix
\section{Appendix}
This appendix provides implementation details, extended results, and supporting analyses in the same section order as the main paper.

\subsection{Additional Related Work}
\label{sec:additional_related_work}
\noindent\textbf{Prompt Engineering.}
Recent studies have shown that retrieval performance is highly sensitive to how an LLM is instructed~\cite{zhang2025scientific}. The initial breakthrough in few-shot learning demonstrated that LLMs can perform tasks using in-context exemplars without gradient updates~\cite{fewshotlearners}. However, traditional few-shot approaches struggled with complex reasoning tasks. This limitation was addressed by Chain-of-Thought (CoT) prompting, which enhances reasoning by guiding LLMs through intermediate natural language reasoning steps~\cite{chainofthought}. The Reasoning and Acting (ReAct) framework improves reliability by allowing LLMs to plan and execute reasoning steps iteratively, reducing hallucination~\cite{reactLLM}. Building on this, the Tree-of-Thoughts (ToT) framework enables LLMs to explore and evaluate multiple reasoning paths~\cite{treeofthought}.

\noindent\textbf{Agentic Information Retrieval.}
Agentic information retrieval represents a new paradigm where LLM agents dynamically manage information access through iterative cycles of observation, reasoning, and action, distinguishing it from traditional IR architectures~\cite{zhang2025agenticinformationretrieval}. It is commonly implemented using multi-agent systems (MAS), which coordinate specialized agents for reflection, planning, and tool use to achieve collective intelligence and outperform single-agent approaches~\cite{singh2026agenticretrievalaugmentedgenerationsurvey}. LLM agents have also been applied to ranking tasks, demonstrating that retrieval and ranking agents can effectively coexist~\cite{xu-etal-2025-iagent}. Despite its potential, agentic IR faces challenges in coordinating complex MAS interactions, reducing the high inference cost of LLMs, and evaluating dynamic agentic behaviors in financial retrieval.

\noindent\textbf{Neural Retrieval and Reranking.}
First-stage retrievers fall into three main families. Dense bi-encoders embed queries and passages into a shared space for approximate nearest-neighbor search~\cite{karpukhin2020dense}, exemplified by commercial models such as OpenAI text-embedding-3 and the dense mode of BGE-M3~\cite{bgem3}. Learned sparse models such as SPLADE~\cite{splade2021} and the sparse mode of BGE-M3 retain the exact-term-matching and inverted-index efficiency of lexical methods while adding learned term expansion. Late-interaction models such as ColBERTv2~\cite{colbertv2} store token-level embeddings and match them at fine granularity. A reranker then rescores retrieved candidates: cross-encoders jointly encode the query and each passage for higher accuracy at greater computational cost~\cite{nogueira2019passage}, while commercial rerankers such as Cohere and Jina reranker v3~\cite{jinav3} offer strong off-the-shelf alternatives.

\noindent\textbf{Rank Fusion and Reciprocal Encoding.}
RRF merges several ranked lists by summing reciprocal-rank weights and can outperform individual systems and learned fusion methods~\cite{reciprocalrankfusion}. $k$-reciprocal encoding, introduced for person re-identification~\cite{zhong2017kreciprocal} and rooted in earlier $k$-reciprocal nearest-neighbor retrieval~\cite{qin2011hello}, instead refines a candidate set by exploiting mutual nearest-neighbor relationships. Each item is encoded from its $k$-reciprocal neighbors and rescored under a Jaccard distance, promoting candidates that are reciprocally close to the query rather than merely close in one direction.

\subsection{FinAgentBench Re-Annotation Details}
\label{sec:annotation_details}
\noindent\textbf{Human-only annotation.}
We deliberately rely on human experts alone and exclude LLMs from annotation because our retrieval pipeline itself uses cross-encoders and LLMs; using LLM-generated labels as ground truth would bias the evaluation toward the systems under test and invite a circularity objection.

\noindent\textbf{Agreement levels.}
The selected chunk pair reaches $\kappa = 0.649$, $\tau_b = 0.582$, RBO@5 $= 0.730$ (Table~\ref{tab:annotation_summary}), and we agree the remaining chunk pairs are much lower. The structural reason is that chunk ranking asks for a top-5 ordering out of the few hundred topically related chunks each query carries, so two experts who both locate relevant material can still disagree on which five and in what order. Document ranking has a small fixed candidate set and correspondingly high agreement ($\tau_b = 0.754$, $\kappa = 0.827$). The principal MAS and grok-versus-GPT-5 conclusions are configuration-ordering claims, and the sweep in Appendix~\ref{sec:label_robustness_appendix} shows that those conclusions remain stable across the tested ground-truth constructions.

For ground-truth construction, we apply the same objective rule to both FinAgentBench tasks: select the annotator pair with the highest mean agreement across Kendall's $\tau_b$, RBO@5, and Cohen's $\kappa$, then merge the two reference rankings by Borda count with ties broken by document or chunk ID. This selects annotators~1 and~2 for document ranking and annotators~1 and~4 for chunk ranking. To validate the reconstructed labels, we re-score the configurations in Table~\ref{tab:ablation} against the new ground truth. Table~\ref{tab:annotation_validation} shows that the reconstructed labels preserve the broad performance pattern of the original private-leaderboard scores.
\begin{table}[!ht]
\footnotesize
\centering
\caption{Validation of the re-annotated FinAgentBench labels: new NDCG@5 vs.\ the original private-leaderboard score. Runs~33--34 and~37--40 were not evaluated on the original test set and are therefore excluded. Run IDs match Appendix Table~\ref{tab:ablation}.}
\label{tab:annotation_validation}
\resizebox{\columnwidth}{!}{
\begin{tabular}{@{}cccccc@{}}
\toprule
\multirow{2}{*}{\textbf{ID}} & \multicolumn{3}{c}{\textbf{New NDCG@5}} & \multirow{2}{*}{\textbf{Orig.\ NDCG@5}} & \multirow{2}{*}{\textbf{Difference}} \\
\cmidrule(lr){2-4}
 & \textbf{Doc.} & \textbf{Chunk} & \textbf{Total} & & \\
\midrule
\multicolumn{6}{c}{\cellcolor{ailens_purple!35}\textbf{Non-Agentic Workflow}} \\
\midrule
1 & 0.93857 & 0.62470 & 0.78164 & 0.66666 & 0.11498 \\
2 & 0.93288 & 0.58229 & 0.75759 & 0.66628 & 0.09131 \\
3 & 0.93288 & 0.62470 & 0.77879 & 0.66328 & 0.11551 \\
4 & 0.93857 & 0.58229 & 0.76043 & 0.66966 & 0.09077 \\
5 & 0.93857 & 0.60145 & 0.77001 & 0.67353 & 0.09648 \\
6 & 0.93288 & 0.59386 & 0.76337 & 0.67021 & 0.09316 \\
7 & 0.93152 & 0.62470 & 0.77811 & 0.66088 & 0.11723 \\
8 & 0.93288 & 0.60145 & 0.76717 & 0.67014 & 0.09703 \\
9 & 0.93152 & 0.60145 & 0.76649 & 0.66774 & 0.09875 \\
10 & 0.94469 & 0.62470 & 0.78470 & 0.66673 & 0.11797 \\
11 & 0.94469 & 0.58229 & 0.76349 & 0.66973 & 0.09376 \\
12 & 0.94800 & 0.57774 & 0.76287 & 0.69019 & 0.07268 \\
13 & 0.94955 & 0.60382 & 0.77669 & 0.68895 & 0.08774 \\
14 & 0.93935 & 0.61278 & 0.77607 & 0.70575 & 0.07032 \\
15 & 0.95227 & 0.60290 & 0.77759 & 0.70977 & 0.06782 \\
16 & 0.94594 & 0.59099 & 0.76847 & 0.69325 & 0.07522 \\
17 & 0.94950 & 0.55340 & 0.75145 & 0.68252 & 0.06893 \\
18 & 0.94891 & 0.61952 & 0.78422 & 0.71446 & 0.06976 \\
19 & 0.93658 & 0.61278 & 0.77468 & 0.71818 & 0.05650 \\
\midrule
\multicolumn{6}{c}{\cellcolor{ailens_purple!35}\textbf{Agentic Workflow}} \\
\midrule
20 & 0.92290 & 0.34595 & 0.63443 & 0.57276 & 0.06167 \\
21 & 0.92794 & 0.35419 & 0.64107 & 0.58234 & 0.05873 \\
22 & 0.56826 & 0.47162 & 0.51994 & 0.51518 & 0.00476 \\
23 & 0.94440 & 0.27232 & 0.60836 & 0.55256 & 0.05580 \\
24 & 0.56614 & 0.47448 & 0.52031 & 0.53400 & -0.01369 \\
25 & 0.94522 & 0.50183 & 0.72353 & 0.66291 & 0.06062 \\
26 & 0.69962 & 0.28918 & 0.49440 & 0.49782 & -0.00342 \\
27 & 0.94464 & 0.46326 & 0.70395 & 0.63654 & 0.06741 \\
28 & 0.93987 & 0.49024 & 0.71506 & 0.62775 & 0.08731 \\
29 & 0.94144 & 0.43765 & 0.68955 & 0.62771 & 0.06184 \\
30 & 0.94710 & 0.41686 & 0.68198 & 0.62294 & 0.05904 \\
31 & 0.94685 & 0.36115 & 0.65400 & 0.59906 & 0.05494 \\
32 & 0.95272 & 0.13185 & 0.54229 & 0.55757 & -0.01529 \\
\midrule
\multicolumn{6}{c}{\cellcolor{ailens_purple!35}\textbf{Hybrid Workflow}} \\
\midrule
35 & 0.92929 & 0.61278 & 0.77104 & 0.70685 & 0.06419 \\
36 & 0.95036 & 0.61278 & 0.78157 & 0.69439 & 0.08718 \\
\midrule
\multicolumn{5}{c}{\textbf{Average NDCG@5 Difference}} & \textbf{0.07020} \\
\bottomrule
\end{tabular}}
\end{table}

\begin{table}[ht]
\footnotesize
\centering
\caption{Pairwise inter-annotator agreement among domain experts on FinAgentBench.}
\label{tab:iaa_pairs}
\resizebox{\columnwidth}{!}{%
\begin{tabular}{llccc}
\toprule
\multicolumn{2}{c}{\textbf{Annotator pair}} & \textbf{Kendall $\tau_b$} & \textbf{RBO@5} & \textbf{Cohen's $\kappa$} \\
\midrule
\multicolumn{5}{c}{\cellcolor{ailens_purple!35}\textbf{Document ranking (3 experts)}} \\
\midrule
\textbf{Annot.\ 1} & \textbf{Annot.\ 2} & \textbf{0.7540} & \textbf{0.9567} & \textbf{0.8270} \\
Annot.\ 1 & Annot.\ 3 & 0.6520 & 0.9347 & 0.7335 \\
Annot.\ 3 & Annot.\ 2 & 0.7420 & 0.9541 & 0.8095 \\
\midrule
\multicolumn{5}{c}{\cellcolor{ailens_purple!35}\textbf{Chunk ranking (4 experts)}} \\
\midrule
\textbf{Annot.\ 1} & \textbf{Annot.\ 4} & \textbf{0.5822} & \textbf{0.7296} & \textbf{0.6485} \\
Annot.\ 1 & Annot.\ 2 & 0.2408 & 0.4926 & 0.2603 \\
Annot.\ 2 & Annot.\ 4 & 0.1015 & 0.5141 & 0.1142 \\
Annot.\ 3 & Annot.\ 1 & 0.1364 & 0.4886 & 0.1459 \\
Annot.\ 3 & Annot.\ 2 & 0.1721 & 0.4678 & 0.1943 \\
Annot.\ 3 & Annot.\ 4 & 0.2174 & 0.5276 & 0.2317 \\
\bottomrule
\end{tabular}%
}
\end{table}

\subsection{Label Robustness of the Re-Annotated Set}
\label{sec:label_robustness_appendix}
Section~\ref{sec:annotation} builds the released ground truth from the most concordant annotator pair per task. This appendix reports what happens to the paper's conclusions when that choice is varied. We rebuild the ground truth 16 different ways using only the annotators of Table~\ref{tab:iaa_pairs} and re-score all 40 configurations against each. The constructions are: Borda over all annotators and majority vote on top-5 membership (2); the remaining document pairs and chunk pairs, that is, all $\binom{3}{2}$ and $\binom{4}{2}$ pairs other than the two the pair rule selected (7); and every annotator alone, 3 for document and 4 for chunk (7).

\begin{figure*}[ht]
\centering
\includegraphics[width=\linewidth]{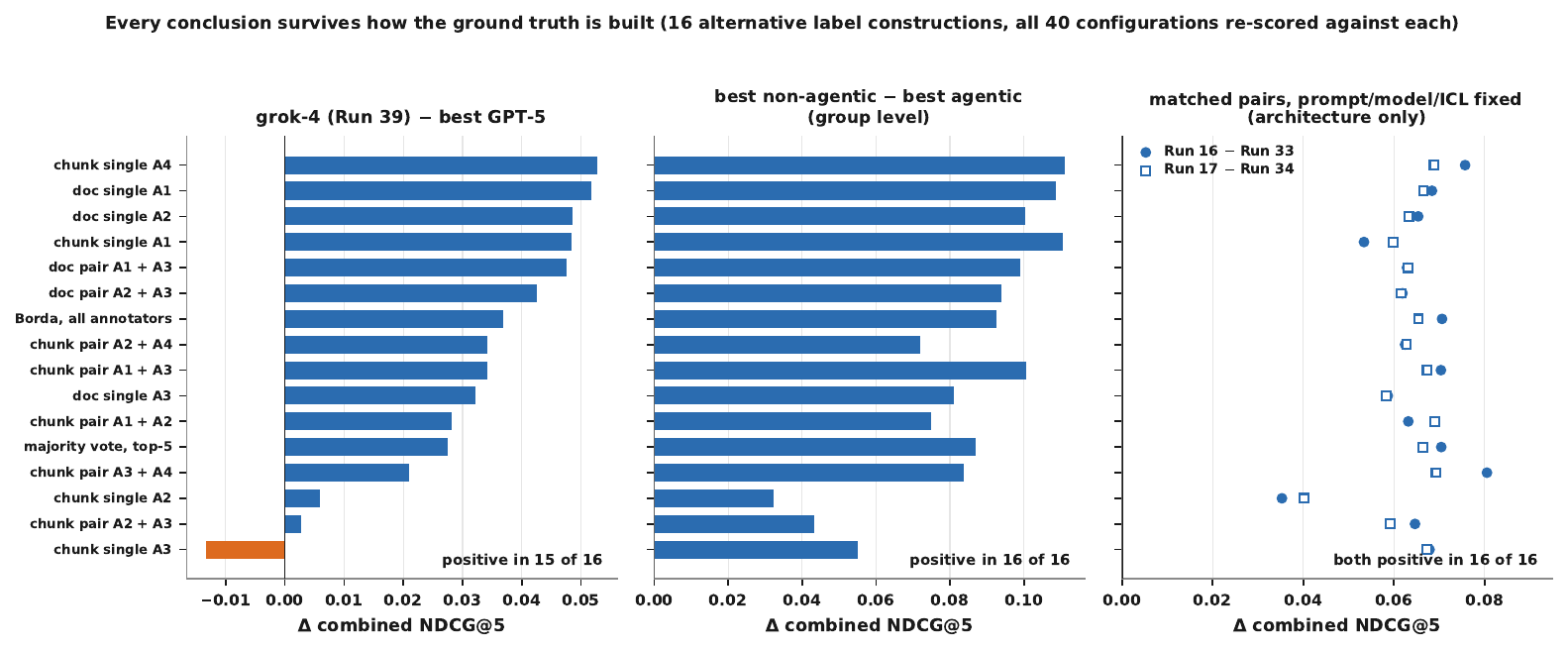}
\caption{Conclusions under the 16 alternative label constructions. Left: Run~39 (grok-4) minus the best GPT-5 configuration, positive in 15 of 16. Middle: best non-agentic minus best agentic, positive in 16 of 16. Right: the two architecture-matched pairs with prompt, model, and ICL held fixed, positive in 16 of 16.}
\label{fig:label_robustness}
\end{figure*}

Table~\ref{tab:label_robustness_full} gives the per-construction outcome and Figure~\ref{fig:label_robustness} plots it. Ordering against the reported labels stays between Spearman 0.812 and 0.998 with median 0.972. Run~39 is top-ranked under 15 of 16 constructions, grok-4-20-reasoning outranks every GPT-5 configuration under 15 of 16, and the best non-agentic configuration exceeds the best agentic one under 16 of 16. The two architecture-matched differences of Section~\ref{sec:mas_matched} are positive under 16 of 16, and the smaller of the two never falls below $+0.035$. The single exception on the grok-4 and top-run claims comes from a ground truth built on one individual rater rather than on a pair. DeepSeek-V4-Flash exceeds the strongest GPT-5-family configuration under 9 of the 16 constructions, which is why that comparison is stated without a superiority claim in Section~\ref{sec:finagent_results}.

A leave-one-annotator-out check rebuilds the ground truth from every subset that drops one rater. Both principal findings hold in 6 of 6 subsets with ordering correlation at least 0.951, and Run~39 remains top-ranked in all of them; DeepSeek again does not hold throughout.

The candidate-refinement robustness statement is limited to FiQA, where it is measured with public qrels and therefore does not depend on FinAgentBench annotator selection; we do not use this analysis to make a FinAgentBench candidate-refinement label-robustness claim.

\begin{table*}[ht]
\footnotesize
\centering
\caption{Per-construction label robustness. Each row rebuilds the FinAgentBench ground truth a different way and re-scores all 40 configurations against it. $\Delta_{16,33}$ and $\Delta_{17,34}$ are the architecture-matched differences of Section~\ref{sec:mas_matched}; both stay positive throughout. Re-annotated basis.}
\label{tab:label_robustness_full}
\begin{adjustbox}{width=\textwidth}
\begin{tabular}{lcccccc}
\toprule
\textbf{Ground-truth construction} & \textbf{Spearman vs.\ reported} & \textbf{Top run} & \textbf{Run~39} & \textbf{Run~37} & $\boldsymbol{\Delta_{16,33}}$ & $\boldsymbol{\Delta_{17,34}}$ \\
\midrule
Borda over all annotators & 0.969 & 39 & 0.8281 & 0.7914 & 0.0706 & 0.0654 \\
Majority vote on top-5 membership & 0.966 & 39 & 0.8067 & 0.7719 & 0.0704 & 0.0664 \\
Document pair (A1, A3) & 0.989 & 39 & 0.8373 & 0.8008 & 0.0629 & 0.0631 \\
Document pair (A2, A3) & 0.983 & 39 & 0.8363 & 0.8009 & 0.0619 & 0.0616 \\
Chunk pair (A1, A2) & 0.964 & 39 & 0.7600 & 0.7413 & 0.0632 & 0.0689 \\
Chunk pair (A1, A3) & 0.974 & 39 & 0.7805 & 0.7399 & 0.0704 & 0.0673 \\
Chunk pair (A2, A3) & 0.929 & 39 & 0.6764 & 0.6593 & 0.0646 & 0.0592 \\
Chunk pair (A2, A4) & 0.933 & 39 & 0.7552 & 0.7375 & 0.0625 & 0.0628 \\
Chunk pair (A3, A4) & 0.975 & 39 & 0.7662 & 0.7378 & 0.0805 & 0.0693 \\
Document annotator A1 alone & 0.998 & 39 & 0.8339 & 0.7967 & 0.0684 & 0.0665 \\
Document annotator A2 alone & 0.992 & 39 & 0.8345 & 0.7968 & 0.0653 & 0.0632 \\
Document annotator A3 alone & 0.943 & 39 & 0.8268 & 0.7889 & 0.0586 & 0.0582 \\
Chunk annotator A1 alone & 0.991 & 39 & 0.8198 & 0.7831 & 0.0534 & 0.0598 \\
Chunk annotator A2 alone & 0.812 & 39 & 0.6442 & 0.6299 & 0.0353 & 0.0401 \\
Chunk annotator A3 alone & 0.906 & 15 & 0.6601 & 0.6355 & 0.0679 & 0.0672 \\
Chunk annotator A4 alone & 0.986 & 39 & 0.8481 & 0.8152 & 0.0757 & 0.0689 \\
\bottomrule
\end{tabular}
\end{adjustbox}
\end{table*}

\noindent\textbf{Label Robustness.}
Because chunk agreement is much lower than document agreement, the conclusions must not depend on which annotators the pair rule selected. We therefore rebuilt the ground truth 16 different ways from the annotators in Appendix Table~\ref{tab:iaa_pairs} and re-scored all 40 configurations against each: Borda over all annotators and majority vote on top-5 membership (2); the remaining 2 document and 5 chunk pairs (7); and each annotator alone (7). The matched-pair agentic penalty stays positive in 16/16 constructions and 6/6 leave-one-annotator-out subsets, MAS stays below the best non-agentic run in 16/16, and grok-4 outranks every GPT-5 configuration in 15/16, with ordering correlation against the reported labels between 0.812 and 0.998 (median 0.972). The one claim that does not survive is DeepSeek-V4-Flash exceeding the strongest GPT-5-family configuration, which holds under only 9 of 16; we state that comparison without a superiority claim throughout. Per-construction scores are in Appendix~\ref{sec:label_robustness_appendix}.

\subsection{EDA on FinAgentBench}
\label{sec:eda_finagentbench}
\begin{table}[htbp]
    \centering
    \footnotesize
    \caption{Distribution of FinAgentBench document types across ranked positions in the document-ranking split.}
    \label{tab:document_ranking}
    \resizebox{\columnwidth}{!}{%
        \begin{tabular}{l ccccc}
        \toprule
        \textbf{Doc. Type} & \textbf{Rank 1} & \textbf{Rank 2} & \textbf{Rank 3} & \textbf{Rank 4} & \textbf{Rank 5} \\
        \midrule
        10-K      & \textbf{2402} & 1546 & 507  & 207  & 324  \\
        Earnings  & \textbf{1360} & 726  & 948  & 893  & 1059 \\
        10-Q      & 554  & \textbf{1558} & 1427 & 866  & 581  \\
        8-K       & 97   & 407  & 1091 & \textbf{1808} & 1583 \\
        DEF 14A   & 573  & 749  & 1013 & 1212 & \textbf{1439} \\
        \bottomrule
        \end{tabular}%
    }
\end{table}

\begin{table*}[htbp]
    \footnotesize
    \centering
    \caption{Top 6 keywords appearing in queries where the respective document type was retrieved at Rank 1 in FinAgentBench. The ratio indicates the frequency of the keyword relative to the entire corpus.}
    \label{tab:top_keywords_desc}
    \resizebox{\linewidth}{!}{%
        \begin{tabular}{l r | l r | l r | l r | l r}
        \toprule
        \multicolumn{2}{c|}{\textbf{DEF 14A}} & \multicolumn{2}{c|}{\textbf{10-K}} & \multicolumn{2}{c|}{\textbf{10-Q}} & \multicolumn{2}{c|}{\textbf{8-K}} & \multicolumn{2}{c}{\textbf{Earnings}} \\
        \textbf{Keyword} & \textbf{Ratio} & \textbf{Keyword} & \textbf{Ratio} & \textbf{Keyword} & \textbf{Ratio} & \textbf{Keyword} & \textbf{Ratio} & \textbf{Keyword} & \textbf{Ratio} \\
        \midrule
        
        dependency & \scriptsize\colorbox{ailens_purple!50}{88.40\%} & 
        evolved & \scriptsize\colorbox{ailens_purple!40}{40.98\%} & 
        quarter & \scriptsize\colorbox{ailens_purple!30}{20.00\%} & 
        compensation & \scriptsize\colorbox{ailens_purple!50}{95.45\%} & 
        asked & \scriptsize\colorbox{ailens_purple!50}{87.68\%} \\
        
        concentration & \scriptsize\colorbox{ailens_purple!50}{88.24\%} & 
        recurring & \scriptsize\colorbox{ailens_purple!40}{40.57\%} & 
        recurring & \scriptsize\colorbox{ailens_purple!20}{6.29\%} & 
        burn & \scriptsize\colorbox{ailens_purple!30}{41.74\%} & 
        questions & \scriptsize\colorbox{ailens_purple!50}{86.56\%} \\
        
        exist & \scriptsize\colorbox{ailens_purple!40}{87.73\%} & 
        ratio & \scriptsize\colorbox{ailens_purple!30}{39.40\%} & 
        evolved & \scriptsize\colorbox{ailens_purple!20}{6.12\%} & 
        award & \scriptsize\colorbox{ailens_purple!30}{41.74\%} & 
        metrics & \scriptsize\colorbox{ailens_purple!40}{78.26\%} \\
        
        risks & \scriptsize\colorbox{ailens_purple!30}{80.51\%} & 
        time & \scriptsize\colorbox{ailens_purple!30}{39.22\%} & 
        time & \scriptsize\colorbox{ailens_purple!20}{5.99\%} & 
        manage & \scriptsize\colorbox{ailens_purple!30}{41.54\%} & 
        customer & \scriptsize\colorbox{ailens_purple!30}{74.62\%} \\
        
        market & \scriptsize\colorbox{ailens_purple!20}{60.77\%} & 
        reporting & \scriptsize\colorbox{ailens_purple!20}{38.19\%} & 
        ratio & \scriptsize\colorbox{ailens_purple!10}{5.97\%} & 
        availability & \scriptsize\colorbox{ailens_purple!30}{40.60\%} & 
        guidance & \scriptsize\colorbox{ailens_purple!20}{52.97\%} \\
        
        share & \scriptsize\colorbox{ailens_purple!10}{57.12\%} & 
        period & \scriptsize\colorbox{ailens_purple!10}{36.01\%} & 
        revenue & \scriptsize\colorbox{ailens_purple!10}{4.02\%} & 
        share & \scriptsize\colorbox{ailens_purple!20}{27.21\%} & 
        offered & \scriptsize\colorbox{ailens_purple!20}{52.80\%} \\
        \bottomrule
        \end{tabular}%
    }
\end{table*}

\noindent\textbf{Document Ranking Data Analysis.}
We first conduct a systematic evaluation across all document types in the document-ranking split of FinAgentBench. Table~\ref{tab:document_ranking} summarizes the distribution of document types across ranking positions. Documents such as 10-K and 8-K appear more frequently near the top and bottom of the rankings, respectively, suggesting that information density and relevance vary across document types. Table~\ref{tab:top_keywords_desc} shows the six most frequent keywords found in Rank~1 documents for each type, revealing distinct lexical signatures across filing categories. These differences in vocabulary concentration across document types highlight the need for adaptive retrieval strategies, which inspired PRISM. Further details are provided in Appendix~\ref{sec:keyword_analysis}.

\begin{figure*}[ht]
    \centering
    \includegraphics[width=\linewidth]{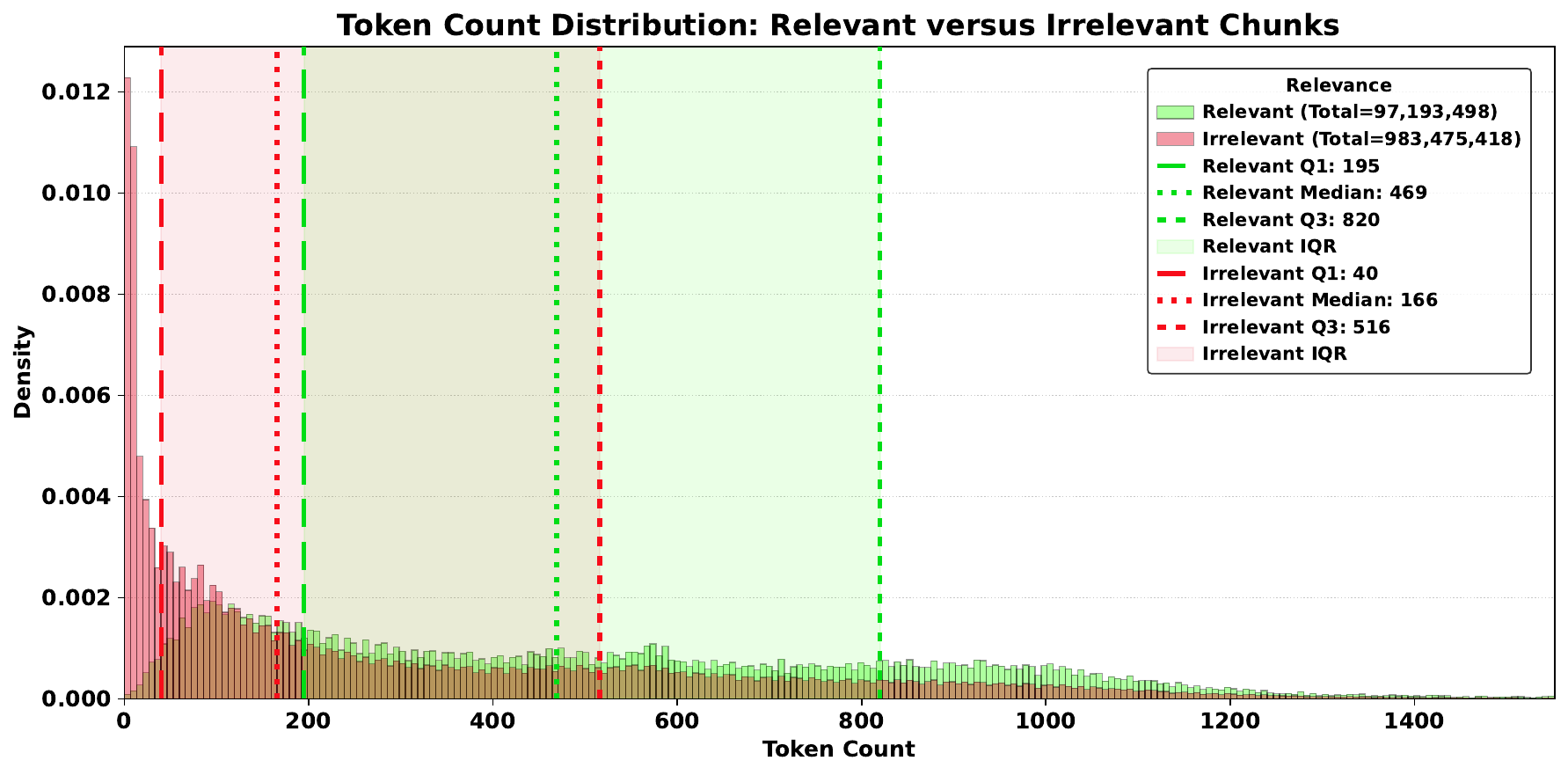}
    \caption{Token count distribution for relevant and irrelevant chunks in the FinAgentBench chunk-ranking split.}
    \label{fig:token_count_comparison}
\end{figure*}

\noindent\textbf{Chunk Ranking Data Analysis.}
We conducted a frequency analysis to better understand chunk-level characteristics in the dataset. As shown in Figure~\ref{fig:token_count_comparison}, relevant chunks are generally more information dense than irrelevant ones. The wider interquartile range (IQR) and higher maximum values, along with the long-tail distribution, suggest that adaptive retrieval methods are needed to handle substantial variability in chunk length. The relatively small proportion of relevant chunks highlights the fine-grained and selective nature of financial retrieval. Our EDA shows that relevant answers are concentrated in specific document types and in a minority of information-dense chunks, requiring models to reason over and manage large variations in chunk length.

\subsubsection{Document Ranking Keyword Analysis}
\label{sec:keyword_analysis}
Table~\ref{tab:top_keywords_desc} reveals distinct keyword concentration patterns across SEC filing types. DEF~14A filings are dominated by high-ratio keywords such as \textit{dependency} (88.40\%) and \textit{concentration} (88.24\%), while 8-K filings focus more on compensation-related terms (95.45\%). 10-Q filings, by contrast, contain notably lower keyword ratios, reflecting greater variation in language use. The spread between these extremes indicates that certain document types carry more distinctive language patterns than others, which can be exploited to guide retrieval, and these patterns informed the design of our domain-specific prompt priors and agent routing logic.

\subsubsection{Chunk Ranking Data Analysis with Word Count}
\begin{figure*}[h]
    \centering
    \includegraphics[width=\linewidth]{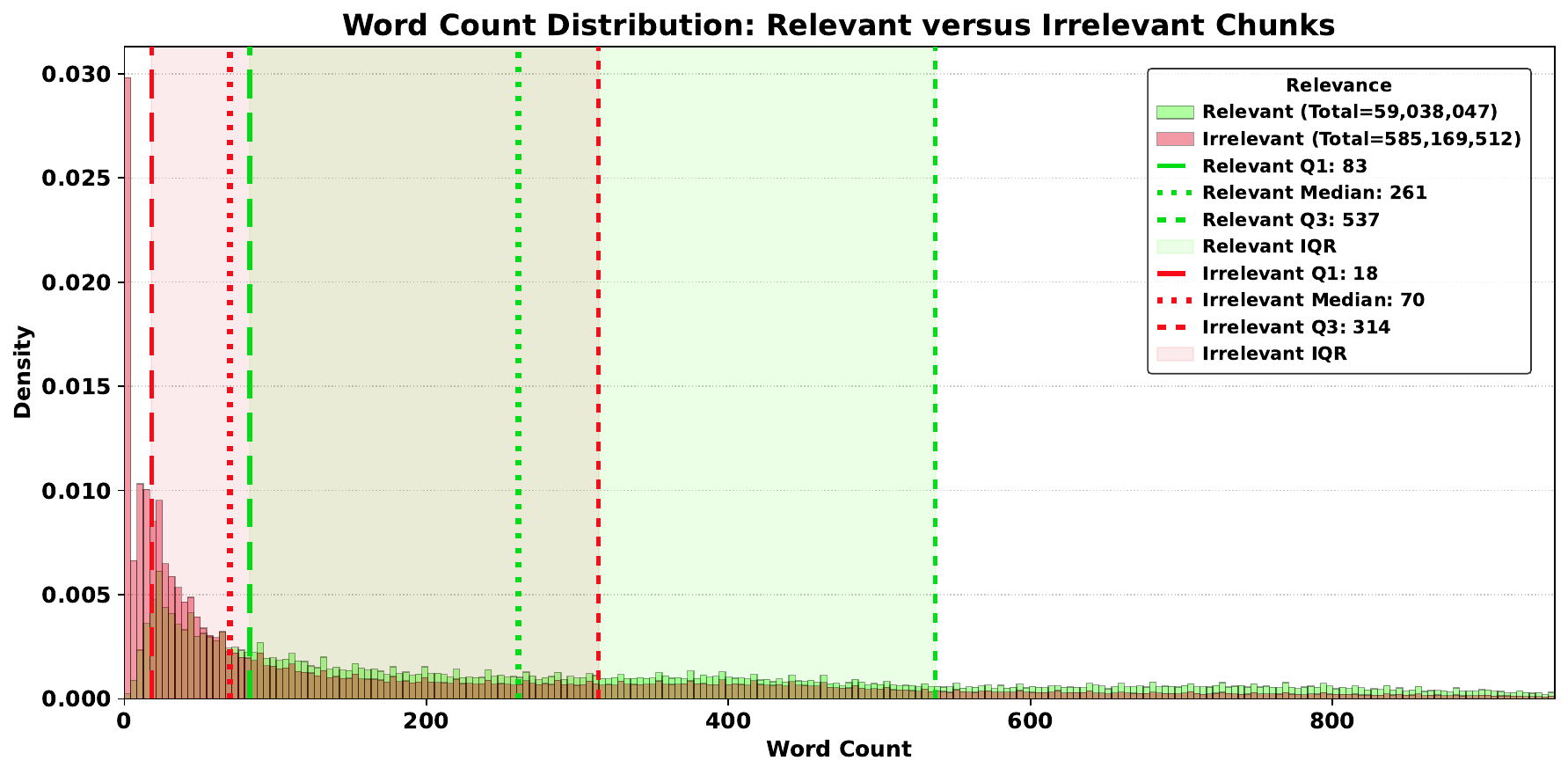}
    \caption{Word count distribution for relevant and irrelevant chunks in the FinAgentBench chunk-ranking split.}
    \label{fig:word_count_comparison}
\end{figure*}
We repeated the frequency analysis on chunk word counts as a tokenizer-independent check. As shown in Figure~\ref{fig:word_count_comparison}, the distributions mirror the token-count results in Figure~\ref{fig:token_count_comparison}, confirming that the length variability reported above is a property of the text rather than an artifact of the tokenizer.

\subsection{EDA on FiQA-2018}
FiQA-2018 contains 57,638 documents and 648 queries with 1,706 relevance judgments. With only 2.96\% of the corpus labeled as relevant and a sparsity of 99.99\%, FiQA-2018 exhibits the extreme label sparsity characteristic of real-world retrieval tasks. Most queries have fewer than three relevant documents, with average query and document lengths of 11 and 133 tokens, respectively. This characteristic places strong emphasis on semantic understanding over lexical overlap, disadvantaging keyword-based models such as BM25 when queries and documents share limited surface-form vocabulary. Dense retrieval models are better suited to this task because they capture latent semantic relevance beyond exact term matching.

\begin{figure*}[ht]
    \centering
    \includegraphics[width=0.8\linewidth, keepaspectratio]{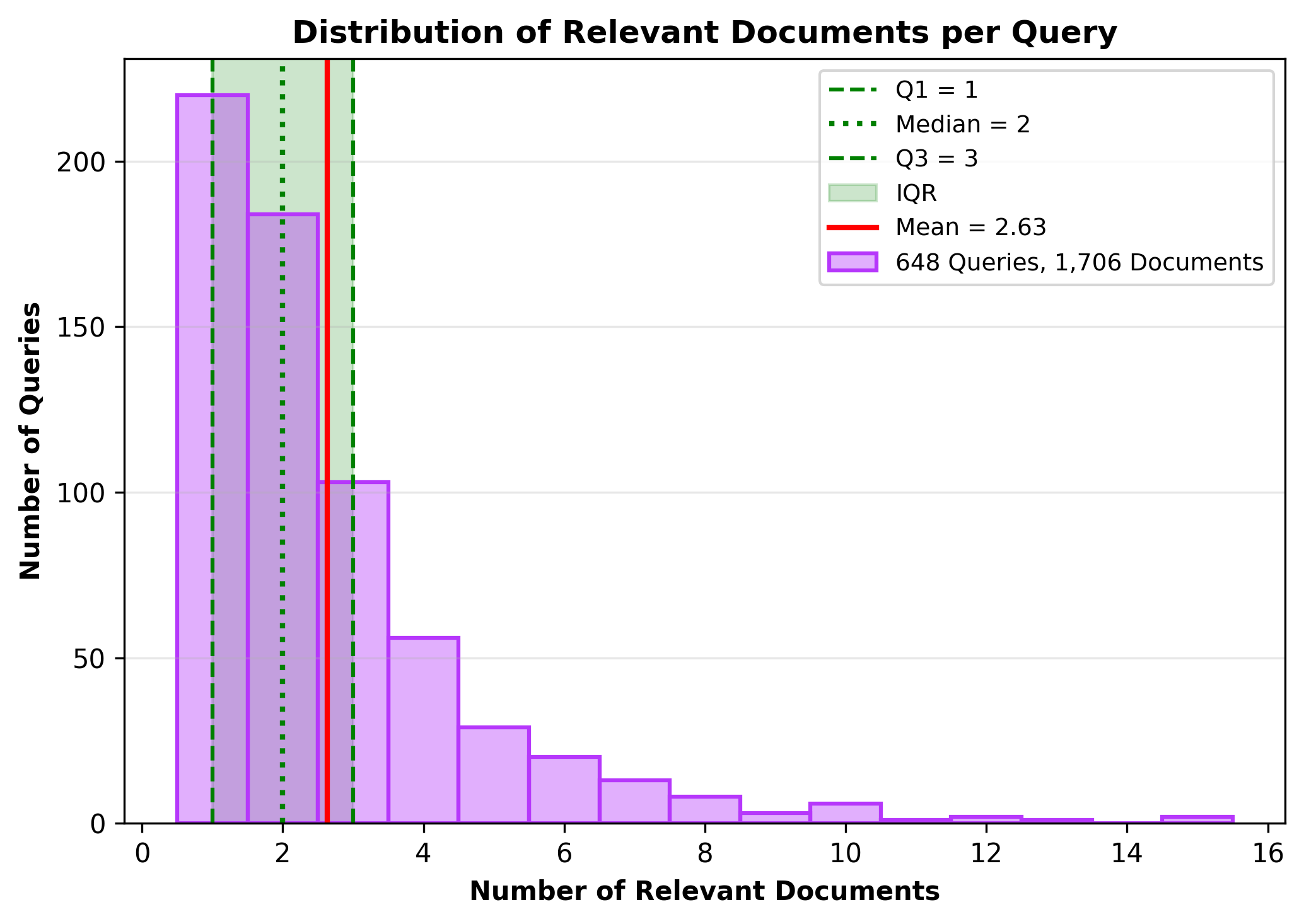}
    \caption{Distribution of relevant documents per query in FiQA-2018.}
    \label{fig:fiqa_relevant_docs}
\end{figure*}

Figure~\ref{fig:fiqa_relevant_docs} shows the distribution of relevant documents per query. The majority of queries have fewer than 3 relevant documents, with a median of 2 and a mean of 2.63. Despite this concentration, the distribution has a long tail extending up to 15 relevant documents for certain queries, indicating significant variability in query complexity. The narrow IQR between Q1 and Q3 confirms that most queries target specific, focused information. This fine-grained relevance pattern requires retrieval models to discriminate precisely between relevant and irrelevant content within a vast corpus of financial documents.

\subsection{EDA on FinanceBench}
\label{sec:financebench_analysis}
FinanceBench contains 150 questions grounded in documents across four types: 10-K (112), 10-Q (15), 8-K (9), and Earnings Transcripts (14), spanning 32 unique companies. These document types exhibit distinct lexical signatures and clearly different document structure: 10-K reports concentrate heavily on company- and industry-specific vocabulary, while Earnings Transcripts show the most diverse vocabulary, reflecting broader analyst Q\&A session patterns. This intra-corpus diversity means that a retrieval strategy optimized for one document type may systematically underperform on others, underscoring the need for type-aware retrieval approaches. These differences informed the construction of our type-specific agent prompts; top keywords by document type are shown in Table~\ref{tab:top_keywords_financebench}.

\begin{table*}[ht]
    \footnotesize
    \centering
    \caption{Top 6 keywords by document type in FinanceBench; ratios show corpus-relative frequency.}
    \label{tab:top_keywords_financebench}
    \resizebox{\textwidth}{!}{%
        \begin{tabular}{l r | l r | l r | l r}
        \toprule
        \multicolumn{2}{c|}{\textbf{10-K}} & \multicolumn{2}{c|}{\textbf{10-Q}} & \multicolumn{2}{c|}{\textbf{8-K}} & \multicolumn{2}{c}{\textbf{Earnings}} \\
        \textbf{Keyword} & \textbf{Ratio} & \textbf{Keyword} & \textbf{Ratio} & \textbf{Keyword} & \textbf{Ratio} & \textbf{Keyword} & \textbf{Ratio} \\
        \midrule
        
        securely & \scriptsize\colorbox{ailens_purple!50}{99.00\%} & 
        mun & \scriptsize\colorbox{ailens_purple!50}{98.61\%} & 
        accession & \scriptsize\colorbox{ailens_purple!50}{96.55\%} & 
        twentieth & \scriptsize\colorbox{ailens_purple!20}{58.90\%} \\
        
        restatement & \scriptsize\colorbox{ailens_purple!50}{99.00\%} & 
        min & \scriptsize\colorbox{ailens_purple!40}{85.71\%} & 
        chargeable & \scriptsize\colorbox{ailens_purple!40}{80.00\%} & 
        fragrance & \scriptsize\colorbox{ailens_purple!10}{43.33\%} \\
        
        athletic & \scriptsize\colorbox{ailens_purple!50}{99.00\%} & 
        noncompensation & \scriptsize\colorbox{ailens_purple!40}{83.13\%} & 
        survivorship & \scriptsize\colorbox{ailens_purple!30}{77.27\%} & 
        recalculate & \scriptsize\colorbox{ailens_purple!10}{42.31\%} \\
        
        george & \scriptsize\colorbox{ailens_purple!50}{98.99\%} & 
        aum & \scriptsize\colorbox{ailens_purple!40}{80.95\%} & 
        abstain & \scriptsize\colorbox{ailens_purple!30}{77.27\%} & 
        translational & \scriptsize\colorbox{ailens_purple!10}{40.24\%} \\
        
        budgetary & \scriptsize\colorbox{ailens_purple!50}{98.99\%} & 
        logged & \scriptsize\colorbox{ailens_purple!40}{80.77\%} & 
        unbudgeted & \scriptsize\colorbox{ailens_purple!30}{74.36\%} & 
        eighteenth & \scriptsize\colorbox{ailens_purple!10}{40.00\%} \\
        
        sport & \scriptsize\colorbox{ailens_purple!50}{98.99\%} & 
        nettable & \scriptsize\colorbox{ailens_purple!40}{80.00\%} & 
        bailiwick & \scriptsize\colorbox{ailens_purple!30}{73.85\%} & 
        shasta & \scriptsize\colorbox{ailens_purple!10}{36.21\%} \\
        \bottomrule
        \end{tabular}%
    }
\end{table*}
Table~\ref{tab:top_keywords_financebench} presents the top 6 keywords for each of the four document types in the corpus. The analysis reveals distinct lexical patterns. 10-K reports exhibit highly concentrated, company- and industry-specific vocabulary (98--99\%), with terms such as \textit{athletic} and \textit{sport} reflecting sector specialization. 10-Q and 8-K filings show strong concentration (74--98\%) with specialized financial and procedural terminology. Earnings Transcripts demonstrate the most diverse vocabulary (36--58\%), with comparative terms such as \textit{twentieth} and \textit{recalculate}, reflecting the broader conversational patterns of analyst Q\&A sessions. These distinct lexical profiles reinforce the need for retrieval strategies tailored to specific document types, and directly informed the construction of our type-specific agent prompts.

\subsection{Candidate Refinement Module Details}
\label{sec:filtering_appendix}

\begin{table}[!hb]
\centering
\caption{End-to-end effect of the candidate-refinement module on FinAgentBench chunk ranking under PRISM settings. Cost and latency are per query.}
\label{tab:finagent_e2e}
\resizebox{\columnwidth}{!}{%
\begin{tabular}{lccccc}
\toprule
\textbf{Pipeline} & \textbf{NDCG@5} & \textbf{MAP@5} & \textbf{MRR@5} & \textbf{\$/query} & \textbf{Lat.\,(s)} \\
\midrule
Baseline (Run 19) & 0.61278 & 0.48092 & 0.83758 & 0.2778 & 136.7 \\
TE3-S (top 100) & 0.61839 & 0.47800 & 0.84358 & 0.2336 & 192.0 \\
TE3-S (top 75) & 0.61298 & 0.46744 & 0.83908 & 0.2214 & 171.9 \\
TE3-S (top 50) & 0.61070 & 0.45060 & 0.82450 & 0.1845 & 142.4 \\
TE3-S$\to$Cohere (top 75) & 0.62418 & 0.47302 & 0.84450 & 0.2146 & 177.6 \\
\textbf{TE3-S$\to$Cohere (top 50)} & \textbf{0.63408} & \textbf{0.48719} & \textbf{0.85042} & \textbf{0.1813} & \textbf{146.3} \\
TE3-S$\to$Cohere$\to$$k$-recip (top 50) & 0.60976 & 0.46944 & 0.81925 & 0.1812 & 148.6 \\
\bottomrule
\end{tabular}}
\end{table}

\begin{table}[!ht]
\footnotesize
\centering
\caption{Embedding comparison on FiQA-2018 top-1000 pool.}
\label{tab:fiqa_embeddings}
\resizebox{\columnwidth}{!}{%
\begin{tabular}{lcccccc}
\toprule
\textbf{Retriever} & \textbf{Recall@100} & \textbf{Recall@200} & \textbf{Recall@300} & \textbf{NDCG@10} & \textbf{MAP@10} & \textbf{MRR@10} \\
\midrule
\textbf{TE3-S} & \textbf{0.7783} & \textbf{0.8423} & \textbf{0.8775} & \textbf{0.54988} & \textbf{0.36847} & \textbf{0.51945} \\
BGE-M3 (dense) & 0.7088 & 0.7800 & 0.8086 & 0.52972 & 0.33316 & 0.50064 \\
BGE-M3 (sparse) & 0.5133 & 0.5838 & 0.6242 & 0.36922 & 0.20623 & 0.32929 \\
SPLADE-v3 & 0.6683 & 0.7301 & 0.7615 & 0.49281 & 0.30611 & 0.45961 \\
BM25 & 0.3590 & 0.4223 & 0.4652 & 0.22782 & 0.11652 & 0.19854 \\
\bottomrule
\end{tabular}}
\end{table}

\begin{table}[!ht]
\centering
\caption{Embedding backbone selection on FinanceBench (150 QA). Cells show \emph{correct answers
(accuracy \%)} under shared-store conditions with prompt $P_4$, ICL-9.}
\label{tab:embedding_selection}
\resizebox{\columnwidth}{!}{%
\begin{tabular}{lccccc}
\toprule
\textbf{Embedding} & \textbf{GPT-5} & \textbf{gpt-oss-120b} & \textbf{DeepSeek-V4-Flash} & \textbf{grok-4-20-reasoning} & \textbf{Llama-4-Maverick-17B-128E-Instruct-FP8} \\
\midrule
\textbf{TE3-S} & \textbf{136 (90.7)} & \textbf{136 (90.7)} & \textbf{109 (72.7)} & 142 (94.7) & 86 (57.3) \\
Cohere embed-english-v3.0 & 136 (90.7) & 130 (86.7) & 106 (70.7) & \textbf{144 (96.0)} & \textbf{90 (60.0)} \\
Cohere embed-v4.0 & 135 (90.0) & 131 (87.3) & 108 (72.0) & 134 (89.3) & 86 (57.3) \\
\bottomrule
\end{tabular}}
\end{table}

\begin{table*}[ht]
\footnotesize
\centering
\caption{Reranker comparison on the FiQA-2018 TE3-S top-1000 pool.}
\label{tab:fiqa_rerankers}
\resizebox{\textwidth}{!}{%
\begin{tabular}{lcccccc}
\toprule
\textbf{Method} & \textbf{Recall@100} & \textbf{Recall@200} & \textbf{Recall@300} & \textbf{NDCG@10} & \textbf{MAP@10} & \textbf{MRR@10} \\
\midrule
Cohere rerank-v4.0-fast & \textbf{0.8115} & \textbf{0.8587} & \textbf{0.8940} & \textbf{0.6258} & \textbf{0.4370} & \textbf{0.6028} \\
BGE reranker v2 m3          & 0.7439  & 0.8061  & 0.8483 & 0.5431 & 0.3465 & 0.5114 \\
BGE-M3 ColBERT                 & 0.7270 & 0.7960 & 0.8366 & 0.5401 & 0.3467 & 0.5120 \\
ColBERTv2                   & 0.6690 & 0.7474 & 0.7928 & 0.4747 & 0.2865 & 0.4338 \\
Jina reranker v3            & 0.6497  & 0.7692 & 0.8209 & 0.4218 & 0.2244 & 0.3759 \\
\bottomrule
\end{tabular}}
\end{table*}

\begin{table}[!hb]
\centering
\caption{End-to-end effect of the candidate-refinement module on FiQA-2018 (TE3-S top-1000 pool).}
\label{tab:fiqa_refine}
\resizebox{\columnwidth}{!}{%
\begin{tabular}{lcccc}
\toprule
\textbf{Pipeline} & \textbf{Recall@100} & \textbf{NDCG@10} & \textbf{MAP@10} & \textbf{MRR@10} \\
\midrule
TE3-S & 0.77830 & 0.54988 & 0.36847 & 0.51945 \\
TE3-S$\to$Cohere & 0.81150 & 0.62580 & 0.43704 & 0.60276 \\
TE3-S$\to$Cohere$\to$GPT-5 & 0.83066 & 0.62426 & 0.43550 & 0.60122 \\
\textbf{TE3-S$\to$Cohere$\to$$k$-recip} & \textbf{0.83260} & \textbf{0.63333} & \textbf{0.45002} & \textbf{0.60058} \\
\bottomrule
\end{tabular}}
\end{table}

Operating on L2-normalized embeddings, where cosine ranking coincides with Euclidean ranking on the unit sphere, we define the base distance between a query $q$ and candidate $c_i$ as

\begin{equation}
d(q,c_i)=\lVert \hat{e}_q-\hat{e}_{c_i}\rVert_2^2 = 2 - 2\,\hat{e}_q^{\top}\hat{e}_{c_i},
\label{eq:basedist}
\end{equation}

and let $N(q,k_1)$ be the set of $k_1$ candidates with the smallest $d(q,\cdot)$. The
\emph{$k$-reciprocal} neighbor set keeps only mutual neighbors,

\begin{equation}
R(q,k_1)=\bigl\{\, c_i \in N(q,k_1) \mid q \in N(c_i,k_1) \,\bigr\}.
\label{eq:krecip}
\end{equation}

To recover true matches that fall just outside the top $k_1$ due to variation, we apply local expansion using a smaller neighborhood $k_2=\lfloor k_1/2\rfloor$, merging in a neighbor's reciprocal set only when the two sets overlap substantially,

\begin{equation}
R^{*}(q,k_1)=R\,\cup\!\!\bigcup_{\substack{c_j\in R\\[2pt] \lvert R\cap R_j\rvert\,\ge\,\frac{2}{3}\lvert R_j\rvert}}\!\! R_j.
\label{eq:expand}
\end{equation}

Writing $R\equiv R(q,k_1)$ and $R_j\equiv R(c_j,k_2)$ for brevity, each query/candidate is then encoded as a sparse vector over the candidate pool, where membership in $R^{*}$ is softened into a Gaussian (exponential) weight so that nearer reciprocal neighbors count more,

\begin{equation}
V_{q,c_i}=
\begin{cases}
e^{-d(q,c_i)}, & c_i \in R^{*}(q,k_1),\\
0, & \text{otherwise.}
\end{cases}
\label{eq:encode}
\end{equation}

The contextual distance between $q$ and $c_i$ is the Jaccard distance between their encoded neighbor vectors,

\begin{equation}
d_{J}(q,c_i)=1-\frac{\sum_{j=1}^{N}\min\!\bigl(V_{q,c_j},\,V_{c_i,c_j}\bigr)}{\sum_{j=1}^{N}\max\!\bigl(V_{q,c_j},\,V_{c_i,c_j}\bigr)},
\label{eq:jaccard}
\end{equation}

which is small when $q$ and $c_i$ share many highly weighted reciprocal neighbors, i.e.\ they are close not only directly but also through common neighbors (multi-hop similarity). The final distance linearly blends the direct and contextual signals,

\begin{equation}
d^{*}(q,c_i)=\lambda\,d(q,c_i)+(1-\lambda)\,d_{J}(q,c_i),
\label{eq:final}
\end{equation}

with $\lambda\in[0,1]$, and candidates are re-sorted in ascending $d^{*}$.

\noindent\textbf{Role of each variable.} Equation~\eqref{eq:basedist}'s $d(q,c_i)$ is the direct semantic distance from the embedding model. The L2 normalization makes it order-equivalent to cosine similarity. The reciprocity condition $q\in N(c_i,k_1)$ in Eq.~\eqref{eq:krecip} is the core robustness mechanism: it discards candidates that lie near $q$ but whose own neighborhoods do not contain $q$ (e.g.\ hub passages that are generically close to everything), keeping only genuinely mutual matches. The two neighborhood sizes trade recall against precision:

\begin{itemize}[itemsep=0pt]
  \item $k_1$ sets how many candidates are eligible as reciprocal neighbors. A larger $k_1$ admits more candidates (higher recall, more noise); a smaller $k_1$ is stricter (higher precision, risk of dropping true matches). Our implementation also allows an asymmetric query-side $k_1$, letting the query use a different neighborhood width from the candidates.
  \item $k_2\!=\!\lfloor k_1/2\rfloor$ controls local expansion (Eq.~\eqref{eq:expand}); the $\tfrac{2}{3}$ overlap threshold guards against merging unrelated neighbor sets, so expansion recovers near-miss true matches without injecting noise.
  \item The exponential weighting in Eq.~\eqref{eq:encode} converts hard set membership into soft weights, so the Jaccard overlap in Eq.~\eqref{eq:jaccard} emphasizes the closest shared neighbors rather than treating all neighbors equally.
  \item $\lambda$ in Eq.~\eqref{eq:final} balances the embedding/reranker score against neighborhood consensus: $\lambda\!\to\!1$ trusts the direct distance alone, while $\lambda\!\to\!0$ relies entirely on reciprocal-neighborhood overlap. Intermediate values let contextual consensus promote true positives that the single-vector distance underranks.
\end{itemize}

\noindent\textbf{Stage ordering.} Because reranking and $k$-reciprocal encoding are independent operations, we ablate their composition to quantify the contribution of each and the effect of order. Results across FiQA-2018 and FinAgentBench (Appendix~\ref{sec:filtering_fiqa} and Appendix~\ref{sec:filtering_finagent}) show that the benefit of $k$-reciprocal encoding is dataset-dependent, motivating 3R's modular, swap-in design.

\begin{table*}[ht]
\footnotesize
\centering
\caption{Reranker comparison on FinAgentBench chunk ranking.}
\label{tab:finagent_rerankers}
\resizebox{\textwidth}{!}{%
\begin{tabular}{lcccccc}
\toprule
\textbf{Method} & \textbf{Recall@25} & \textbf{Recall@50} & \textbf{Recall@75} & \textbf{NDCG@5} & \textbf{MAP@5} & \textbf{MRR@5} \\
\midrule
Cohere rerank-v4.0-fast     & \textbf{0.7850} & \textbf{0.8725} & \textbf{0.9088} & \textbf{0.5464} & \textbf{0.4070}  & \textbf{0.7632} \\
BGE reranker v2 m3    & 0.5975 & 0.7600 & 0.8675 & 0.3211 & 0.2330 & 0.5197   \\
BGE-M3 ColBERT & 0.6500 & 0.7875 & 0.8838 & 0.3115 & 0.2155 & 0.5113 \\
ColBERTv2   & 0.5825  & 0.7500 & 0.8675 & 0.2928 & 0.2053 & 0.5218 \\
Jina reranker v3        & 0.7200 & 0.8250 & 0.8888 & 0.4508 & 0.3071 & 0.6692  \\
\bottomrule
\end{tabular}}
\end{table*}

\subsubsection{Embedding Backbone Selection}
\label{sec:embedding_selection}
The retrieve stage is a dense retriever whose recall upper-bounds everything downstream. We compare text-embedding-3-small (TE3-S) against open dense, sparse, and late-interaction retrievers on FiQA-2018 and against Cohere embed-english-v3.0 and embed-v4.0 on FinanceBench, under identical pools and cutoffs. TE3-S attains the highest first-stage recall and ranking quality on FiQA (Appendix Table~\ref{tab:fiqa_embeddings}), and supplies the most consistent downstream QA accuracy across five LLM backbones on FinanceBench (Appendix Table~\ref{tab:embedding_selection}). We therefore adopt it as the default first-stage retriever, since it generalizes across retrieval-centric and answer-centric tasks and across model families, so one backbone serves both paths.

\subsubsection{Reranker Comparison and Selection}
\label{sec:filtering_fiqa}
We compare Cohere rerank-v4.0-fast, ColBERTv2, BGE-M3 ColBERT, BGE reranker v2 m3, and Jina reranker v3 on the same TE3-S pool for FiQA-2018 (top-1000 of the 57.6K-passage corpus; Appendix Table~\ref{tab:fiqa_rerankers}) and FinAgentBench chunk ranking (Appendix Table~\ref{tab:finagent_rerankers}). On FinAgentBench, refinement applies only to chunk ranking, since document ranking uses no embeddings and each query already arrives with a few hundred candidate chunks that the reranker compresses to the top ${\le}100$. One pattern holds across both datasets: reranker quality is decisive and weak rerankers degrade performance. Only Cohere improves over the TE3-S first stage on both recall and ranking, raising FiQA Recall@100 from 0.7783 to 0.8115 and NDCG@10 from 0.5499 to 0.6258, and reaching the best reranker-only Recall@50 of 0.8725 and NDCG@5 of 0.5464 on FinAgentBench. The others generally fall below the embedding baseline on recall, pushing relevant candidates below the cutoff instead of promoting them. Reranker selection, not merely its presence, is the design decision, and we adopt Cohere as the default.

\subsection{K-Reciprocal Encoding \& Stacking Order}
\label{sec:filtering_finagent}
The refine stage, $k$-reciprocal encoding, is a geometric, training-free step that reorders candidates by mutual-neighborhood consensus rather than by new semantic signal. We study $k$-reciprocal alone, both stacking orders with each reranker (reranker$\to$$k$-recip and $k$-recip$\to$reranker), and how the benefit depends on corpus size.

On the large FiQA corpus (Table~\ref{tab:fiqa_stacking}), $k$-reciprocal encoding is consistently beneficial when applied after a reranker, and the after order dominates the before order. It recovers recall that weaker rerankers discard, lifting Jina reranker v3 from $0.6497$ to $0.7809$ Recall@100, ColBERTv2 from $0.6690$ to $0.7520$, and BGE reranker v2 m3 from $0.7439$ to $0.7896$ while still adding to the strongest reranker, taking Cohere rerank-v4.0-fast from $0.8115$ to the best overall $0.8326$ Recall@100 (NDCG@10 $0.6333$). On a large pool with many near-neighbors, geometric consensus thus supplies complementary signal, disproportionately helping weak rerankers but still improving the strong one.

The picture inverts on FinAgentBench (Table~\ref{tab:finagent_stacking}). As each query is already paired with a few hundred topically related chunks that are filtered to ${\le}100$, the candidate set is small and semantically homogeneous, leaving little geometric structure for $k$-reciprocal encoding to exploit. No stacking configuration exceeds the reranker-only Cohere rerank-v4.0-fast result (cf.\ Table~\ref{tab:finagent_rerankers}), and stacking often harms it: applying $k$-reciprocal after Cohere rerank-v4.0-fast drops NDCG@5 from $0.5464$ to $0.2829$, as the geometric encoding reorders an already well-ordered, semantically tight set. The guidance is therefore corpus-dependent: stack $k$-reciprocal after a strong reranker for large, sparse corpora, but rely on the reranker alone for small, pre-filtered candidate sets, where the extra stage adds cost without accuracy gain.

\noindent\textbf{Where the crossover is.}
Not a single number, but a measurable predictor: how much recall the upstream stage leaves behind. Across the rerankers in Tables~\ref{tab:fiqa_stacking} and~\ref{tab:finagent_stacking}, the gain from adding $k$-reciprocal falls monotonically as the upstream stage's own Recall@100 rises, as Table~\ref{tab:krecip_crossover} shows. Two corpora cannot fix a threshold, but they do show what the threshold would be tracking. This is why the effect tracks corpus size, and it makes the guidance concrete rather than replacing it. FiQA's top-1000 pool is large and sparse, so a reranker still leaves relevant passages behind and $k$-reciprocal recovers some of them. FinAgentBench pairs each query with a few hundred topically related chunks already filtered to ${\le}100$, where Cohere reaches Recall@50 $=0.8725$ (Table~\ref{tab:finagent_rerankers}) against Recall@100 $=0.8115$ on FiQA (Table~\ref{tab:fiqa_rerankers}), so it is better covered at a tighter cutoff and has less headroom left. So the answer we can give is a test rather than a threshold: measure the upstream stage's recall at your operating cutoff, and add $k$-reciprocal only if it is leaving relevant candidates behind and a downstream ranker will re-order its output. Corpus size remains the right rule of thumb, and this is the quantity a practitioner can check directly on their own pool instead of guessing where the size boundary falls.

\begin{table}[ht]
\footnotesize
\centering
\caption{The gain from adding $k$-reciprocal encoding declines monotonically as upstream recall rises. FiQA-2018 TE3-S top-1000 pool, Recall@100.}
\label{tab:krecip_crossover}
\resizebox{\columnwidth}{!}{%
\begin{tabular}{lrrr}
\toprule
\textbf{Upstream reranker} & \textbf{Recall@100 alone} & $\boldsymbol{+\,k}$\textbf{-recip} & \textbf{Gain} \\
\midrule
Jina reranker v3 & 0.6497 & 0.7809 & $+0.131$ \\
ColBERTv2 & 0.6690 & 0.7520 & $+0.083$ \\
BGE-M3 ColBERT & 0.7270 & 0.7910 & $+0.064$ \\
BGE reranker v2 m3 & 0.7439 & 0.7896 & $+0.046$ \\
Cohere rerank-v4.0-fast & 0.8115 & 0.8326 & $+0.021$ \\
\bottomrule
\end{tabular}}
\end{table}

\begin{table*}[ht]
\footnotesize
\centering
\caption{Stacking $k$-reciprocal encoding with rerankers on FinAgentBench chunk ranking
(TE3-S top-100).}
\label{tab:finagent_stacking}
\resizebox{\textwidth}{!}{%
\begin{tabular}{lccccccc}
\toprule
\textbf{Method} & \textbf{Recall@25} & \textbf{Recall@50} & \textbf{Recall@75} & \textbf{Recall@100} & \textbf{NDCG@5} & \textbf{MAP@5} & \textbf{MRR@5} \\
\midrule
TE3-S$\to$$k$-recip (only) & 0.5925 & 0.7850 & 0.8875 & 0.9325 & 0.28638 & 0.20450 & 0.44708 \\
\midrule
\multicolumn{8}{c}{\cellcolor{ailens_purple!35}\textit{$k$-reciprocal $\to$ reranker (before)}} \\
\midrule
TE3-S$\to$$k$-recip$\to$Cohere rerank-v4.0-fast     & 0.7713 & 0.8575 & - & - & 0.54171 & 0.40515 & 0.75558  \\
TE3-S$\to$$k$-recip$\to$BGE-M3 ColBERT & 0.6625 & 0.8075    & - & - & 0.31584 & 0.21854 & 0.52108 \\
TE3-S$\to$$k$-recip$\to$BGE reranker v2 m3      & 0.5975   & 0.7863 & - & - & 0.32410 & 0.23556 & 0.52400 \\
TE3-S$\to$$k$-recip$\to$Jina reranker v3        & 0.7188 & 0.8250  & - & - & 0.44356 & 0.30575 & 0.64333 \\
TE3-S$\to$$k$-recip$\to$ColBERTv2   & 0.5888  & 0.7488   & - & - & 0.29552 & 0.20675 & 0.52258  \\
\midrule
\multicolumn{8}{c}{\cellcolor{ailens_purple!35}\textit{reranker $\to$ $k$-reciprocal (after)}} \\
\midrule
TE3-S$\to$Cohere rerank-v4.0-fast$\to$$k$-recip      & 0.6625 & 0.8725  & - & - & 0.28286 & 0.20215  & 0.44825 \\
TE3-S$\to$BGE-M3 ColBERT$\to$$k$-recip & 0.6313 & 0.8113    & - & - & 0.28828 & 0.20427 & 0.46533   \\
TE3-S$\to$BGE reranker v2 m3$\to$$k$-recip      & 0.6288   & 0.8013 & - & - & 0.29497 & 0.20388 & 0.45925 \\
TE3-S$\to$Jina reranker v3$\to$$k$-recip        & 0.6338 & 0.8138  & - & - & 0.29448 & 0.20833 & 0.47100 \\
TE3-S$\to$ColBERTv2$\to$$k$-recip   & 0.5988 & 0.8088   & - & - & 0.29540 & 0.21052 & 0.47683 \\
\midrule
\multicolumn{8}{c}{\cellcolor{ailens_purple!35}\textit{End to end pipeline}} \\
\midrule
TE3-S (top-100)$\to$GPT-5   & 0.6938 & 0.8263 & 0.8888 & 0.9325 & 0.61839 & 0.47800 & 0.84358 \\
TE3-S (top-75)$\to$GPT-5   & 0.6938 & 0.8263 & 0.8888 & - & 0.61298 & 0.46744 & 0.83908 \\
TE3-S (top-100)$\to$Cohere rerank-v4.0-fast (top-75)$\to$GPT-5   & 0.7850 & 0.8725 & 0.9088 & - & 0.62418 & 0.47302 & 0.84450 \\
TE3-S (top-50)$\to$GPT-5   & 0.6938 & 0.8263 & - & - & 0.61070 & 0.45060 & 0.82450 \\
TE3-S (top-100)$\to$Cohere rerank-v4.0-fast (top-50)$\to$GPT-5   & 0.7850 & 0.8725 & - & - & 0.63408 & 0.48719 & 0.85042 \\
TE3-S (top-100)$\to$$k$-recip$\to$Cohere rerank-v4.0-fast(top-50)$\to$GPT-5   & 0.7713 & 0.8575 & - & - & 0.60976 & 0.46944 & 0.81925 \\
\bottomrule
\end{tabular}}
\end{table*}

\begin{table*}[ht]
\footnotesize
\centering
\caption{Stacking $k$-reciprocal encoding with rerankers on FiQA-2018 (TE3-S top-1000).}
\label{tab:fiqa_stacking}
\resizebox{\textwidth}{!}{%
\begin{tabular}{lcccccc}
\toprule
\textbf{Method} & \textbf{Recall@100} & \textbf{Recall@200} & \textbf{Recall@300} & \textbf{NDCG@10} & \textbf{MAP@10} & \textbf{MRR@10} \\
\midrule
TE3-S$\to$$k$-recip (only) & 0.7824 & 0.8484 & 0.8839 & 0.53021 & 0.34970 & 0.48821 \\
\midrule
\multicolumn{7}{c}{\cellcolor{ailens_purple!35}\textit{$k$-reciprocal $\to$ reranker (before)}} \\
\midrule
TE3-S$\to$$k$-recip$\to$Cohere rerank-v4.0-fast     & 0.8140   & 0.8687  & - & 0.62853 & 0.44082 & 0.60662 \\
TE3-S$\to$$k$-recip$\to$BGE-M3 ColBERT & 0.7520 & 0.8356 & - & 0.54163 & 0.34939 & 0.51403 \\
TE3-S$\to$$k$-recip$\to$BGE reranker v2 m3      & 0.7690   & 0.8490   & - & 0.54820  & 0.35141 & 0.51668 \\
TE3-S$\to$$k$-recip$\to$Jina reranker v3       & 0.7755 & 0.8595 & - & 0.49708 & 0.28377 & 0.45946 \\
TE3-S$\to$$k$-recip$\to$ColBERTv2   & 0.7160 & 0.8114 & - & 0.48290  & 0.29042 & 0.43978 \\
\midrule
\multicolumn{7}{c}{\cellcolor{ailens_purple!35}\textit{reranker $\to$ $k$-reciprocal (after)}} \\
\midrule
TE3-S$\to$Cohere rerank-v4.0-fast$\to$$k$-recip      & \textbf{0.8326} & \textbf{0.8743} & - & \textbf{0.63333} & \textbf{0.45002} & 0.60058 \\
TE3-S$\to$BGE-M3 ColBERT$\to$$k$-recip & 0.7910 & 0.8247 & - & 0.56720  & 0.38213 & 0.53533 \\
TE3-S$\to$BGE reranker v2 m3$\to$$k$-recip      & 0.7896  & 0.8332  & - & 0.58122 & 0.39079 & 0.55393 \\
TE3-S$\to$Jina reranker v3$\to$$k$-recip        & 0.7809 & 0.8132 & - & 0.58378 & 0.38692 & 0.55994 \\
TE3-S$\to$ColBERTv2$\to$$k$-recip   & 0.7520 & 0.7826 & - & 0.48667 & 0.29597 & 0.44702 \\
\midrule
\multicolumn{7}{c}{\cellcolor{ailens_purple!35}\textit{End to end pipeline}} \\
\midrule
TE3-S$\to$Cohere rerank-v4.0-fast$\to$GPT-5   & 0.8307 & 0.8566 & - & 0.62426 & 0.43550 & 0.60122 \\
\bottomrule
\end{tabular}}
\end{table*}

\paragraph{When to add $k$-reciprocal encoding.}
\label{sec:krecip_rule}
Corpus size is the usual rule of thumb, but the quantity it tracks is directly measurable: how much recall the upstream stage leaves behind. Across the five rerankers on the same FiQA pool, the gain from adding $k$-reciprocal falls monotonically as the upstream stage's own Recall@100 rises, from $+0.131$ at 0.6497 (Jina v3) to $+0.021$ at 0.8115 (Cohere), passing through $+0.083$, $+0.064$, and $+0.046$ (Appendix Tables~\ref{tab:fiqa_stacking} and~\ref{tab:finagent_stacking}). FinAgentBench sits at the saturated end of that same relationship rather than in a separate regime: its chunk pools arrive filtered to ${\le}100$ and Cohere reaches Recall@50 of 0.8725 there, so adding the stage instead drops reranker-only chunk NDCG@5 from 0.5464 to 0.2829. The guidance is therefore a test rather than a threshold: add $k$-reciprocal only if the upstream stage is still leaving relevant candidates behind at your operating cutoff and a downstream ranker will re-order its output. For FinAgentBench-like pre-filtered pools, use Cohere refinement alone, at the price of roughly 10\,s of added latency per query.

\subsection{Different Prompt Versions}
\label{sec:prompt_version}
The non-agentic prompt templates are designed for single LLM workflows, where the model independently performs the complete ranking process using structured system instructions. These prompts emphasize logical consistency, domain alignment, and interpretable reasoning. For document ranking, the progression moves from domain-guided heuristics to combined reasoning frameworks. For chunk ranking, prompts evolve from a pure ReAct framework to integrated multi-strategy approaches. Specifically, the prompt iterations are as follows:

\begin{itemize}
    \item \(P_1\): For document ranking, introduces domain-specific keyword hints and statistical cues to guide relevance assessment. For chunk ranking, employs a ReAct-style framework that interleaves internal reasoning with action steps, requiring the model to think step-by-step before outputting ranked indices.
    \item \(P_2\): For document ranking, enhances \(P_1\) with detailed statistical keyword associations (e.g., word frequency ratios per document type) to improve discriminative power. For chunk ranking, combines Chain-of-Thought (CoT), ReAct, and Tree-of-Thought (ToT) strategies, introducing multi-expert simulation where three analysts reason collaboratively and prune incorrect paths.
    \item \(P_3\): Adopts a unified multi-strategy framework for both document and chunk ranking, integrating CoT decomposition, ReAct alternation, and ToT multi-expert simulation. Emphasizes explicit step-by-step reasoning with domain-specific hints for financial concepts.
    \item \(P_4\): Refines the \(P_3\) framework with streamlined instructions and consistent terminology across document and chunk ranking tasks, maintaining the combined CoT, ReAct, and ToT reasoning strategies.
\end{itemize}

The prompts differ in a specific way: \(P_1\) supplies domain keyword hints and statistical cues only, while \(P_2\) to \(P_4\) add a structured relevance scaffold combining CoT decomposition, ReAct alternation and ToT multi-expert simulation. \(P_4\) is \(P_3\) with streamlined instructions and terminology unified across the document and chunk stages, so the difference between them is consistency and output discipline, not length or verbosity.

Matched on model and ICL, GPT-5-mini at both stages with no exemplars, Table~\ref{tab:prompt_both_bases} reports the four prompts on both evaluation bases. The explanation therefore has two parts. The step that carries the effect is adding the scaffold: \(P_1\) is the weakest of the four on both evaluation bases. Within the scaffolded set, \(P_4\) is the strongest on the re-annotated set and within 0.0012 of the strongest on the original leaderboard, which is why it is standardized for the later experiments: it is the only prompt that is at or near the top on both bases, so it is the safest default rather than a large win. We state the claim at that strength, as reliability rather than as a per-prompt gap, which is also consistent with the observation that the best prompt form differs by benchmark.

We also make explicit a dependency that is easy to miss: chunk scores depend on the document stage, because the chunk ranker only sees chunks from the documents the first stage selected. That is why prompt effects are discussed at the pipeline level rather than attributed to the chunk prompt alone.

\begin{table}[ht]
\footnotesize
\centering
\caption{Prompt variants matched on model and ICL (GPT-5-mini at both stages, no exemplars), on both evaluation bases. Combined NDCG@5.}
\label{tab:prompt_both_bases}
\begin{tabular}{ccrr}
\toprule
\textbf{Prompt} & \textbf{Run} & \textbf{Re-annotated} & \textbf{Original private} \\
\midrule
\(P_1\) & 2 & 0.75759 & 0.66628 \\
\(P_2\) & 9 & 0.76649 & 0.66774 \\
\(P_3\) & 12 & 0.76287 & 0.69019 \\
\(P_4\) & 13 & 0.77668 & 0.68895 \\
\bottomrule
\end{tabular}
\end{table}

The complete prompt templates for document ranking (Table~\ref{tab:doc-prompts}) and chunk ranking (Table~\ref{tab:chunk-prompts}) are provided in Appendix~\ref{sec:prompt_template}.

\subsection{In-Context Learning Retrieval Details}
\label{sec:icl_retrieval_details}
The FAISS index is built offline from the training split: each query--document/chunk pair is embedded using OpenAI text-embedding-3-small, and all embeddings are stored with their ground-truth rankings. At inference time, the $k$ nearest neighbors of the incoming query are retrieved by cosine similarity and formatted as ordered few-shot demonstrations prepended to the system prompt. The resulting prompt is therefore dynamic, adapting to the semantic characteristics of each user query rather than relying on a static template. ICL exemplar count $k$ was varied from 5 to 15 in ablation experiments. Beyond $k=5$, chunk ranking performance degraded due to context overload, whereas document ranking benefited from $k=5$ shots with no gain from additional examples.

\noindent\textbf{ICL retriever against candidate-refinement retriever.}
They are not redundant: the two share TE3-S as an embedding backbone but query different indexes and act at different points in the pipeline (Table~\ref{tab:two_retrievers}). The distinction is visible empirically, because the two have opposite effects on the same stage. ICL improves document ranking but is not the default for chunk ranking, since exemplars consume the context chunk evidence needs. Candidate refinement shortens chunk input by shrinking the pool the LLM sees: on FinAgentBench chunk ranking, TE3-S to Cohere top-50 yields a higher observed NDCG@5 (0.63408 vs. 0.61278) and reduces cost from USD 0.2778 to USD 0.1813 per query (Appendix Table~\ref{tab:finagent_e2e}). One adds context, the other removes it, and the results follow that difference.

\begin{table}[ht]
\footnotesize
\centering
\caption{The two retrievers in PRISM share an embedding backbone but query different indexes and act at different points.}
\label{tab:two_retrievers}
\resizebox{\columnwidth}{!}{%
\begin{tabular}{lll}
\toprule
 & \textbf{ICL retriever} & \textbf{Candidate-refinement retriever} \\
\midrule
Index queried & training-split exemplar store & document or chunk corpus for the current query \\
Output & few-shot demonstrations prepended to the prompt & candidates to be ranked \\
Changes the candidate set? & no & yes \\
Role & task adaptation & pool construction and context compression \\
\bottomrule
\end{tabular}}
\end{table}

\subsection{Multi-Agent System Architecture Details}
\label{sec:mas_architecture_details}

This section provides comprehensive implementation details for the Multi-Agent System (MAS) modeling architectures used in PRISM, including the document ranking workflow and four chunk ranking variants (\(A_1\)--\(A_4\)). All architectures are illustrated in Figure~\ref{fig:mas_architectures}.

\subsubsection{Agent Definitions} 
We adopt a state-graph framework implemented using LangGraph~\cite{langgraph}, where each agent is an LLM instance with a specialized system prompt defining its role and scoring criteria. Document ranking employs a fixed configuration: a \textit{Question Analyzer} that extracts key entities and intent from the query, followed by five \textit{Document Expert} agents (one per SEC filing type: 10-K, 10-Q, 8-K, DEF~14A, and Earnings Transcripts) that score documents based on domain-specific relevance criteria. For chunk ranking, we define specialized agents including: \textit{Financial Analyst} (evaluates quantitative metrics and financial data), \textit{Risk Analyst} (assesses operational and regulatory risks), \textit{Evidence Extractor} (identifies concrete supporting facts), \textit{Contextual Reasoner} (evaluates explanatory content), and filtering agents such as \textit{Quick Filter} and \textit{Noise Remover}.

\subsubsection{Interaction Process} 
Agents communicate through a shared state graph where each node represents an agent and edges define information flow. In a typical workflow: (1)~input chunks are passed to filtering agents that assign preliminary relevance scores; (2)~filtered candidates flow to scoring agents that independently evaluate chunks on a 1--10 scale based on their specialized criteria; (3)~scores are aggregated through weighted averaging or voting to produce final rankings. The graph topology determines whether agents operate in parallel (simultaneous scoring) or sequentially (pipeline filtering). All agent prompts are provided in Table~\ref{tab:agentic-chunk-prompts}.

\subsubsection{State Management and Information Flow} 
All MAS architectures maintain a shared state graph implemented through LangGraph's StateGraph abstraction. Information flows through the graph according to predefined edges connecting agent nodes. Parallel agents (e.g., scoring agents in \(A_1\)) execute simultaneously without inter-agent communication, reducing coordination complexity. Sequential pipelines (\(A_2\), \(A_3\)) enforce strict ordering where downstream agents only receive filtered candidates from upstream stages, creating potential bottlenecks if early filtering is overly aggressive. The state contains the input data, intermediate outputs, and aggregated results.

\subsubsection{Run-Level Findings and Deployment Implications}
After defining the document agents, chunk-ranking variants, and LangGraph information flow above, we summarize how these MAS designs behave empirically. The ablations show that agentic workflows are constrained by model capacity and graph depth. With GPT-4o-mini, deeper chunk-ranking pipelines perform poorly: \(A_2\) reaches 0.51994 combined NDCG@5 (Run~22) and \(A_3\) reaches 0.52031 (Run~24), indicating that filtering and scoring errors compound across sequential stages. Scaling the same \(A_3\) design to GPT-5-mini raises combined NDCG@5 to 0.72353 (Run~25), showing that MAS requires sufficiently capable agents to maintain coherent intermediate outputs.

Hybrid workflows further clarify where MAS is useful. Configurations that keep agentic reasoning at the document level but use a non-agentic chunk ranker perform strongly, reaching 0.77104 (Run~35) and 0.78157 (Run~36). This supports the deployment recommendation in the main text: use MAS selectively for coarse document-level routing or reasoning, but prefer simpler single-model or lightly coordinated designs for chunk ranking, where latency, token cost, and false-negative filtering risk are higher.

\subsubsection{Document Ranking Workflow}
The document ranking workflow (Figure~\ref{fig:document_agent}) follows a two-stage architecture designed to handle heterogeneous financial document types with specialized domain expertise. The workflow begins with a \textit{Question Analyzer} agent that processes the user query to extract key entities and classify the query relevance to the corresponding agents' expertise. This structured analysis is then distributed to five parallel \textit{Document Expert} agents, each specialized in a specific SEC filing type:

\begin{itemize}
    \item 10-K Expert: Evaluates annual reports, focusing on comprehensive financial statements, business operations, and risk factors disclosed in Form 10-K filings.
    \item 10-Q Expert: Assesses quarterly reports, prioritizing interim financial performance, management discussion and analysis (MD\&A), and quarterly updates.
    \item 8-K Expert: Analyzes current reports for material events, corporate announcements, and time-sensitive disclosures.
    \item DEF14A Expert: Reviews proxy statements for governance information, executive compensation, and shareholder proposals.
    \item Earnings Transcript Expert: Evaluates earnings call transcripts for management commentary, forward guidance, and analyst Q\&A insights.
\end{itemize}

Each document expert scores documents on a 0--4 scale based on relevance to the analyzed query intent. Scores are then aggregated through weighted averaging, where weights are determined by document type relevance to the query category provided by the question analyzer. This parallel scoring approach ensures domain-appropriate evaluation while maintaining computational efficiency, as document experts operate simultaneously without sequential dependencies.

\subsubsection{Chunk Ranking Workflow Variants}
\begin{figure*}[htbp]
    \centering
    \begin{subfigure}[b]{0.405\textwidth}
        \centering
        \includegraphics[width=\textwidth]{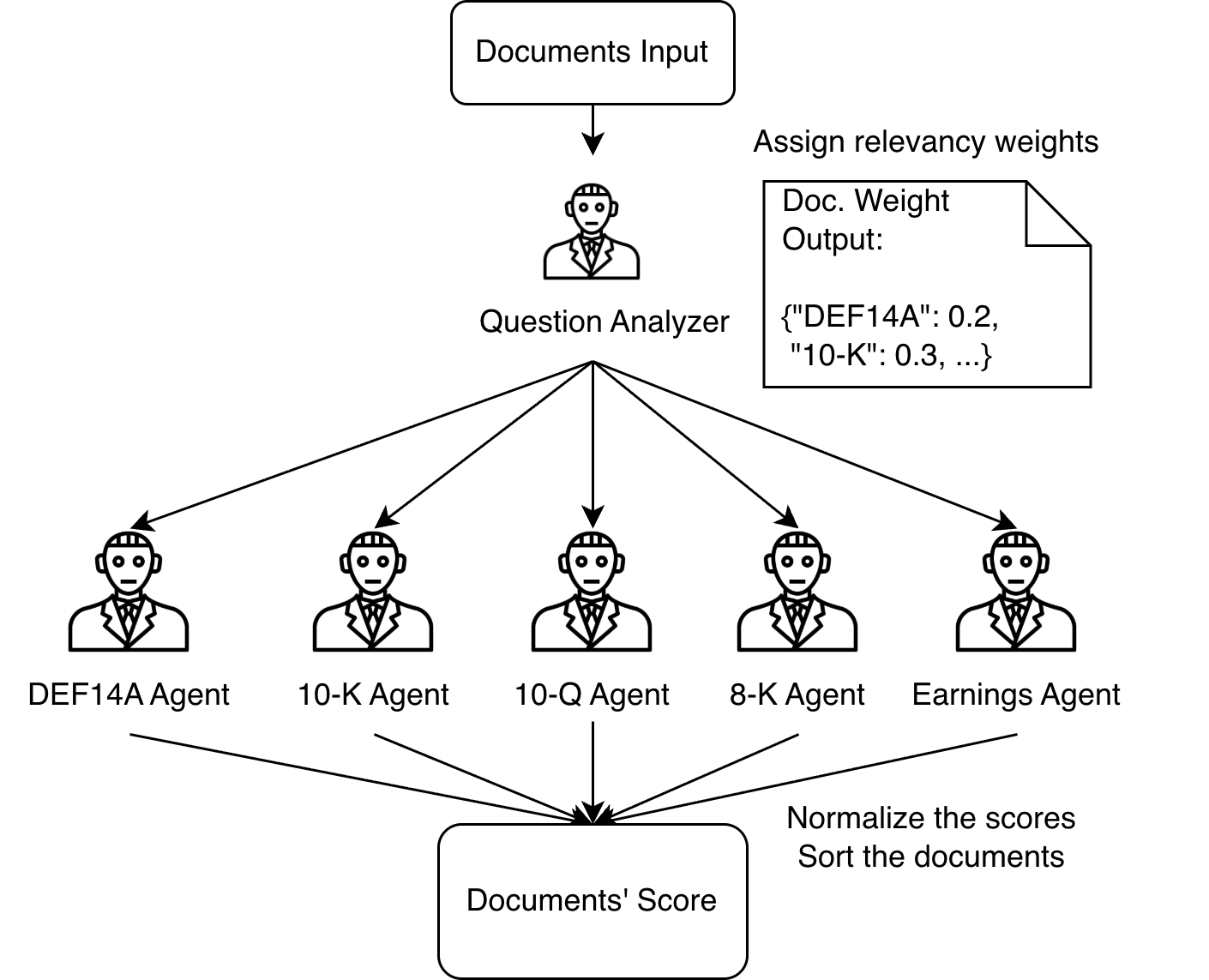}
        \caption{Document ranking workflow}
        \label{fig:document_agent}
    \end{subfigure}
    \hfill
    \begin{subfigure}[b]{0.405\textwidth}
        \centering
        \includegraphics[width=\textwidth]{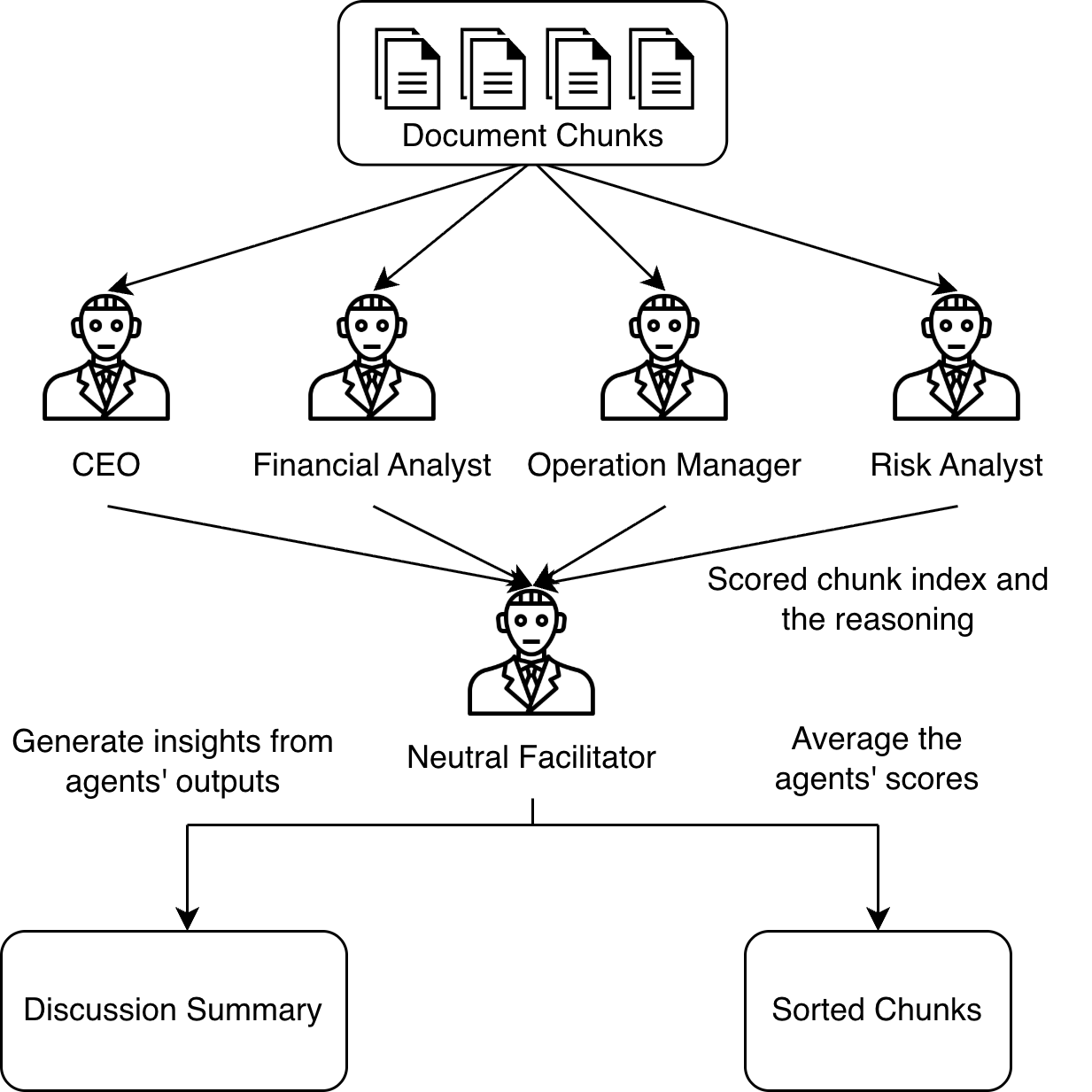}
        \caption{\(A_1\): Direct scoring ensemble}
        \label{fig:chunk_ranking1}
    \end{subfigure}
    
    \begin{subfigure}[b]{0.405\textwidth}
        \centering
        \includegraphics[width=\textwidth]{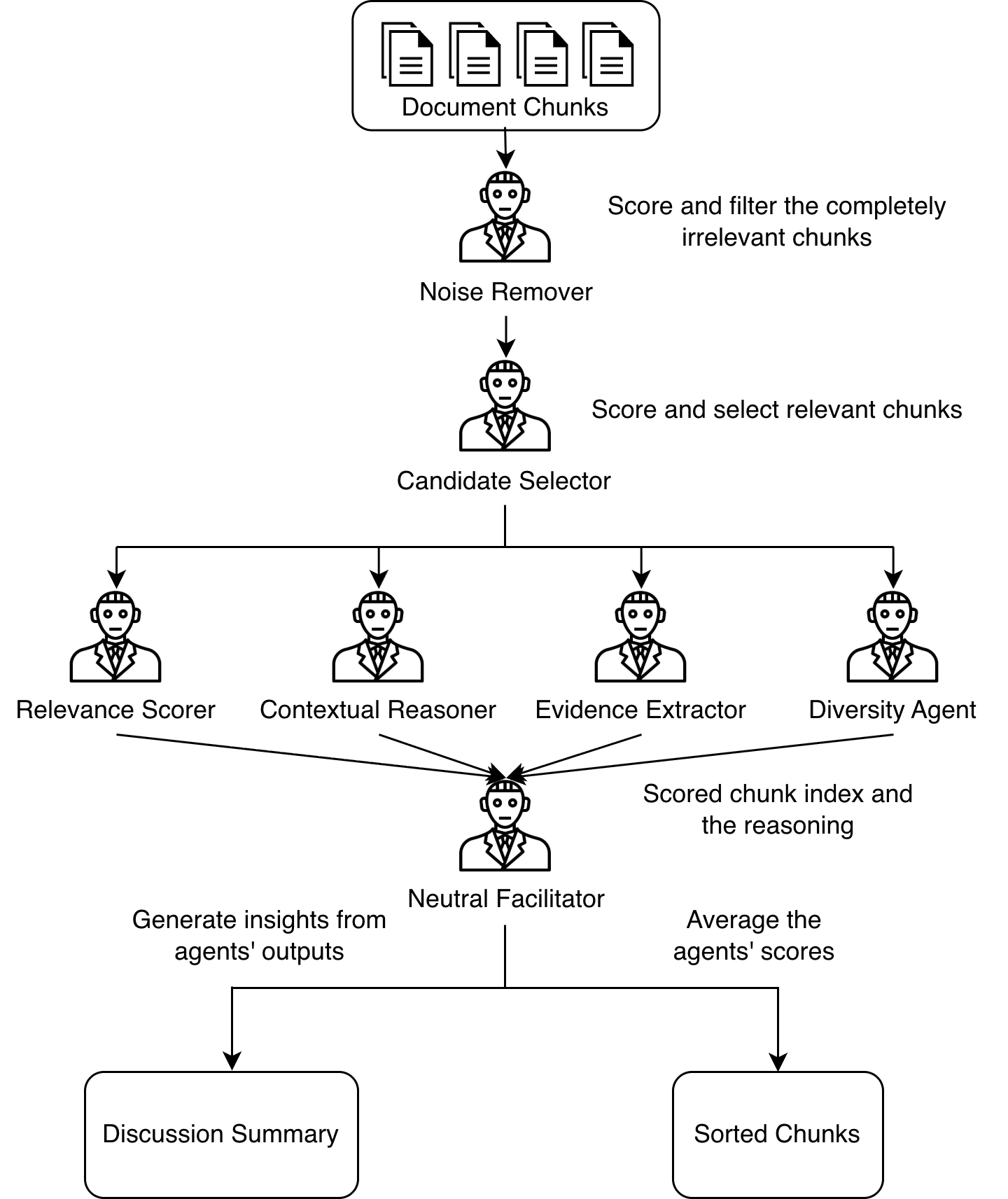}
        \caption{\(A_2\): Three-stage deep pipeline}
        \label{fig:chunk_ranking2}
    \end{subfigure}
    \hfill
    \begin{subfigure}[b]{0.405\textwidth}
        \centering
        \includegraphics[width=\textwidth]{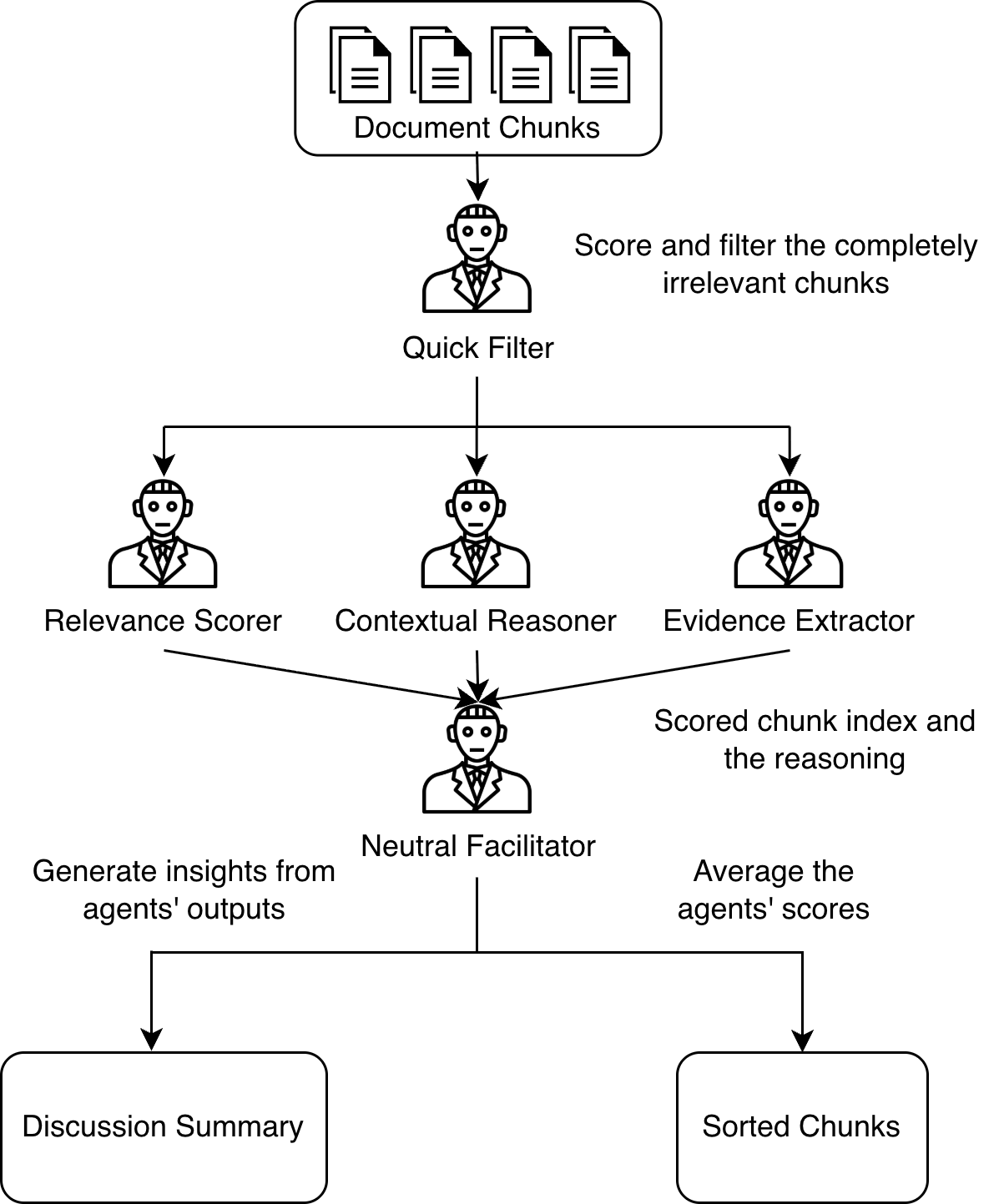}
        \caption{\(A_3\): Two-stage balanced pipeline}
        \label{fig:chunk_ranking3}
    \end{subfigure}
    
    \begin{subfigure}[b]{0.405\textwidth}
        \centering
        \includegraphics[width=\textwidth]{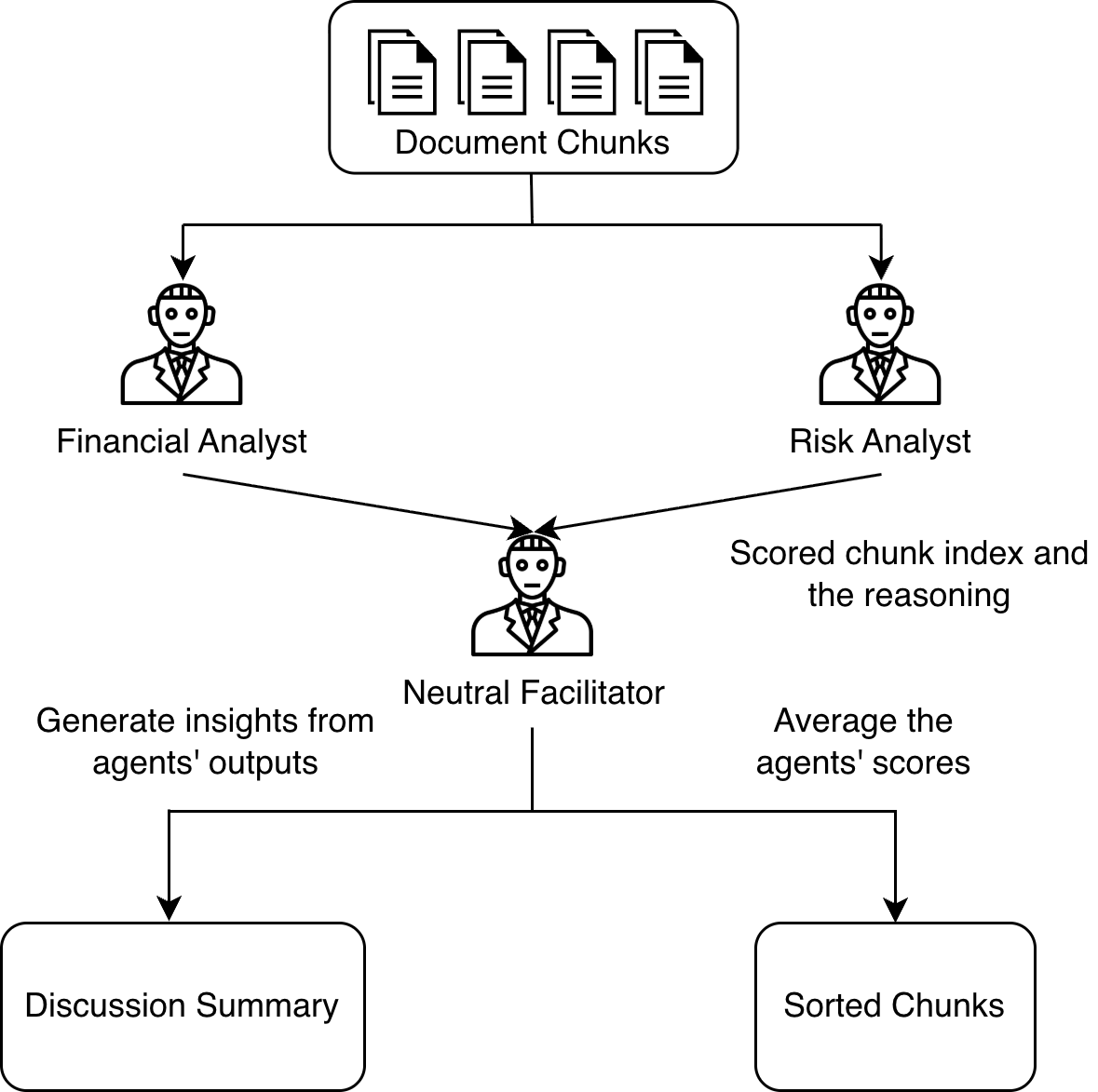}
        \caption{\(A_4\): Minimal two-agent system}
        \label{fig:chunk_ranking4}
    \end{subfigure}
    
    \caption{Multi-Agent System architectures for PRISM. (a)~Document ranking workflow. (b)--(e)~Chunk ranking variants.}
    \label{fig:mas_architectures}
\end{figure*}

The chunk ranking workflows (Figures~\ref{fig:chunk_ranking1}--\ref{fig:chunk_ranking4}) represent four architectural variants with progressively simplified designs, each exploring different trade-offs between reasoning depth, error propagation, and computational cost.

\(A_1\): Direct Scoring Ensemble (Figure~\ref{fig:chunk_ranking1}) implements a single-stage parallel scoring approach with four domain-expert agents operating simultaneously:
\begin{itemize}
    \item \textit{CEO Agent}: Evaluates strategic relevance and high-level business implications.
    \item \textit{Financial Analyst}: Assesses quantitative metrics, financial ratios, and numerical evidence.
    \item \textit{Operations Manager}: Focuses on operational details, business processes, and execution aspects.
    \item \textit{Risk Analyst}: Identifies risk factors, uncertainties, and potential adverse impacts.
\end{itemize}

Each agent independently scores all candidate chunks on a 1--10 scale according to its specialized criteria. Final rankings are determined through simple averaging of the four scores. This architecture minimizes pipeline depth to reduce error accumulation and analyzes chunks from different perspectives.

\(A_2\): Three-Stage Deep Pipeline (Figure~\ref{fig:chunk_ranking2}) implements the most complex workflow with three sequential stages designed for sequential filtering and specialized evaluation:

\begin{enumerate}[itemsep=0pt]
    \item Noise Removal Stage: A \textit{Noise Remover} agent filters out clearly irrelevant chunks (boilerplate text, legal disclaimers, table fragments) to reduce downstream computational burden.
    \item Candidate Selection Stage: A \textit{Candidate Selector} agent performs secondary filtering, retaining only chunks with potential relevance based on keyword matching and semantic proximity.
    \item Ensemble Scoring Stage: Four specialized agents score the filtered candidates:
    \begin{itemize}
        \item \textit{Relevance Scorer}: Direct query-chunk semantic alignment.
        \item \textit{Contextual Reasoner}: Evaluates explanatory content and contextual completeness.
        \item \textit{Evidence Extractor}: Identifies concrete supporting facts and data points.
        \item \textit{Diversity Agent}: Ensures ranking diversity to avoid redundant information.
    \end{itemize}
\end{enumerate}

While this deep architecture aims to progressively refine candidates, our experiments reveal it suffers from over-filtering and cascading errors, where early-stage false negatives cannot be recovered in later stages, leading to degraded recall performance.

\(A_3\): Two-Stage Balanced Pipeline (Figure~\ref{fig:chunk_ranking3}) simplifies \(A_2\) by reducing to two stages while maintaining filtering benefits:
\begin{enumerate}[itemsep=0pt]
    \item Quick Filtering Stage: A \textit{Quick Filter} agent performs lightweight relevance assessment, applying looser thresholds than \(A_2\)'s Noise Remover to preserve recall.
    \item Specialized Scoring Stage: Three focused agents evaluate filtered chunks:
    \begin{itemize}
        \item \textit{Relevance Scorer}: Query-chunk alignment assessment.
        \item \textit{Contextual Reasoner}: Contextual adequacy and explanatory quality.
        \item \textit{Evidence Extractor}: Concrete evidence identification.
    \end{itemize}
\end{enumerate}
This architecture balances pipeline depth with stability, reducing error propagation risks while maintaining some computational efficiency gains from preliminary filtering.

\(A_4\): Minimal Two-Agent System (Figure~\ref{fig:chunk_ranking4}) represents the simplest viable MAS configuration with direct parallel scoring by two complementary agents:
\begin{itemize}
    \item \textit{Financial Analyst}: Focuses on quantitative evidence and financial metrics.
    \item \textit{Risk Analyst}: Evaluates qualitative risk factors and uncertainties.
\end{itemize}
By eliminating filtering stages and reducing agent count, it minimizes interaction complexity and error propagation pathways. This minimalist design performed comparably to or better than deeper architectures in our experiments, as reduced error propagation outweighs ensemble benefits.

\subsubsection{Prompt Engineering for Agent Roles} 
Each agent receives a specialized system prompt defining its role, evaluation criteria, and output format. For document ranking, expert prompts specify document-type-specific relevance factors. For chunk ranking, scoring agents receive detailed rubrics mapping scores 1--10 to specific evidence characteristics. Filtering agents use binary decision criteria with explicit instructions to prefer recall over precision. All agent prompts are provided in Table~\ref{tab:agentic-doc-prompts} and Table~\ref{tab:agentic-chunk-prompts}.

\subsubsection{Aggregation and Ranking Logic} 
Score aggregation varies by architecture. Simple averaging is used in chunk ranking where all agents have equal weight. In document ranking, we implement weighted averaging where weights reflect agent reliability and query-dependent relevance. Final rankings are produced by sorting candidates by aggregated score in descending order, with ties broken by original retrieval rank.

\subsection{Implementation Details}
\label{sec:implementation_details}
\subsubsection{Model Provider Selection}
Experiments used an Azure-hosted API deployment for enterprise access control and reproducibility, with all models accessed through the Azure ecosystem. This deployment setting motivated our focus on OpenAI GPT models as the closed-source model family, while the open-source models assess PRISM's generalizability beyond a single model family. Beyond operational accessibility, OpenAI's models remain among the most benchmarked proprietary LLMs on financial tasks~\cite{bigeard2025finance}, offering strong reliability as a baseline. The exact versions of the eight foundation models used in our experiments are provided below:
\begin{enumerate}[itemsep=0pt]
    \item GPT-4o-mini: gpt-4o-mini-2024-07-18
    \item GPT-4.1: gpt-4.1-2025-04-14
    \item GPT-5-mini: gpt-5-mini-2025-08-07
    \item GPT-5: gpt-5-2025-08-07
    \item DeepSeek-V4-Flash
    \item gpt-oss-120b
    \item grok-4-20-reasoning
    \item Llama-4-Maverick-17B-128E-Instruct-FP8
\end{enumerate}
The retrieval pipeline was implemented using a FAISS vector store with two OpenAI embedding backbones: text-embedding-3-small (TE3-S) and text-embedding-3-large (TE3-L). Multi-agent workflows were constructed with LangGraph~(v1.0.3), and all models were accessed through the OpenAI Python SDK~(v2.3.0).

\subsubsection{LangGraph Selection}
LangGraph was selected to manage multi-agentic workflows due to its modular design, strong community support, and active maintenance. As a widely adopted framework, it provides a stable and extensible foundation for orchestrating complex agent interactions with precise control over information flow. Its high degree of customizability enables flexible integration with diverse tools, APIs, and models, making it adaptable to various use cases. Additionally, its emphasis on maintainability and observability supports both rapid prototyping during development and scalable deployment in production. These factors make LangGraph a suitable choice for orchestrating complex MAS financial reasoning pipelines while ensuring long-term sustainability in both research and applied settings.

\subsubsection{Prompt Selection in Agentic Workflow}
We fixed the prompt version to \(P_1\) for both chunk- and document-ranking tasks in all agentic workflow experiments to ensure controlled and interpretable results. This allowed us to isolate the effects of architectural modifications, ensuring that any performance differences could be attributed to changes in agent coordination rather than confounded by prompt variation. Furthermore, system prompts in a multi-agent setup serve distinct roles across agents, and later prompt versions contain more complex instructions that do not generalize well across heterogeneous agent responsibilities. For example, in Run~32, performance deteriorated when applying the \(P_4\) prompt to the document-ranking task, as the increased reasoning complexity introduced confusion among agents. Similarly, attempting to use \(P_4\) for chunk-ranking tasks in the hybrid workflow revealed practical constraints as the expanded prompt frequently exceeded token limits due to the large chunk sizes. For these reasons, we maintained a consistent and lightweight prompt configuration while experimenting with different agentic architectures, ensuring comparability, scalability, and clarity in performance attribution. This setup introduces a prompt asymmetry, since most agentic experiments use \(P_1\) while the best non-agentic result uses \(P_4\). That asymmetry is why the paper's multi-agent conclusion does not rest on those runs: Runs~33 and~34 apply \(P_4\) under \(A_4\) and are matched to non-agentic controls on prompt, model, and exemplars, so the comparison in Section~\ref{sec:mas_matched} isolates architecture. Run~32, which is \(P_4\) on document and \(P_1\) on chunk, is prompt-mismatched and is excluded from that analysis.

\subsection{Quantitative Results}

\subsubsection{FinAgentBench}
\label{sec:quanti_finagent_appendix}
\begin{table}[!ht]
    \footnotesize
    \centering
    \caption{FinAgentBench leaderboard results.}
    \label{tab:kaggle_lb}
    \resizebox{\columnwidth}{!}{%
    \centering
    \begin{tabular}{clccc}
    \toprule
    \textbf{Rank} & \textbf{Team} & \textbf{NDCG@5 Score (Private)} & \textbf{Code Released} & \textbf{Fine-tuning} \\
    \midrule
        1 & memex & 0.72497 & N/A & Yes \\
        2 & Yuzhen Hu & 0.72328 & N/A & Yes \\
        \rowcolor{ailens_purple!50}3 & Ours & 0.71181\textsuperscript{$\dagger$} & On acceptance & No \\
        4 & rlawnsxo & 0.70601 & N/A & N/A \\
        5 & currio & 0.70195 & N/A & N/A \\
        6 & kaggleslasher & 0.69985 & N/A & N/A \\
        7 & Gautam Jajoo & 0.69774 & N/A & N/A \\
        8 & RagTag & 0.69173 & N/A & N/A \\
        9 & ART YOUNG & 0.69162 & N/A & N/A \\
        10 & liangjiang chen & 0.69081 & N/A & N/A \\
    \bottomrule
    \end{tabular}}
    \par\noindent{\footnotesize \textsuperscript{$\dagger$}Score from the submission selected during the competition. Our best observed score across repeated runs of the same configuration (Run~19) is 0.71818 (Appendix~\ref{sec:reproducibility_details}).}
\end{table}

\noindent\textbf{Full per-run grid.}
Table~\ref{tab:ablation} is the complete configuration grid, moved here from the main text so that the main narrative carries only the decision-relevant comparisons. All NDCG@5, MAP@5, and MRR@5 values in it are on the re-annotated basis (Section~\ref{sec:bases}); leaderboard-basis scores for the subset of configurations that have them are in Table~\ref{tab:annotation_validation}. Runs~37--40 are post-competition runs and exist on the re-annotated basis only.

\begin{table*}[ht]
\footnotesize
\centering
\caption{Ablation studies of document and chunk ranking configurations across different workflows, scored against our re-annotated evaluation set. Run~19 is shown in bold because it is the configuration submitted during the competition, not because it is the highest-scoring run here.}
\label{tab:ablation}
\begin{adjustbox}{width=\textwidth}
\begin{tabular}{@{}c c c c c c c c c c c c@{}}
\toprule
\multirow{2}{*}{\textbf{ID}} & \multicolumn{3}{c}{\textbf{Document Ranking Configuration}} & \multicolumn{3}{c}{\textbf{Chunk Ranking Configuration}} & \multicolumn{2}{c}{\textbf{Agentic Configuration}} & \multicolumn{3}{c}{\textbf{Total Scores}} \\
\cmidrule(lr){2-4} \cmidrule(lr){5-7} \cmidrule(lr){8-9} \cmidrule(lr){10-12}
 & \textbf{Prompt} & \textbf{Model} & \textbf{ICL/Embedding} & \textbf{Prompt} & \textbf{Model} & \textbf{ICL/Embedding} & \textbf{Doc. Agent} & \textbf{Chunk Agent} & \textbf{NDCG@5} & \textbf{MAP@5} & \textbf{MRR@5} \\
\midrule
\multicolumn{12}{c}{\cellcolor{ailens_purple!35}\textbf{Non-Agentic Workflow}} \\
\midrule
1 & \(P_1\) & GPT-4.1 & N/A & \(P_1\) & GPT-4.1 & N/A & \multicolumn{2}{c}{N/A} & 0.78163 & 0.70815 & 0.91392 \\
2 & \(P_1\) & GPT-5-mini & N/A & \(P_1\) & GPT-5-mini & N/A & \multicolumn{2}{c}{N/A} & 0.75758 & 0.69190 & 0.90284 \\
3 & \(P_1\) & GPT-5-mini & N/A & \(P_1\) & GPT-4.1 & N/A & \multicolumn{2}{c}{N/A} & 0.77879 & 0.70778 & 0.91517 \\
4 & \(P_1\) & GPT-4.1 & N/A & \(P_1\) & GPT-5-mini & N/A & \multicolumn{2}{c}{N/A} & 0.76043 & 0.69228 & 0.90158 \\
5 & \(P_1\) & GPT-4.1 & N/A & \(P_2\) & GPT-5-mini & N/A & \multicolumn{2}{c}{N/A} & 0.77001 & 0.69818 & 0.91263 \\
6 & \(P_1\) & GPT-5-mini & N/A & \(P_3\) & GPT-5-mini & N/A & \multicolumn{2}{c}{N/A} & 0.76337 & 0.69812 & 0.90734 \\ \hline
7 & \(P_2\) & GPT-5-mini & N/A & \(P_1\) & GPT-4.1 & N/A & \multicolumn{2}{c}{N/A} & 0.77811 & 0.70686 & 0.91641 \\
8 & \(P_2\) & GPT-5-mini & N/A & \(P_1\) & GPT-5-mini & N/A & \multicolumn{2}{c}{N/A} & 0.76716 & 0.69781 & 0.91387 \\
9 & \(P_2\) & GPT-5-mini & N/A & \(P_2\) & GPT-5-mini & N/A & \multicolumn{2}{c}{N/A} & 0.76648 & 0.69688 & 0.91512 \\ \hline
10 & \(P_3\) & GPT-5-mini & N/A & \(P_1\) & GPT-4.1 & N/A & \multicolumn{2}{c}{N/A} & 0.78470 & 0.70724 & 0.91641 \\
11 & \(P_3\) & GPT-5-mini & N/A & \(P_1\) & GPT-5-mini & N/A & \multicolumn{2}{c}{N/A} & 0.76349 & 0.69137 & 0.90409 \\
12 & \(P_3\) & GPT-5-mini & N/A & \(P_3\) & GPT-5-mini & N/A & \multicolumn{2}{c}{N/A} & 0.76287 & 0.69142 & 0.91079 \\ \hline
13 & \(P_4\) & GPT-5-mini & N/A & \(P_4\) & GPT-5-mini & N/A & \multicolumn{2}{c}{N/A} & 0.77668 & 0.71053 & 0.92225 \\
14 & \(P_4\) & GPT-5-mini & N/A & \(P_4\) & GPT-5 & N/A & \multicolumn{2}{c}{N/A} & 0.77607 & 0.72303 & 0.91754 \\
15 & \(P_4\) & GPT-5 & N/A & \(P_4\) & GPT-5 & N/A & \multicolumn{2}{c}{N/A} & 0.77758 & 0.72024 & 0.91566 \\ \hline
16 & \(P_4\) & GPT-5-mini & ICL-5/TE3-S & \(P_4\) & GPT-5-mini & N/A & \multicolumn{2}{c}{N/A} & 0.76846 & 0.70326 & 0.91063 \\
17 & \(P_4\) & GPT-5-mini & ICL-5/TE3-S & \(P_4\) & GPT-5-mini & ICL-5/TE3-S & \multicolumn{2}{c}{N/A} & 0.75145 & 0.68958 & 0.89650 \\
18 & \(P_4\) & GPT-5-mini & ICL-5/TE3-S & \(P_4\) & GPT-5 & N/A & \multicolumn{2}{c}{N/A} & 0.78421 & 0.72223 & 0.91529 \\
\textbf{19} & \textbf{\(P_4\)} & \textbf{GPT-5} & \textbf{ICL-5/TE3-S} & \textbf{\(P_4\)} & \textbf{GPT-5} & \textbf{N/A} & \multicolumn{2}{c}{\textbf{N/A}} & \textbf{0.77468} & \textbf{0.71981} & \textbf{0.91629} \\
\midrule
\multicolumn{12}{c}{\cellcolor{ailens_purple!35}\textbf{Agentic Workflow}} \\
\midrule
20 & N/A & GPT-4o-mini & N/A & N/A & GPT-4o-mini & N/A & \multicolumn{2}{c}{\(A_1\)} & 0.63443 & 0.60910 & 0.77267 \\ \hline
21 & \(P_1\) & GPT-4o-mini & N/A & \(P_1\) & GPT-4o-mini & N/A & \multicolumn{2}{c}{\(A_1\)} & 0.64106 & 0.61049 & 0.78100 \\
22 & \(P_1\) & GPT-4o-mini & N/A & \(P_1\) & GPT-4o-mini & N/A & \multicolumn{2}{c}{\(A_2\)} & 0.51994 & 0.55245 & 0.71429 \\
23 & \(P_1\) & GPT-5-mini & N/A & \(P_1\) & GPT-5-mini & N/A & \multicolumn{2}{c}{\(A_2\)} & 0.60836 & 0.58235 & 0.71917 \\
24 & \(P_1\) & GPT-4o-mini & N/A & \(P_1\) & GPT-4o-mini & N/A & \multicolumn{2}{c}{\(A_3\)} & 0.52031 & 0.55441 & 0.73021 \\
25 & \(P_1\) & GPT-5-mini & N/A & \(P_1\) & GPT-5-mini & N/A & \multicolumn{2}{c}{\(A_3\)} & 0.72352 & 0.67801 & 0.88679 \\
26 & \(P_1\) & GPT-4o-mini & N/A & \(P_1\) & GPT-4o-mini & N/A & \multicolumn{2}{c}{\(A_4\)} & 0.49440 & 0.52616 & 0.63596 \\
27 & \(P_1\) & GPT-5-mini & N/A & \(P_1\) & GPT-5-mini & N/A & \multicolumn{2}{c}{\(A_4\)} & 0.70395 & 0.65719 & 0.85388 \\ \hline
28 & \(P_1\) & GPT-5-mini & ICL-5/TE3-S & \(P_1\) & GPT-5-mini & ICL-5/TE3-S & \multicolumn{2}{c}{\(A_4\)} & 0.71505 & 0.66283 & 0.86816 \\
29 & \(P_1\) & GPT-5-mini & ICL-10/TE3-S & \(P_1\) & GPT-5-mini & ICL-10/TE3-S & \multicolumn{2}{c}{\(A_4\)} & 0.68954 & 0.64463 & 0.82759 \\
30 & \(P_1\) & GPT-5-mini & ICL-10/TE3-L & \(P_1\) & GPT-5-mini & ICL-10/TE3-L & \multicolumn{2}{c}{\(A_4\)} & 0.68198 & 0.63658 & 0.83562 \\
31 & \(P_1\) & GPT-5-mini & ICL-15/TE3-L & \(P_1\) & GPT-5-mini & ICL-15/TE3-L & \multicolumn{2}{c}{\(A_4\)} & 0.65400 & 0.61208 & 0.78917 \\ \hline
32 & \(P_4\) & GPT-5-mini & ICL-10/TE3-S & \(P_1\) & GPT-5-mini & ICL-10/TE3-S & \multicolumn{2}{c}{\(A_4\)} & 0.54229 & 0.52599 & 0.61554 \\
33 & \(P_4\) & GPT-5-mini & ICL-5/TE3-S & \(P_4\) & GPT-5-mini & N/A & \multicolumn{2}{c}{\(A_4\)} & 0.69861 & 0.64794 & 0.84533 \\
34 & \(P_4\) & GPT-5-mini & ICL-5/TE3-S & \(P_4\) & GPT-5-mini & ICL-5/TE3-S & \multicolumn{2}{c}{\(A_4\)} & 0.68368 & 0.63129 & 0.81529 \\
\midrule
\multicolumn{12}{c}{\cellcolor{ailens_purple!35}\textbf{Hybrid Workflow}} \\
\midrule
35 & \(P_4\) & GPT-5-mini & N/A & \(P_4\) & GPT-5 & N/A & \(A_3\) & N/A & 0.77104 & 0.71613 & 0.91504 \\
36 & \(P_4\) & GPT-5-mini & N/A & \(P_4\) & GPT-5 & N/A & \(A_4\) & N/A & 0.78157 & 0.72115 & 0.91754 \\
\midrule
\multicolumn{12}{c}{\cellcolor{ailens_purple!35}\textbf{Open-Source Models}} \\
\midrule
37 & \(P_4\) & DeepSeek-V4-Flash & ICL-5/TE3-S & \(P_4\) & DeepSeek-V4-Flash & N/A & \multicolumn{2}{c}{N/A} & 0.80157 & 0.74708 & 0.94612 \\
38 & \(P_4\) & gpt-oss-120b & ICL-5/TE3-S & \(P_4\) & gpt-oss-120b & N/A & \multicolumn{2}{c}{N/A} & 0.65819 & 0.63978 & 0.83319 \\
39 & \(P_4\) & grok-4-20-reasoning & ICL-5/TE3-S & \(P_4\) & grok-4-20-reasoning & N/A & \multicolumn{2}{c}{N/A} & 0.83721 & 0.77711 & 0.97604 \\
40 & \(P_4\) & Llama-4-Maverick-17B-128E-Instruct-FP8 & ICL-5/TE3-S & \(P_4\) & Llama-4-Maverick-17B-128E-Instruct-FP8 & N/A & \multicolumn{2}{c}{N/A} & 0.76274 & 0.71905 & 0.92175 \\
\bottomrule
\end{tabular}
\end{adjustbox}
\end{table*}

Table~\ref{tab:kaggle_lb} reports the final private-subset leaderboard rankings. The top three methods show a small performance gap relative to the remaining teams. The memex team's solution employed a pool-of-experts of five different LLMs for pre-filtering and 65 agents for chunk retrieval and reranking, with an embedding model fine-tuned for 30 epochs on FinAgentBench training data. The second-ranked Yuzhen Hu team's solution used three MiniLM models fine-tuned separately for document retrieval, chunk retrieval, and cross-encoder reranking. In contrast, PRISM is training-free and relies on off-the-shelf LLMs with engineered prompts and lightweight multi-agent coordination. The performance gap between PRISM and the top two fine-tuned approaches ($\Delta \approx 0.005$--$0.007$ NDCG@5 on the private subset) is modest and reflects the expected trade-off of forgoing task-specific fine-tuning.

\begin{table*}[!ht]
\footnotesize
\centering
\caption{Fine-grained document, chunk and total scores for every configuration. Run IDs correspond to Table~\ref{tab:ablation}; Run~19 is bolded as the competition-submitted configuration.}
\label{tab:ablation_fine}
\begin{adjustbox}{width=\textwidth}
\begin{tabular}{@{}c c c c c c c c c c@{}}
\toprule
\multirow{2}{*}{\textbf{ID}} & \multicolumn{3}{c}{\textbf{NDCG@5}} & \multicolumn{3}{c}{\textbf{MAP@5}} & \multicolumn{3}{c}{\textbf{MRR@5}} \\
\cmidrule(lr){2-4} \cmidrule(lr){5-7} \cmidrule(lr){8-10}
 & \textbf{Doc.} & \textbf{Chunk} & \textbf{Total} & \textbf{Doc.} & \textbf{Chunk} & \textbf{Total} & \textbf{Doc.} & \textbf{Chunk} & \textbf{Total} \\
\midrule
\multicolumn{10}{c}{\cellcolor{ailens_purple!35}\textbf{Non-Agentic Workflow}} \\
\midrule
1 & 0.93857 & 0.62470 & 0.78163 & 0.96154 & 0.45477 & 0.70815 & 0.99500 & 0.83283 & 0.91392 \\
2 & 0.93288 & 0.58229 & 0.75758 & 0.96079 & 0.42302 & 0.69190 & 0.99750 & 0.80817 & 0.90284 \\
3 & 0.93288 & 0.62470 & 0.77879 & 0.96079 & 0.45477 & 0.70778 & 0.99750 & 0.83283 & 0.91517 \\
4 & 0.93857 & 0.58229 & 0.76043 & 0.96154 & 0.42302 & 0.69228 & 0.99500 & 0.80817 & 0.90158 \\
5 & 0.93857 & 0.60145 & 0.77001 & 0.96154 & 0.43483 & 0.69818 & 0.99500 & 0.83025 & 0.91263 \\
6 & 0.93288 & 0.59386 & 0.76337 & 0.96079 & 0.43544 & 0.69812 & 0.99750 & 0.81717 & 0.90734 \\ \hline
7 & 0.93152 & 0.62470 & 0.77811 & 0.95894 & 0.45477 & 0.70686 & 1.00000 & 0.83283 & 0.91641 \\
8 & 0.93288 & 0.60145 & 0.76716 & 0.96079 & 0.43483 & 0.69781 & 0.99750 & 0.83025 & 0.91387 \\
9 & 0.93152 & 0.60145 & 0.76648 & 0.95894 & 0.43483 & 0.69688 & 1.00000 & 0.83025 & 0.91512 \\ \hline
10 & 0.94469 & 0.62470 & 0.78470 & 0.95971 & 0.45477 & 0.70724 & 1.00000 & 0.83283 & 0.91641 \\
11 & 0.94469 & 0.58229 & 0.76349 & 0.95971 & 0.42302 & 0.69137 & 1.00000 & 0.80817 & 0.90409 \\
12 & 0.94800 & 0.57774 & 0.76287 & 0.96077 & 0.42206 & 0.69142 & 1.00000 & 0.82158 & 0.91079 \\ \hline
13 & 0.94955 & 0.60382 & 0.77668 & 0.96454 & 0.45652 & 0.71053 & 1.00000 & 0.84450 & 0.92225 \\
14 & 0.93935 & 0.61278 & 0.77607 & 0.96515 & 0.48092 & 0.72303 & 0.99750 & 0.83758 & 0.91754 \\
15 & 0.95227 & 0.60290 & 0.77758 & 0.96348 & 0.47700 & 0.72024 & 1.00000 & 0.83133 & 0.91566 \\ \hline
16 & 0.94594 & 0.59099 & 0.76846 & 0.96492 & 0.44160 & 0.70326 & 0.99500 & 0.82625 & 0.91063 \\
17 & 0.94950 & 0.55340 & 0.75145 & 0.96581 & 0.41335 & 0.68958 & 1.00000 & 0.79300 & 0.89650 \\
18 & 0.94891 & 0.61952 & 0.78421 & 0.96569 & 0.47877 & 0.72223 & 0.99750 & 0.83308 & 0.91529 \\
\textbf{19} & \textbf{0.93658} & \textbf{0.61278} & \textbf{0.77468} & \textbf{0.95869} & \textbf{0.48092} & \textbf{0.71981} & \textbf{0.99500} & \textbf{0.83758} & \textbf{0.91629} \\
\midrule
\multicolumn{10}{c}{\cellcolor{ailens_purple!35}\textbf{Agentic Workflow}} \\
\midrule
20 & 0.92290 & 0.34595 & 0.63443 & 0.96154 & 0.25665 & 0.60910 & 0.98500 & 0.56033 & 0.77267 \\ \hline
21 & 0.92794 & 0.35419 & 0.64106 & 0.96075 & 0.26023 & 0.61049 & 0.98750 & 0.57450 & 0.78100 \\
22 & 0.56826 & 0.47162 & 0.51994 & 0.76221 & 0.34269 & 0.55245 & 0.74000 & 0.68858 & 0.71429 \\
23 & 0.94440 & 0.27232 & 0.60836 & 0.96106 & 0.20363 & 0.58235 & 0.99750 & 0.44083 & 0.71917 \\
24 & 0.56614 & 0.47448 & 0.52031 & 0.76256 & 0.34625 & 0.55441 & 0.74000 & 0.72042 & 0.73021 \\
25 & 0.94522 & 0.50183 & 0.72352 & 0.96075 & 0.39527 & 0.67801 & 0.99750 & 0.77608 & 0.88679 \\
26 & 0.69962 & 0.28918 & 0.49440 & 0.84046 & 0.21185 & 0.52616 & 0.85000 & 0.42192 & 0.63596 \\
27 & 0.94464 & 0.46326 & 0.70395 & 0.96508 & 0.34931 & 0.65719 & 0.99750 & 0.71025 & 0.85388 \\ \hline
28 & 0.93987 & 0.49024 & 0.71505 & 0.96779 & 0.35787 & 0.66283 & 0.99500 & 0.74133 & 0.86816 \\
29 & 0.94144 & 0.43765 & 0.68954 & 0.96948 & 0.31979 & 0.64463 & 0.99500 & 0.66017 & 0.82759 \\
30 & 0.94710 & 0.41686 & 0.68198 & 0.96806 & 0.30510 & 0.63658 & 1.00000 & 0.67125 & 0.83562 \\
31 & 0.94685 & 0.36115 & 0.65400 & 0.96771 & 0.25644 & 0.61208 & 0.99500 & 0.58333 & 0.78917 \\ \hline
32 & 0.95272 & 0.13185 & 0.54229 & 0.96381 & 0.08817 & 0.52599 & 0.99500 & 0.23608 & 0.61554 \\
33 & 0.95359 & 0.44364 & 0.69861 & 0.96808 & 0.32779 & 0.64794 & 0.99750 & 0.69317 & 0.84533 \\
34 & 0.95259 & 0.41476 & 0.68368 & 0.96783 & 0.29475 & 0.63129 & 0.99750 & 0.63308 & 0.81529 \\
\midrule
\multicolumn{10}{c}{\cellcolor{ailens_purple!35}\textbf{Hybrid Workflow}} \\
\midrule
35 & 0.92929 & 0.61278 & 0.77104 & 0.95133 & 0.48092 & 0.71613 & 0.99250 & 0.83758 & 0.91504 \\
36 & 0.95036 & 0.61278 & 0.78157 & 0.96137 & 0.48092 & 0.72115 & 0.99750 & 0.83758 & 0.91754 \\
\midrule
\multicolumn{10}{c}{\cellcolor{ailens_purple!35}\textbf{Open-Source Models}} \\
\midrule
37 & 0.93376 & 0.66938 & 0.80157 & 0.96665 & 0.52750 & 0.74708 & 0.99750 & 0.89475 & 0.94612 \\
38 & 0.92732 & 0.38906 & 0.65819 & 0.95109 & 0.32847 & 0.63978 & 0.98770 & 0.67868 & 0.83319 \\
39 & 0.92620 & 0.74823 & 0.83721 & 0.94944 & 0.60477 & 0.77711 & 0.99250 & 0.95958 & 0.97604 \\
40 & 0.95573 & 0.56974 & 0.76274 & 0.96969 & 0.46842 & 0.71905 & 1.00000 & 0.84350 & 0.92175 \\
\bottomrule
\end{tabular}
\end{adjustbox}
\end{table*}

\subsubsection{FiQA-2018}
\label{sec:quanti_fiqa}

\begin{table}[!ht]
    \footnotesize
    \centering
    \caption{FiQA evaluation results.}
    \label{tab:fiqa_results}
    \resizebox{\columnwidth}{!}{%
        \begin{tabular}{clccc}
        \toprule
        \textbf{Prompt} & \textbf{Model} & \textbf{ICL} & \textbf{NDCG@10} & \textbf{Recall@100} \\
        \midrule
        N/A & BM25+CE & N/A & 0.3470 & 0.5390 \\
        N/A & TE3-S & N/A & 0.5499 & 0.7783 \\
        N/A & BGE-M3 (dense) & N/A & 0.5297 & 0.7088 \\
        N/A & BGE-M3 (sparse) & N/A & 0.3692 & 0.5133 \\
        N/A & SPLADE-v3 & N/A & 0.4928 & 0.6683 \\
        \midrule
        \multicolumn{5}{c}{\cellcolor{ailens_purple!35}\textbf{Closed-Source Models}} \\
        \midrule
        \(P_1\) & GPT-5 & ICL-5 & 0.6104 & 0.8083 \\
        \(P_2\) & GPT-5 & ICL-5 & 0.6027 & 0.7941 \\
        \(P_3\) & GPT-5 & ICL-5 & 0.6197 & 0.8206 \\
        \(P_4\) & GPT-5 & ICL-5 & 0.6097 & 0.8021 \\
        \midrule
        \multicolumn{5}{c}{\cellcolor{ailens_purple!35}\textbf{Open-Source Models}} \\
        \midrule
        \(P_4\) & DeepSeek-V4-Flash & ICL-5 & 0.5778 & 0.7905 \\
        \(P_4\) & gpt-oss-120b & ICL-5 & 0.5833 & 0.6758 \\
        \(P_4\) & grok-4-20-reasoning & ICL-5 & 0.5741 & 0.7881 \\
        \(P_4\) & Llama-4-Maverick-17B-128E-Instruct-FP8 & ICL-5 & 0.5581 & 0.7726 \\
        \bottomrule
        \end{tabular}%
    }
\end{table}

Table~\ref{tab:fiqa_baselines} reports the traditional baselines, and Table~\ref{tab:full_fiqa_results} provides the full FiQA model-by-prompt breakdown, including zero-shot and ICL results.

Examining the prompt design progression reveals nuanced performance patterns. $P_1$, which employs a straightforward prompting strategy, establishes a strong baseline with GPT-5 achieving 0.6130 NDCG@10 and 0.8045 Recall@100. $P_2$, which integrates key contextual information, demonstrates comparable performance of 0.6095 NDCG@10 and 0.8094 Recall@100 with GPT-5, suggesting that additional context maintains effectiveness while potentially offering enhanced interpretability. $P_3$, which utilizes Chain-of-Thought (CoT) reasoning, achieves the highest NDCG@10 of 0.6197, indicating that structured reasoning particularly benefits ranking precision at top positions. Interestingly, $P_4$, which refines $P_3$ with streamlined instructions and stricter output formatting while maintaining the combined CoT, ReAct, and ToT strategies, shows slightly lower performance of 0.6070 NDCG@10, suggesting that its more constrained output structure is less well aligned with FiQA's passage reranking format than $P_3$'s more expressive reasoning scaffold.

The impact of ICL varies significantly by model capability, revealing important insights for practical deployment. ICL proves particularly beneficial for models with more limited reasoning capabilities, as evidenced by GPT-4.1's consistent performance gains across all prompts (e.g., 23.5\% improvement in NDCG@10 for $P_1$). This occurs because ICL provides concrete demonstrations of task structure and expected reasoning patterns, effectively compensating for limited inherent reasoning capacity. In contrast, GPT-5 shows only marginal variation with ICL (ranging from small declines to small gains depending on prompt), while GPT-5-mini is consistently hurt by ICL on FiQA (e.g., $P_1$: 0.5565 to 0.5171, $P_4$: 0.5269 to 0.4615), suggesting that for capable models the in-context examples consume context budget without providing proportional guidance benefit on this passage-level task. This differential ICL sensitivity suggests that smaller or less capable models can be substantially enhanced through careful example selection, while more advanced models require task-specific validation before activating ICL.

The open-source model evaluations using $P_4$ further validate PRISM's generalizability beyond a single model family. All four models substantially outperform the BM25+CE baseline under both zero-shot and ICL-5 conditions, with NDCG@10 ranging from 0.552 to 0.583, trailing GPT-5 by a modest margin. Notably, gpt-oss-120b achieves the strongest open-source result of 0.5833 NDCG@10 under ICL-5, approaching the performance of GPT-5-mini. The effect of ICL across these models is consistent but marginal, with ICL-5 yielding small improvements over zero-shot across all four models, suggesting that they already possess sufficient instruction-following capacity to perform the reranking task without extensive demonstration examples. The relatively uniform performance across architecturally diverse models, spanning mixture-of-experts, dense, and reasoning-specialized designs indicates that PRISM's prompt-driven approach is robust to underlying model architecture.

Overall, all PRISM configurations maintain higher Recall@100, with GPT-5 results consistently near or above 0.80, demonstrating strong retrieval coverage, while improvements in NDCG@10 highlight enhanced precision in ranking highly relevant documents at top positions, a critical capability for practical information retrieval systems.

\begin{table}[htbp]
    \centering
    \caption{Traditional baseline performance on FiQA-2018.}
    \label{tab:fiqa_baselines}
    \resizebox{\columnwidth}{!}{
    \begin{tabular}{llcc}
    \toprule
    \textbf{Method} & \textbf{Model} & \textbf{NDCG@10} & \textbf{Recall@100} \\
    \midrule
    BM25       & Baseline & 0.2360 & 0.5390 \\
    DeepCT     & Sparse & 0.1910 & 0.4890 \\
    SPARTA     & Sparse & 0.1980 & 0.4460 \\
    docT5query & Sparse & 0.2910 & 0.5980 \\
    DPR        & Dense & 0.1120 & 0.3420 \\
    ANCE       & Dense &0.2950 & 0.5810 \\
    TAS-B      & Dense &0.3000 & 0.5930 \\
    GenQ       & Dense &0.3080 & 0.6180 \\
    ColBERT    & Late-Interaction & 0.3170 & 0.6030 \\
    BM25+CE    & Re-ranking & 0.3470 & 0.5390 \\
    \bottomrule
    \end{tabular}}
\end{table}

\begin{table}[t]
\footnotesize
\centering
\caption{PRISM performance on FiQA-2018.}
\label{tab:full_fiqa_results}
\resizebox{\columnwidth}{!}{
\begin{tabular}{llccc}
\toprule
\textbf{Prompt} & \textbf{Model} & \textbf{ICL} & \textbf{NDCG@10} & \textbf{Recall@100} \\
\midrule
\multicolumn{5}{c}{\cellcolor{ailens_purple!35}\textbf{Closed-Source Models}} \\
\midrule
\(P_1\) & GPT-4.1    & N/A & 0.4307 & 0.6024 \\
\(P_1\) & GPT-5-mini & N/A & 0.5565 & 0.7268 \\
\(P_1\) & GPT-5      & N/A & 0.6130 & 0.8045 \\
\(P_1\) & GPT-4.1    & ICL-5 & 0.5320 & 0.7247 \\
\(P_1\) & GPT-5-mini & ICL-5 & 0.5171 & 0.6412 \\
\(P_1\) & GPT-5      & ICL-5 & 0.6104 & 0.8083 \\
\midrule
\(P_2\) & GPT-4.1    & N/A & 0.4871 & 0.6711 \\
\(P_2\) & GPT-5-mini & N/A & 0.5521 & 0.7094 \\
\(P_2\) & GPT-5      & N/A & 0.6095 & 0.8094 \\
\(P_2\) & GPT-4.1    & ICL-5 & 0.4974 & 0.6863 \\
\(P_2\) & GPT-5-mini & ICL-5 & 0.4960 & 0.6142 \\
\(P_2\) & GPT-5      & ICL-5 & 0.6027 & 0.7941 \\
\midrule
\(P_3\) & GPT-4.1    & N/A & 0.4479 & 0.5948 \\
\(P_3\) & GPT-5-mini & N/A & 0.5642 & 0.7374 \\
\(P_3\) & GPT-5      & N/A & 0.6094 & 0.8136 \\
\(P_3\) & GPT-4.1    & ICL-5 & 0.4702 & 0.6471 \\
\(P_3\) & GPT-5-mini & ICL-5 & 0.5182 & 0.6506 \\
\(P_3\) & GPT-5      & ICL-5 & 0.6197 & 0.8206 \\
\midrule
\(P_4\) & GPT-4.1    & N/A & 0.4178 & 0.5739 \\
\(P_4\) & GPT-5-mini & N/A & 0.5269 & 0.6812 \\
\(P_4\) & GPT-5      & N/A & 0.6070 & 0.8195 \\
\(P_4\) & GPT-4.1    & ICL-5 & 0.4544 & 0.6212 \\
\(P_4\) & GPT-5-mini & ICL-5 & 0.4615 & 0.5610 \\
\(P_4\) & GPT-5      & ICL-5 & 0.6097 & 0.8021 \\
\midrule
\multicolumn{5}{c}{\cellcolor{ailens_purple!35}\textbf{Open-Source Models}} \\
\midrule
\(P_4\) & DeepSeek-V4-Flash    & N/A & 0.5696 & 0.7862 \\
\(P_4\) & gpt-oss-120b & N/A & 0.5592 & 0.6609 \\
\(P_4\) & grok-4-20-reasoning      & N/A & 0.5642 & 0.7823 \\
\(P_4\) & Llama-4-Maverick-17B-128E-Instruct-FP8      & N/A & 0.5515 & 0.7628 \\
\(P_4\) & DeepSeek-V4-Flash    & ICL-5 & 0.5778 & 0.7905 \\
\(P_4\) & gpt-oss-120b & ICL-5 & 0.5833 & 0.6758 \\
\(P_4\) & grok-4-20-reasoning      & ICL-5 & 0.5741 & 0.7881 \\
\(P_4\) & Llama-4-Maverick-17B-128E-Instruct-FP8      & ICL-5 & 0.5581 & 0.7726 \\
\bottomrule
\end{tabular}
}
\end{table}

\subsubsection{FinanceBench} 
\label{sec:quanti_financebench}
Table~\ref{tab:financebench_results} summarizes the oracle setting, and Table~\ref{tab:full_financebench_results} provides the full FinanceBench evaluation-mode breakdown and open-source model comparison.

\begin{table}[!ht]
    \footnotesize
    \centering
    \caption{FinanceBench oracle evaluation results.}
    \label{tab:financebench_results}
    \resizebox{0.9\columnwidth}{!}{%
        \begin{tabular}{clccc}
        \toprule
        \textbf{Prompt} & \textbf{Model} & \textbf{Correct} & \textbf{Incorrect} & \textbf{Unable} \\
        \midrule
        \multicolumn{5}{c}{\cellcolor{ailens_purple!35}\textbf{Closed-Source Models}} \\
        \midrule
        Baseline & GPT-5 & 141 & 9 & 0 \\
        \(P_1\) & GPT-5 & 147 & 3 & 0 \\
        \(P_2\) & GPT-5 & 142 & 8 & 0 \\
        \(P_3\) & GPT-5 & 140 & 10 & 0 \\
        \(P_4\) & GPT-5 & 146 & 4 & 0 \\
        \midrule
        \multicolumn{5}{c}{\cellcolor{ailens_purple!35}\textbf{Open-Source Models}} \\
        \midrule
        \(P_4\) & DeepSeek-V4-Flash & 125 & 25 & 0 \\
        \(P_4\) & gpt-oss-120b & 145 & 4 & 1 \\
        \(P_4\) & grok-4-20-reasoning & 143 & 7 & 0 \\
        \(P_4\) & Llama-4-Maverick-17B-128E-Instruct-FP8 & 86 & 64 & 0 \\
        \bottomrule
        \end{tabular}%
    }
\end{table}

An important observation emerges regarding prompt complexity and task alignment. While $P_4$ shows slightly lower performance in reranking tasks, it maintains competitive performance in QA settings, with GPT-5 Oracle achieving 97.33\% accuracy (146/150). This suggests that optimal prompt design is indeed task-dependent, as we observed in the reranking experiments. Simpler prompts ($P_1$) often match or exceed more complex variants, indicating that the structured reasoning scaffolding beneficial for nuanced ranking decisions may be less critical for direct factual retrieval tasks inherent to question-answering.

The limited utility of ICL in FinanceBench can be attributed to its task-specific characteristics, in contrast to its more substantial benefits in reranking tasks. FinanceBench queries require finding specific answers within individual documents rather than comparing multiple candidates, making similarity-based ICL selection less effective at retrieving truly relevant question-answer exemplars. This can produce ICL examples that misdirect rather than guide the model, explaining why non-ICL configurations often perform comparably or better. Nevertheless, despite the limited dataset size (150 questions) and the already high baseline performance of up to 96\% accuracy, PRISM achieves measurable improvements, demonstrating its effectiveness across different task types beyond reranking. While the absolute gains are necessarily constrained in such a high-performing regime approaching the theoretical ceiling of 100\%, PRISM's ability to push accuracy toward 98\% supports its versatility as a general-purpose framework for LLM-based information retrieval tasks.

The open-source model evaluations under Oracle conditions using $P_4$ provide further evidence of PRISM's generalizability. gpt-oss-120b achieves the strongest open-source result at 145/150 (96.7\%) with ICL, closely matching the best GPT-5 prompt variants and surpassing the GPT-5 oracle baseline. grok-4-20-reasoning similarly performs strongly at 143/150 (95.3\%) with ICL, confirming that PRISM's prompt-driven approach transfers effectively across model families. The ICL effect is more pronounced for these models on FinanceBench than on FiQA: grok-4-20-reasoning gains substantially from ICL (143 vs.\ 126), while gpt-oss-120b shows a modest improvement (145 vs.\ 142), suggesting that structured financial QA generally benefits more from demonstration examples than open-ended reranking does. DeepSeek-V4-Flash shows a larger ICL benefit (125 vs.\ 112), though its overall accuracy remains lower than that of the stronger open-source models. The notable exception is Llama-4-Maverick-17B-128E-Instruct-FP8, which achieves only 86/150 with ICL and 70/150 zero-shot, suggesting that this model requires additional adaptation or fine-tuning to handle the precise numerical and financial reasoning demands of FinanceBench effectively.

\onecolumn
\begin{longtable}{lp{3cm}lc ccc}
    \caption{Comprehensive evaluation results of different prompt versions and models across various evaluation modes on FinanceBench.}
    \label{tab:full_financebench_results} \\
    
    \toprule
    \multirow{2}{*}{\textbf{Prompt}} & \multirow{2}{*}{\textbf{Model}} & \multirow{2}{*}{\textbf{Eval Mode}} & \multirow{2}{*}{\textbf{ICL}} & \multicolumn{3}{c}{\textbf{Results (Out of 150)}} \\
    \cmidrule(lr){5-7}
    & & & & \textbf{Correct} & \textbf{Incorrect} & \textbf{Unable to Answer} \\
    \midrule
    \endfirsthead
    
    \multicolumn{7}{c}%
    {\tablename\ \thetable{} -- \textit{Continued from previous page}} \\
    \toprule
    \multirow{2}{*}{\textbf{Prompt}} & \multirow{2}{*}{\textbf{Model}} & \multirow{2}{*}{\textbf{Eval Mode}} & \multirow{2}{*}{\textbf{ICL}} & \multicolumn{3}{c}{\textbf{Results (Out of 150)}} \\
    \cmidrule(lr){5-7}
    & & & & \textbf{Correct} & \textbf{Incorrect} & \textbf{Unable to Answer} \\
    \midrule
    \endhead
    
    \midrule
    \multicolumn{7}{r}{\textit{Continued on next page}} \\
    \endfoot
    
    \bottomrule
    \endlastfoot

    \multicolumn{7}{c}{\cellcolor{ailens_purple!35}\textbf{Baselines}} \\
    \midrule
    Baseline & GPT-4.1 & Single Store & N/A & 132 & 18 & 0 \\
    Baseline & GPT-4.1 & Shared Store & N/A & 124 & 26 & 0 \\
    Baseline & GPT-4.1 & In Context   & N/A & 129 & 21 & 0 \\
    Baseline & GPT-4.1 & Oracle       & N/A & 133 & 16 & 1 \\
    Baseline & GPT-5-mini & Single Store & N/A & 144 & 4 & 2 \\
    Baseline & GPT-5-mini & Shared Store & N/A & 143 & 7 & 0 \\
    Baseline & GPT-5-mini & In Context   & N/A & 142 & 7 & 1 \\
    Baseline & GPT-5-mini & Oracle       & N/A & 145 & 5 & 0 \\
    Baseline & GPT-5   & Single Store & N/A & 141 & 8 & 1 \\
    Baseline & GPT-5   & Shared Store & N/A & 143 & 7 & 0 \\
    Baseline & GPT-5   & In Context   & N/A & 139 & 11 & 0 \\
    Baseline & GPT-5   & Oracle       & N/A & 141 & 9 & 0 \\
    \midrule
    \multicolumn{7}{c}{\cellcolor{ailens_purple!35}\textbf{PRISM}} \\
    \midrule
    \(P_1\) & GPT-4.1 & Single Store & ICL-9 & 124 & 26 & 0 \\
    \(P_1\) & GPT-4.1 & Shared Store & ICL-9 & 126 & 24 & 0 \\
    \(P_1\) & GPT-4.1 & In Context   & ICL-9 & 124 & 26 & 0 \\
    \(P_1\) & GPT-4.1 & Oracle       & ICL-9 & 131 & 19 & 0 \\
    \(P_1\) & GPT-5-mini & Single Store & ICL-9 & 139 & 11 & 0 \\
    \(P_1\) & GPT-5-mini & Shared Store & ICL-9 & 144 & 6 & 0 \\
    \(P_1\) & GPT-5-mini & In Context   & ICL-9 & 140 & 10 & 0 \\
    \(P_1\) & GPT-5-mini & Oracle       & ICL-9 & 143 & 7 & 0 \\
    \(P_1\) & GPT-5   & Single Store & ICL-9 & 146 & 4 & 0 \\
    \(P_1\) & GPT-5   & Shared Store & ICL-9 & 145 & 5 & 0 \\
    \(P_1\) & GPT-5   & In Context   & ICL-9 & 136 & 14 & 0 \\
    \(P_1\) & GPT-5   & Oracle       & ICL-9 & 147 & 3 & 0 \\
    \midrule
    \(P_1\) & GPT-4.1 & Single Store & N/A & 132 & 18 & 0 \\
    \(P_1\) & GPT-4.1 & Shared Store & N/A & 122 & 27 & 1 \\
    \(P_1\) & GPT-4.1 & In Context   & N/A & 133 & 17 & 0 \\
    \(P_1\) & GPT-4.1 & Oracle       & N/A & 129 & 21 & 0 \\
    \(P_1\) & GPT-5-mini & Single Store & N/A & 142 & 8 & 0 \\
    \(P_1\) & GPT-5-mini & Shared Store & N/A & 147 & 3 & 0 \\
    \(P_1\) & GPT-5-mini & In Context   & N/A & 143 & 7 & 0 \\
    \(P_1\) & GPT-5-mini & Oracle       & N/A & 141 & 9 & 0 \\
    \(P_1\) & GPT-5   & Single Store & N/A & 140 & 10 & 0 \\
    \(P_1\) & GPT-5   & Shared Store & N/A & 142 & 8 & 0 \\
    \(P_1\) & GPT-5   & In Context   & N/A & 136 & 14 & 0 \\
    \(P_1\) & GPT-5   & Oracle       & N/A & 140 & 10 & 0 \\
    \midrule
    \(P_2\) & GPT-4.1 & Single Store & ICL-9 & 122 & 28 & 0 \\
    \(P_2\) & GPT-4.1 & Shared Store & ICL-9 & 114 & 35 & 1 \\
    \(P_2\) & GPT-4.1 & In Context   & ICL-9 & 126 & 24 & 0 \\
    \(P_2\) & GPT-4.1 & Oracle       & ICL-9 & 132 & 18 & 0 \\
    \(P_2\) & GPT-5-mini & Single Store & ICL-9 & 145 & 5 & 0 \\
    \(P_2\) & GPT-5-mini & Shared Store & ICL-9 & 133 & 16 & 0 \\
    \(P_2\) & GPT-5-mini & In Context   & ICL-9 & 146 & 4 & 0 \\
    \(P_2\) & GPT-5-mini & Oracle       & ICL-9 & 142 & 7 & 1 \\
    \(P_2\) & GPT-5   & Single Store & ICL-9 & 139 & 11 & 0 \\
    \(P_2\) & GPT-5   & Shared Store & ICL-9 & 136 & 12 & 2 \\
    \(P_2\) & GPT-5   & In Context   & ICL-9 & 138 & 12 & 0 \\
    \(P_2\) & GPT-5   & Oracle       & ICL-9 & 142 & 8 & 0 \\
    \midrule
    \(P_2\) & GPT-4.1 & Single Store & N/A & 130 & 20 & 0 \\
    \(P_2\) & GPT-4.1 & Shared Store & N/A & 126 & 23 & 1 \\
    \(P_2\) & GPT-4.1 & In Context   & N/A & 126 & 24 & 0 \\
    \(P_2\) & GPT-4.1 & Oracle       & N/A & 140 & 10 & 0 \\
    \(P_2\) & GPT-5-mini & Single Store & N/A & 145 & 5 & 0 \\
    \(P_2\) & GPT-5-mini & Shared Store & N/A & 141 & 9 & 0 \\
    \(P_2\) & GPT-5-mini & In Context   & N/A & 140 & 10 & 0 \\
    \(P_2\) & GPT-5-mini & Oracle       & N/A & 141 & 9 & 0 \\
    \(P_2\) & GPT-5   & Single Store & N/A & 134 & 15 & 1 \\
    \(P_2\) & GPT-5   & Shared Store & N/A & 144 & 6 & 0 \\
    \(P_2\) & GPT-5   & In Context   & N/A & 140 & 10 & 0 \\
    \(P_2\) & GPT-5   & Oracle       & N/A & 143 & 6 & 1 \\
    \midrule
    \(P_3\) & GPT-4.1 & Single Store & ICL-9 & 128 & 22 & 0 \\
    \(P_3\) & GPT-4.1 & Shared Store & ICL-9 & 120 & 30 & 0 \\
    \(P_3\) & GPT-4.1 & In Context   & ICL-9 & 129 & 21 & 0 \\
    \(P_3\) & GPT-4.1 & Oracle       & ICL-9 & 129 & 21 & 0 \\
    \(P_3\) & GPT-5-mini & Single Store & ICL-9 & 141 & 9 & 0 \\
    \(P_3\) & GPT-5-mini & Shared Store & ICL-9 & 140 & 10 & 0 \\
    \(P_3\) & GPT-5-mini & In Context   & ICL-9 & 142 & 8 & 0 \\
    \(P_3\) & GPT-5-mini & Oracle       & ICL-9 & 143 & 7 & 0 \\
    \(P_3\) & GPT-5   & Single Store & ICL-9 & 146 & 3 & 1 \\
    \(P_3\) & GPT-5   & Shared Store & ICL-9 & 147 & 3 & 0 \\
    \(P_3\) & GPT-5   & In Context   & ICL-9 & 138 & 12 & 0 \\
    \(P_3\) & GPT-5   & Oracle       & ICL-9 & 140 & 10 & 0 \\
    \midrule
    \(P_3\) & GPT-4.1 & Single Store & N/A & 128 & 22 & 0 \\
    \(P_3\) & GPT-4.1 & Shared Store & N/A & 121 & 29 & 0 \\
    \(P_3\) & GPT-4.1 & In Context   & N/A & 123 & 27 & 0 \\
    \(P_3\) & GPT-4.1 & Oracle       & N/A & 136 & 14 & 0 \\
    \(P_3\) & GPT-5-mini & Single Store & N/A & 140 & 10 & 0 \\
    \(P_3\) & GPT-5-mini & Shared Store & N/A & 140 & 9 & 1 \\
    \(P_3\) & GPT-5-mini & In Context   & N/A & 144 & 6 & 0 \\
    \(P_3\) & GPT-5-mini & Oracle       & N/A & 146 & 0 & 0 \\
    \(P_3\) & GPT-5   & Single Store & N/A & 133 & 17 & 0 \\
    \(P_3\) & GPT-5   & Shared Store & N/A & 135 & 13 & 2 \\
    \(P_3\) & GPT-5   & In Context   & N/A & 133 & 17 & 0 \\
    \(P_3\) & GPT-5   & Oracle       & N/A & 139 & 11 & 0 \\
    \midrule
    \(P_4\) & GPT-4.1 & Single Store & ICL-9 & 120 & 30 & 0 \\
    \(P_4\) & GPT-4.1 & Shared Store & ICL-9 & 109 & 39 & 2 \\
    \(P_4\) & GPT-4.1 & In Context   & ICL-9 & 125 & 25 & 0 \\
    \(P_4\) & GPT-4.1 & Oracle       & ICL-9 & 128 & 22 & 0 \\
    \(P_4\) & GPT-5-mini & Single Store & ICL-9 & 138 & 12 & 0 \\
    \(P_4\) & GPT-5-mini & Shared Store & ICL-9 & 139 & 11 & 0 \\
    \(P_4\) & GPT-5-mini & In Context   & ICL-9 & 143 & 7 & 0 \\
    \(P_4\) & GPT-5-mini & Oracle       & ICL-9 & 140 & 10 & 0 \\
    \(P_4\) & GPT-5   & Single Store & ICL-9 & 141 & 9 & 0 \\
    \(P_4\) & GPT-5   & Shared Store & ICL-9 & 135 & 14 & 1 \\
    \(P_4\) & GPT-5   & In Context   & ICL-9 & 137 & 12 & 1 \\
    \(P_4\) & GPT-5   & Oracle       & ICL-9 & 146 & 4 & 0 \\
    \midrule
    \(P_4\) & GPT-4.1 & Single Store & N/A & 125 & 25 & 0 \\
    \(P_4\) & GPT-4.1 & Shared Store & N/A & 119 & 31 & 0 \\
    \(P_4\) & GPT-4.1 & In Context   & N/A & 125 & 25 & 0 \\
    \(P_4\) & GPT-4.1 & Oracle       & N/A & 133 & 17 & 0 \\
    \(P_4\) & GPT-5-mini & Single Store & N/A & 137 & 12 & 1 \\
    \(P_4\) & GPT-5-mini & Shared Store & N/A & 136 & 12 & 2 \\
    \(P_4\) & GPT-5-mini & In Context   & N/A & 138 & 12 & 0 \\
    \(P_4\) & GPT-5-mini & Oracle       & N/A & 137 & 13 & 0 \\
    \(P_4\) & GPT-5   & Single Store & N/A & 137 & 13 & 0 \\
    \(P_4\) & GPT-5   & Shared Store & N/A & 136 & 13 & 1 \\
    \(P_4\) & GPT-5   & In Context   & N/A & 137 & 13 & 0 \\
    \(P_4\) & GPT-5   & Oracle       & N/A & 142 & 8 & 0 \\
    \midrule
    \multicolumn{7}{c}{\cellcolor{ailens_purple!35}\textbf{Open-Source Models}} \\
    \midrule
    \(P_4\) & DeepSeek-V4-Flash & Oracle & ICL-9 & 125 & 25 & 0 \\
    \(P_4\) & gpt-oss-120b & Oracle & ICL-9 & 145 & 4 & 1 \\
    \(P_4\) & grok-4-20-reasoning & Oracle   & ICL-9 & 143 & 7 & 0 \\
    \(P_4\) & Llama-4-Maverick-17B-128E-Instruct-FP8      & Oracle       & ICL-9 & 86 & 64 & 0 \\
    \(P_4\) & DeepSeek-V4-Flash & Oracle & N/A & 112 & 38 & 0 \\
    \(P_4\) & gpt-oss-120b & Oracle & N/A & 142 & 7 & 1 \\
    \(P_4\) & grok-4-20-reasoning & Oracle   & N/A & 126 & 24 & 0 \\
    \(P_4\) & Llama-4-Maverick-17B-128E-Instruct-FP8      & Oracle       & N/A & 70 & 79 & 1 \\
\end{longtable}
\twocolumn

\subsection{Feasibility Analysis}
\label{sec:feasibility_appendix}
\begin{figure*}[!ht]
    \centering
    \includegraphics[width=\linewidth]{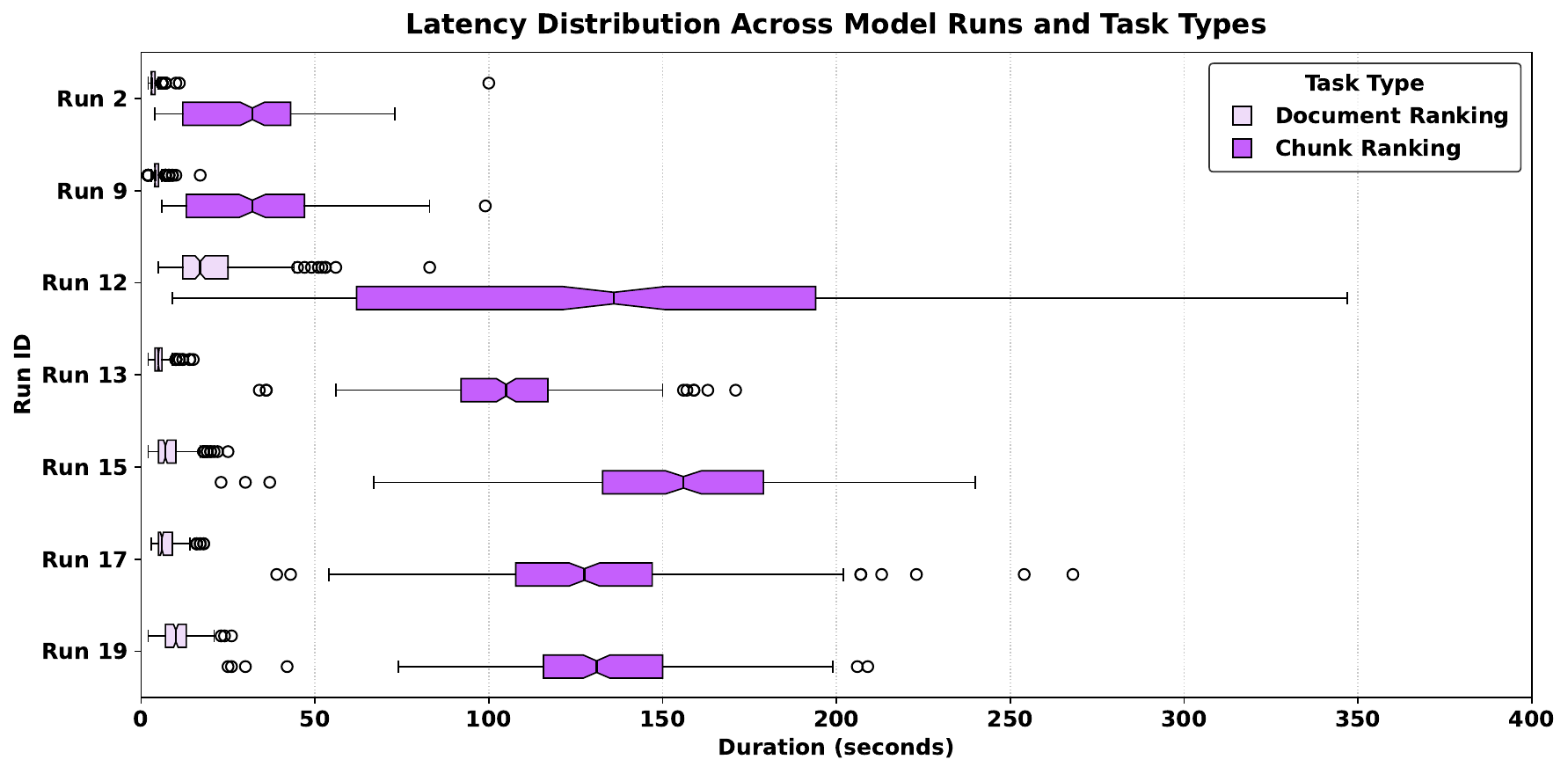}
    \caption{Latency distribution box plots, where each box shows how the latency varies across runs with different LLMs and configurations. For chunk ranking tasks, we observe that \(P_3\) (Run~12) has the highest median latency for both tasks due to its heavy reasoning prompt design, while \(P_4\) (Runs~13--19) achieves a good balance between latency and performance.}
    \label{fig:latency_box}
\end{figure*}

\begin{figure*}[!ht]
    \centering
    \includegraphics[width=\linewidth]{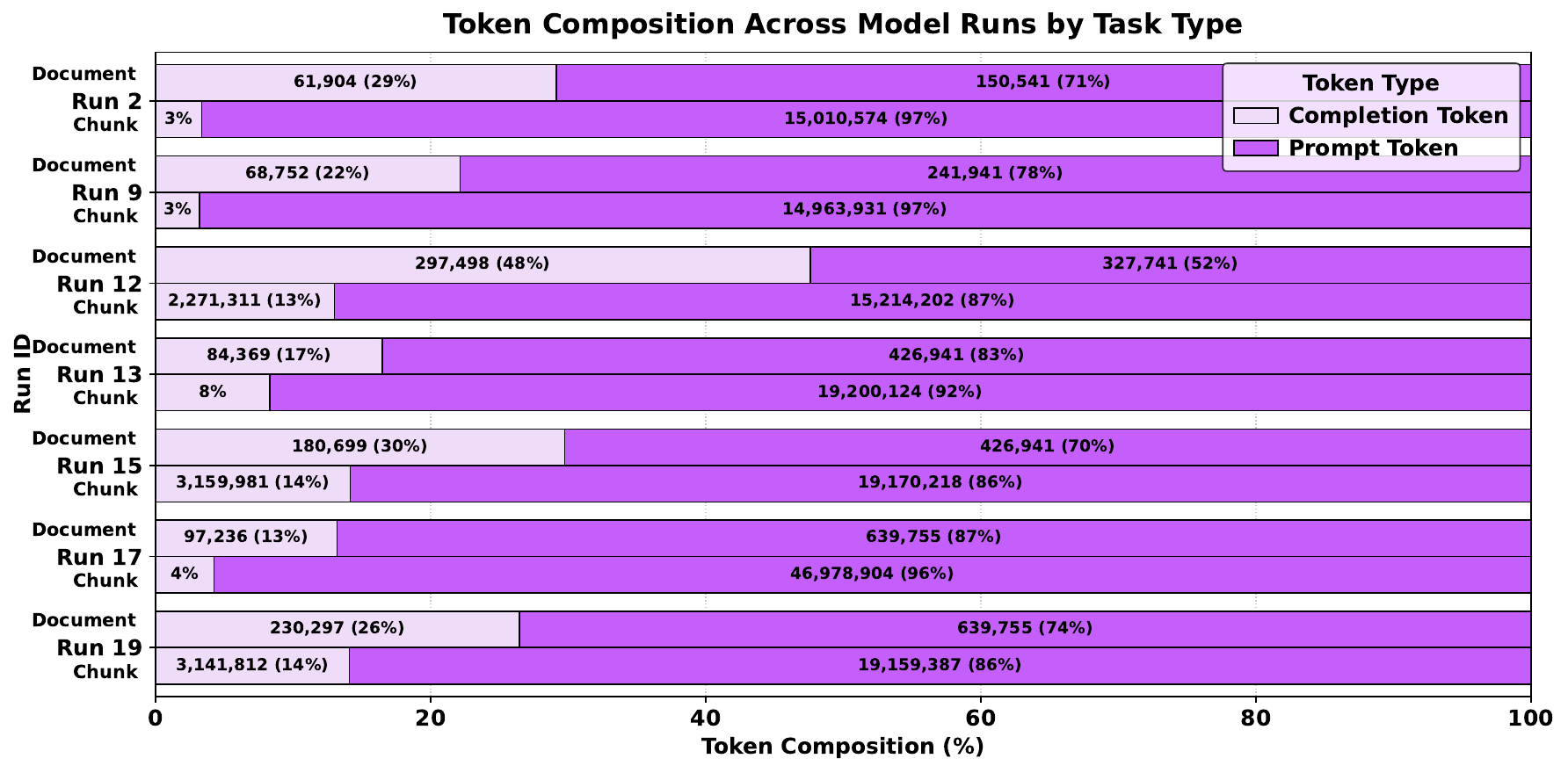}
    \caption{Token distribution bar charts, where each bar details the proportion of prompt and completion token usage across different runs and tasks, highlighting that chunk ranking tasks use more prompt tokens due to longer context and complexity.}
    \label{fig:token_box}
\end{figure*}

No prior work on FinAgentBench reports latency or cost, so these are the first such baselines. Run~19 has median latencies of 10\,s for document ranking and 131\,s for chunk ranking, at per-query token costs of \$0.0155 and \$0.2768 (per-run statistics in Appendix Table~\ref{tab:full_analysis}, distributions in Appendix Figures~\ref{fig:latency_box} and~\ref{fig:token_box}). Chunk ranking dominates resource use because of its token-heavy context, which makes candidate-pool control the effective cost lever: a Cohere-refined top-50 pool cuts chunk-ranking cost by about 35\% (\$0.278 to \$0.181 per query) with no retraining, and refinement itself is cheap, since $k$-reciprocal and the open rerankers run locally and Cohere adds under \$0.01 per query (Appendix Tables~\ref{tab:finagent_refine_cost} and~\ref{tab:fiqa_refine_cost}). For cost-sensitive deployments, GPT-5-mini configurations trade moderate NDCG@5 for much lower cost; full breakdowns are in Appendix~\ref{sec:feasibility_appendix}.

\noindent\textbf{Reproducibility.}
Repeated private-subset submissions show low stochastic variance (CV $<1.6\%$), with Run~19 the most stable high-performing configuration (mean 0.71163, SD 0.00433 over nine runs, leaderboard basis). Welch tests~\cite{welch1938significance} against Run~12 are significant for Runs~15, 18, and 19, so those gains exceed run noise. Query-level uncertainty on the re-annotated basis is quantified separately by the paired bootstrap of Appendix~\ref{sec:adjacent_significance}; full statistics for all repeated runs are in Appendix~\ref{sec:reproducibility_details}.

\subsubsection{Cost Analysis}
\begin{table*}[htbp]
\centering
\caption{Latency, token usage, and cost statistics for different runs used in feasibility analysis on FinAgentBench.}
\label{tab:full_analysis}
\begin{adjustbox}{width=\textwidth}
\begin{tabular}{@{}ccccccccccccccc@{}}
\toprule
\multirow{2}{*}{\textbf{ID}} & \multirow{2}{*}{\textbf{Task}} & \multicolumn{5}{c}{\textbf{Latency Statistics (seconds)}} & \multicolumn{4}{c}{\textbf{Token Usage Statistics}} & \multicolumn{3}{c}{\textbf{Cost Statistics (USD)}}\\
\cmidrule(lr){3-7} \cmidrule(lr){8-11} \cmidrule(lr){12-14}
 & & \textbf{Min} & \textbf{\(Q_1\)} & \textbf{Median} & \textbf{\(Q_3\)} & \textbf{Max} & \textbf{$\sum_{prompt}$} & \textbf{$\sum_{completion}$} & \textbf{$\mu_{prompt}$} & \textbf{$\mu_{completion}$} & \textbf{$\sum_{prompt}$} & \textbf{$\sum_{completion}$} & \textbf{$\sum_{combined}$} \\
\midrule
2 & Document & 2 & 3 & 3 & 4 & 100 & 150,541 & 61,904 & 752.71 & 309.52 & 0.04 & 0.12 & 0.16 \\
2 & Chunk & 4 & 12 & 32 & 43 & 4190 & 15,010,574 & 519,830 & 75,052.87 & 2,599.15 & 3.75 & 1.04 & 4.79 \\ 
2 & Combined & 2 & 3 & 7 & 32 & 4190 & 15,161,115 & 581,734 & 37,902.79 & 1,454.34 & 3.79 & 1.16 & 4.95 \\ \hline
9 & Document & 2 & 4 & 4 & 5 & 17 & 241,941 & 68,752 & 1,209.71 & 343.76 & 0.06 & 0.14 & 0.20 \\
9 & Chunk & 6 & 13 & 32 & 47 & 99 & 14,963,931 & 490,556 & 74,819.66 & 2,452.78 & 3.74 & 0.98 & 4.72 \\
9 & Combined & 2 & 4 & 8 & 32 & 99 & 15,205,872 & 559,308 & 38,014.68 & 1,398.27 & 3.80 & 1.12 & 4.92 \\ \hline
12 & Document & 5 & 12 & 17 & 25 & 83 & 327,741 & 297,498 & 1,638.71 & 1,487.49 & 0.08 & 0.59 & 0.68 \\
12 & Chunk & 9 & 62 & 136 & 194 & 665 & 15,214,202 & 2,271,311 & 76,071.01 & 11,356.56 & 3.80 & 4.54 & 8.35 \\
12 & Combined & 5 & 17 & 38 & 135 & 665 & 15,541,943 & 2,568,809 & 38,854.86 & 6,422.02 & 3.89 & 5.14 & 9.02 \\ \hline
13 & Document & 2 & 4 & 5 & 6 & 15 & 426,941 & 84,369 & 2,134.71 & 421.85 & 0.11 & 0.17 & 0.28 \\
13 & Chunk & 34 & 92 & 105 & 117 & 171 & 19,200,124 & 1,734,012 & 96,000.62 & 8,670.06 & 4.80 & 3.47 & 8.27 \\
13 & Combined & 2 & 5 & 25 & 105 & 171 & 19,627,065 & 1,818,381 & 49,067.66 & 4,545.95 & 4.91 & 3.64 & 8.54 \\ \hline
15 & Document & 2 & 5 & 7 & 10 & 25 & 426,941 & 180,699 & 2,134.71 & 903.50 & 0.53 & 1.81 & 2.34 \\
15 & Chunk & 23 & 133 & 156 & 179 & 240 & 19,170,218 & 3,159,981 & 95,851.09 & 15,799.91 & 23.96 & 31.60 & 55.56 \\
15 & Combined & 2 & 7 & 24 & 156 & 240 & 19,597,159 & 3,340,680 & 48,992.90 & 8,351.70 & 24.50 & 33.41 & 57.90 \\ \hline
17 & Document & 3 & 5 & 6 & 9 & 18 & 639,755 & 97,236 & 3,198.78 & 486.18 & 0.16 & 0.19 & 0.35 \\
17 & Chunk & 39 & 108 & 128 & 147 & 268 & 46,978,904 & 2,079,562 & 234,894.52 & 10,397.81 & 11.74 & 4.16 & 15.90 \\
17 & Combined & 3 & 6 & 29 & 127 & 268 & 47,618,659 & 2,176,798 & 119,046.65 & 5,442.00 & 11.90 & 4.35 & 16.26 \\ \hline
19 & Document & 2 & 7 & 10 & 13 & 26 & 639,755 & 230,297 & 3,198.78 & 1,151.49 & 0.80 & 2.30 & 3.10 \\
19 & Chunk & 25 & 116 & 131 & 150 & 209 & 19,159,387 & 3,141,812 & 95,796.94 & 15,709.06 & 23.95 & 31.42 & 55.37 \\
19 & Combined & 2 & 10 & 26 & 131 & 209 & 19,799,142 & 3,372,109 & 49,497.86 & 8,430.27 & 24.75 & 33.72 & 58.47 \\
\hline
37 & Document & 1 & 1 & 1 & 4 & 79 & 530,222 & 3,200 & 2,651.11 & 16.00 & 0.31 & 0.01 & 0.32 \\
37 & Chunk & 9 & 15 & 21 & 36 & 169 & 18,919,395 & 33,723 & 94,596.98 & 168.62 & 10.97 & 0.06 & 11.03 \\
37 & Combined & 1 & 1 & 12 & 22 & 169 & 19,449,617 & 36,923 & 48,624.04 & 92.31 & 11.28 & 0.06 & 11.34 \\
\hline
38 & Document & 1 & 1 & 3 & 4 & 67 & 542,683 & 88,314 & 2,713.41 & 441.57 & 0.08 & 0.05 & 0.13 \\
38 & Chunk & 7 & 17 & 23 & 34 & 141 & 19,338,311 & 682,834 & 96,691.55 & 3,414.17 & 2.90 & 0.41 & 3.31 \\
38 & Combined & 1 & 3 & 9 & 23 & 141 & 19,880,994 & 771,148 & 49,702.49 & 1,927.87 & 2.98 & 0.46 & 3.44 \\
\hline
39 & Document & 5 & 12 & 19 & 25 & 109 & 520,274 & 3,000 & 2,601.37 & 15.00 & 1.56 & 0.05 & 1.61 \\
39 & Chunk & 28 & 83 & 94 & 112 & 226 & 18,704,741 & 31,681 & 93,523.71 & 158.41 & 56.11 & 0.48 & 56.59 \\
39 & Combined & 5 & 19 & 47 & 94 & 226 & 19,225,015 & 34,681 & 48,062.54 & 86.70 & 57.68 & 0.52 & 58.20 \\
\hline
40 & Document & 4 & 6 & 7 & 8 & 10 & 526,118 & 138,691 & 2,630.59 & 694.46 & 0.13 & 0.14 & 0.27 \\
40 & Chunk & 13 & 30 & 34 & 38 & 129 & 18,892,676 & 610,697 & 94,463.38 & 3,053.49 & 4.72 & 0.61 & 5.33 \\
40 & Combined & 4 & 7 & 11 & 34 & 129.4315 & 19,418,794 & 749,388 & 48,546.99 & 1,873.47 & 4.85 & 0.75 & 5.60 \\
\bottomrule
\end{tabular}
\end{adjustbox}
\end{table*}

\begin{table*}[!ht]
\footnotesize
\centering
\caption{Latency, token usage and cost for the candidate-refinement module on FinAgentBench chunk ranking. Cost is API cost: local rerankers and $k$-reciprocal incur no API cost, while commercial rerankers are counted where used. (a)~Component stages emit no LLM completion tokens. (b)~End-to-end pipelines pass the refined pool to \(P_4\)/GPT-5 with ICL-5 and report both prompt and completion tokens.}
\label{tab:finagent_refine_cost}
 
\begin{subtable}{\textwidth}
\centering
\caption{Component stages on the TE3-S pool (retrieval, reranking, and $k$-reciprocal encoding; no completion tokens).}
\label{tab:finagent_refine_cost_stages}
\begin{adjustbox}{max width=\textwidth}
\begin{tabular}{lcccc}
    \toprule
    \multirow{2}{*}{\textbf{Method}} & \textbf{Total Prompt} & \multirow{2}{*}{\textbf{Total Time (Ks)}} & \multirow{2}{*}{\textbf{Time Per Query (s)}} & \multirow{2}{*}{\textbf{Total Cost (USD)}} \\
    & \textbf{Tokens (M)} & & & \\
    \midrule
    \multicolumn{5}{c}{\cellcolor{ailens_purple!35}\textbf{Embedding (first stage)}} \\
    \midrule
    TE3-S & 7.8687 & 1.32 & 6.61 & 0.25 \\
    \midrule
    \multicolumn{5}{c}{\cellcolor{ailens_purple!35}\textbf{Rerankers (on TE3-S pool)}} \\
    \midrule
    TE3-S$\to$Cohere rerank-v4.0-fast     & 8.9715 & 1.75 & 8.73  & 0.65 \\
    TE3-S$\to$Jina reranker v3       & 8.9715 & 2.39 & 11.94 & 0.25 \\
    TE3-S$\to$BGE reranker v2 m3      & 9.1155 & 0.47 & 2.33  & 0.25 \\
    TE3-S$\to$BGE-M3 ColBERT & 8.9715 & 2.09 & 10.43 & 0.25 \\
    TE3-S$\to$ColBERTv2   & 7.2718 & 0.04 & 0.20  & 0.25 \\
    \midrule
    \multicolumn{5}{c}{\cellcolor{ailens_purple!35}\textbf{K-Reciprocal Encoding and Stacking (on TE3-S pool)}} \\
    \midrule
    TE3-S$\to$$k$-recip                      & 7.8687 & 0.84 & 4.18  & 0.25 \\
    TE3-S$\to$$k$-recip$\to$Cohere rerank-v4.0-fast          & 7.2029 & 1.39 & 6.95  & 0.65 \\
    TE3-S$\to$Cohere rerank-v4.0-fast$\to$$k$-recip           & 8.9715 & 2.16 & 10.78 & 0.65 \\
    TE3-S$\to$$k$-recip$\to$BGE reranker v2 m3           & 7.1673 & 0.49 & 2.43  & 0.25 \\
    TE3-S$\to$BGE reranker v2 m3$\to$$k$-recip           & 9.1155 & 0.89 & 4.43  & 0.25 \\
    TE3-S$\to$$k$-recip$\to$BGE-M3 ColBERT      & 7.2029 & 1.56 & 7.78  & 0.25 \\
    TE3-S$\to$BGE-M3 ColBERT$\to$$k$-recip      & 8.9715 & 2.49 & 12.43 & 0.25 \\
    TE3-S$\to$$k$-recip$\to$ColBERTv2        & 7.2029 & 0.15 & 0.75  & 0.25 \\
    TE3-S$\to$ColBERTv2$\to$$k$-recip        & 7.2718 & 0.44 & 2.20  & 0.25 \\
    TE3-S$\to$$k$-recip$\to$Jina reranker v3           & 7.2029 & 1.75 & 8.76  & 0.25 \\
    TE3-S$\to$Jina reranker v3$\to$$k$-recip             & 8.9715 & 2.81 & 14.03 & 0.25 \\
    \bottomrule
\end{tabular}
\end{adjustbox}
\end{subtable}
 
\begin{subtable}{\textwidth}
\centering
\caption{End-to-end pipelines (\(P_4\)/GPT-5, ICL-5) including LLM completion tokens.}
\label{tab:finagent_refine_cost_e2e}
\begin{adjustbox}{max width=\textwidth}
\begin{tabular}{lccccc}
    \toprule
    \multirow{2}{*}{\textbf{Method}} & \textbf{Total Prompt} & \textbf{Total Completion} & \multirow{2}{*}{\textbf{Total Time (Ks)}} & \multirow{2}{*}{\textbf{Time Per Query (s)}} & \multirow{2}{*}{\textbf{Total Cost (USD)}} \\
    & \textbf{Tokens (M)} & \textbf{Tokens (K)} & & & \\
    \midrule
    Baseline (Run 19)                       & 19.1702 & 3.1600 & 27.34 & 136.69 & 55.56 \\
    TE3-S (top 100)                         & 14.5196 & 2.8568 & 38.39 & 191.96 & 46.72 \\
    TE3-S (top 75)                          & 13.0323 & 2.7998 & 34.39 & 171.95 & 44.29 \\
    TE3-S (top 50)                          & 11.2818 & 2.2798 & 28.47 & 142.37 & 36.90 \\
    TE3-S$\to$Cohere rerank-v4.0-fast (top 75)               & 13.1100 & 2.6533 & 35.53 & 177.63 & 42.92 \\
    TE3-S$\to$Cohere rerank-v4.0-fast (top 50)               & 11.2846 & 2.2157 & 29.25 & 146.26 & 36.26 \\
    TE3-S$\to$Cohere rerank-v4.0-fast$\to$$k$-recip (top 50) & 11.1845 & 2.2263 & 29.72 & 148.59 & 36.24 \\
    \bottomrule
\end{tabular}
\end{adjustbox}
\end{subtable}
\end{table*}

\paragraph{FinAgentBench.}
In Table~\ref{tab:full_analysis}, we report the full set of statistics for the runs selected in our feasibility analysis, covering all prompt variants and the ICL-5/TE3-S configuration within the non-agentic workflow. Additionally, we assess the practical feasibility of the proposed approach by analyzing the cost distribution across the selected runs. Document ranking tasks consistently incur low total costs (up to about~\$3.1), reflecting their modest inference requirements. In contrast, chunk ranking tasks dominate the overall expenditure, with total costs ranging from~\$4.7 to~\$56.6, depending on input size and reasoning depth. Combined task runs follow the same pattern, as their total costs are primarily driven by chunk ranking computations. The highest combined costs occur for Runs~19, 39, and~15 (\$58.5, \$58.2, and \$57.9, respectively), whereas earlier runs such as Runs~2 and~9 remain significantly more economical (~\$5) when using smaller models. These results confirm that document-level ranking remains cost-efficient, while chunk ranking incurs substantially higher costs due to its larger context size, remaining the dominant contributor to overall expenditure. Table~\ref{tab:finagent_refine_cost} reports the corresponding figures for the candidate-refinement module, separating the component stages (Table~\ref{tab:finagent_refine_cost_stages}) from the end-to-end pipelines (Table~\ref{tab:finagent_refine_cost_e2e}).

\begin{table*}[!ht]
\footnotesize
\centering
\caption{Comprehensive latency, token usage and cost statistics for different prompt versions and models on FiQA-2018.}
\label{tab:full_fiqa_cost}
\begin{adjustbox}{max width=\textwidth}
\begin{tabular}{lllccccc}
    \toprule
    \multirow{2}{*}{\textbf{Prompt}} & \multirow{2}{*}{\textbf{Model}} & \multirow{2}{*}{\textbf{ICL}} & \textbf{Total Prompt} & \textbf{Total Completion} & \multirow{2}{*}{\textbf{Total Time (Ks)}} & \multirow{2}{*}{\textbf{Time Per Query (s)}} & \multirow{2}{*}{\textbf{Total Cost (USD)}} \\
    & & & \textbf{Tokens (M)} & \textbf{Tokens (K)} & & \\
    \midrule
    \multicolumn{8}{c}{\cellcolor{ailens_purple!35}\textbf{Closed-Source Models}} \\
    \midrule
    \(P_1\) & GPT-4.1    & N/A   & 70.02 & 237.63 & 71.70 & 110.65 & 141.90 \\
    \(P_1\) & GPT-5-mini & N/A   & 82.44 & 293.29 & 38.63 & 59.61 & 21.20 \\
    \(P_1\) & GPT-5      & N/A   & 66.27 & 479.44 & 51.17 & 78.97 & 87.63 \\
    \(P_1\) & GPT-4.1    & ICL-5 & 73.02 & 249.23 & 62.19 & 95.97 & 148.04 \\
    \(P_1\) & GPT-5-mini & ICL-5 & 82.58 & 234.98 & 31.20 & 48.15 & 21.11 \\
    \(P_1\) & GPT-5      & ICL-5 & 82.58 & 474.73 & 56.23 & 86.77 & 107.97 \\
    \midrule
    \(P_2\) & GPT-4.1    & N/A   & 61.60 & 442.55 & 85.43 & 131.84 & 126.74 \\
    \(P_2\) & GPT-5-mini & N/A   & 82.42 & 480.63 & 35.75 & 55.17 & 21.57 \\
    \(P_2\) & GPT-5      & N/A   & 82.42 & 790.47 & 54.24 & 83.70 & 110.93 \\
    \(P_2\) & GPT-4.1    & ICL-5 & 60.85 & 396.73 & 81.37 & 125.57 & 124.87 \\
    \(P_2\) & GPT-5-mini & ICL-5 & 82.56 & 389.64 & 29.29 & 45.20 & 21.42 \\
    \(P_2\) & GPT-5      & ICL-5 & 82.47 & 787.34 & 53.78 & 82.99 & 110.96 \\
    \midrule
    \(P_3\) & GPT-4.1    & N/A   & 40.40 & 344.30 & 82.18 & 126.82 & 83.55 \\
    \(P_3\) & GPT-5-mini & N/A   & 82.43 & 505.36 & 35.09 & 54.15 & 21.62 \\
    \(P_3\) & GPT-5      & N/A   & 82.30 & 606.21 & 49.82 & 76.88 & 108.94 \\
    \(P_3\) & GPT-4.1    & ICL-5 & 58.99 & 328.15 & 77.96 & 120.30 & 120.60 \\
    \(P_3\) & GPT-5-mini & ICL-5 & 82.57 & 442.93 & 29.32 & 45.25 & 21.53 \\
    \(P_3\) & GPT-5      & ICL-5 & 82.57 & 607.09 & 48.14 & 74.29 & 109.28 \\
    \midrule
    \(P_4\) & GPT-4.1    & N/A   & 38.70 & 272.99 & 73.19 & 112.95 & 79.58 \\
    \(P_4\) & GPT-5-mini & N/A   & 82.42 & 341.96 & 34.59 & 53.58 & 21.29 \\
    \(P_4\) & GPT-5      & N/A   & 82.30 & 511.02 & 55.24 & 85.25 & 107.98 \\
    \(P_4\) & GPT-4.1    & ICL-5 & 43.11 & 283.30 & 86.78 & 133.92 & 88.48 \\
    \(P_4\) & GPT-5-mini & ICL-5 & 82.56 & 256.27 & 29.02 & 44.78 & 21.15 \\
    \(P_4\) & GPT-5      & ICL-5 & 82.56 & 516.58 & 52.37 & 80.82 & 108.37 \\
    \midrule
    \multicolumn{8}{c}{\cellcolor{ailens_purple!35}\textbf{Open-Source Models}} \\
    \midrule
    \(P_4\) & DeepSeek-V4-Flash    & N/A   & 49.88 & 594.62 & 15.07 & 23.26 & 68.30 \\
    \(P_4\) & gpt-oss-120b & N/A   & 48.24 & 362.28 & 33.71 & 52.42 & 63.92 \\
    \(P_4\) & grok-4-20-reasoning      & N/A   & 49.88 & 323.22 & 38.16 & 58.89 & 154.50 \\
    \(P_4\) & Llama-4-Maverick-17B-128E-Instruct-FP8    & N/A & 48.91 & 343.41 & 25.89 & 39.96 & 12.57 \\
    \(P_4\) & DeepSeek-V4-Flash    & ICL-5   & 50.21 & 601.99 & 16.60 & 25.61 & 68.55 \\
    \(P_4\) & gpt-oss-120b & ICL-5   & 49.91 & 326.05 & 26.72 & 41.23 & 65.65 \\
    \(P_4\) & grok-4-20-reasoning      & ICL-5   & 50.02 & 318.83 & 35.44 & 54.69 & 154.85 \\
    \(P_4\) & Llama-4-Maverick-17B-128E-Instruct-FP8    & ICL-5 & 49.67 & 360.15 & 24.79 & 38.26 & 13.50 \\
    \bottomrule
\end{tabular}
\end{adjustbox}
\end{table*}

\begin{table*}[!ht]
\footnotesize
\centering
\caption{Latency, token usage and cost for the candidate-refinement module on FiQA-2018. Cost is API cost: local rerankers and $k$-reciprocal incur no API cost, while commercial rerankers are counted where used. (a)~Component stages emit no LLM completion tokens. (b)~The end-to-end pipeline passes the refined pool to GPT-5 and reports both prompt and completion tokens.}
\label{tab:fiqa_refine_cost}
 
\begin{subtable}{\textwidth}
\centering
\caption{Component stages on the TE3-S pool (embedding, reranking, and $k$-reciprocal encoding; no completion tokens).}
\label{tab:fiqa_refine_cost_stages}
\begin{adjustbox}{max width=\textwidth}
\begin{tabular}{lcccc}
    \toprule
    \multirow{2}{*}{\textbf{Method}} & \textbf{Total Prompt} & \multirow{2}{*}{\textbf{Total Time (Ks)}} & \multirow{2}{*}{\textbf{Time Per Query (s)}} & \multirow{2}{*}{\textbf{Total Cost (USD)}} \\
    & \textbf{Tokens (M)} & & & \\
    \midrule
    \multicolumn{5}{c}{\cellcolor{ailens_purple!35}\textbf{Embedding Models}} \\
    \midrule
    TE3-S           & 9.7989 & 1.50 & 2.31 & 0.20 \\
    \midrule
    \multicolumn{5}{c}{\cellcolor{ailens_purple!35}\textbf{Rerankers (on TE3-S pool)}} \\
    \midrule
    TE3-S$\to$Cohere rerank-v4.0-fast     & 172.9717 & 30.53 & 47.11 & 12.84 \\
    TE3-S$\to$BGE reranker v2 m3      & 165.3046 & 1.31  & 2.01  & 0.20 \\
    TE3-S$\to$BGE-M3 ColBERT & 172.9735 & 12.51 & 19.30 & 0.20 \\
    TE3-S$\to$ColBERTv2   & 172.9707 & 2.34  & 3.62  & 0.20 \\
    TE3-S$\to$Jina reranker v3       & 172.9762 & 17.96 & 27.72 & 0.20 \\
    \midrule
    \multicolumn{5}{c}{\cellcolor{ailens_purple!35}\textbf{K-Reciprocal Encoding (on TE3-S pool)}} \\
    \midrule
    TE3-S$\to$$k$-recip                  & 9.7989   & 2.02  & 3.11  & 0.20 \\
    TE3-S$\to$Cohere rerank-v4.0-fast$\to$$k$-recip       & 172.9717 & 8.25  & 12.73 & 3.53 \\
    TE3-S$\to$BGE reranker v2 m3$\to$$k$-recip       & 165.3046 & 12.90 & 19.92 & 0.20 \\
    TE3-S$\to$BGE-M3 ColBERT$\to$$k$-recip  & 172.9735 & 44.22 & 68.24 & 0.20 \\
    TE3-S$\to$ColBERTv2$\to$$k$-recip    & 172.9707 & 9.24  & 14.25 & 0.20 \\
    TE3-S$\to$Jina reranker v3$\to$$k$-recip         & 172.9762 & 28.58 & 44.11 & 0.20 \\
    TE3-S$\to$$k$-recip$\to$Cohere rerank-v4.0-fast      & 47.8230  & 32.54 & 50.22 & 12.96 \\
    TE3-S$\to$$k$-recip$\to$BGE reranker v2 m3       & 50.2152  & 3.32  & 5.12  & 0.20 \\
    TE3-S$\to$$k$-recip$\to$BGE-M3 ColBERT  & 52.5944  & 14.52 & 22.41 & 0.20 \\
    TE3-S$\to$$k$-recip$\to$ColBERTv2    & 52.6015  & 4.36  & 6.73  & 0.20 \\
    TE3-S$\to$$k$-recip$\to$Jina reranker v3        & 52.6015  & 19.98 & 30.83 & 0.20 \\
    \bottomrule
\end{tabular}
\end{adjustbox}
\end{subtable}
 
\begin{subtable}{\textwidth}
\centering
\caption{End-to-end pipeline (refined pool passed to GPT-5) including LLM completion tokens.}
\label{tab:fiqa_refine_cost_e2e}
\begin{adjustbox}{max width=\textwidth}
\begin{tabular}{lccccc}
    \toprule
    \multirow{2}{*}{\textbf{Method}} & \textbf{Total Prompt} & \textbf{Total Completion} & \multirow{2}{*}{\textbf{Total Time (Ks)}} & \multirow{2}{*}{\textbf{Time Per Query (s)}} & \multirow{2}{*}{\textbf{Total Cost (USD)}} \\
    & \textbf{Tokens (M)} & \textbf{Tokens (K)} & & & \\
    \midrule
    TE3-S$\to$Cohere rerank-v4.0-fast$\to$GPT-5 & 52.1589 & 483.5090 & 98.41 & 151.87 & 70.03 \\
    \bottomrule
\end{tabular}
\end{adjustbox}
\end{subtable}
\end{table*}

\paragraph{FiQA-2018.}
Table~\ref{tab:full_fiqa_cost} presents a comprehensive computational cost analysis across all prompt variants and models on the FiQA-2018 dataset, and Table~\ref{tab:fiqa_refine_cost} reports the same statistics for the candidate-refinement stages (Table~\ref{tab:fiqa_refine_cost_stages}) and pipelines (Table~\ref{tab:fiqa_refine_cost_e2e}). GPT-5-mini emerges as a compelling choice for production deployment within the PRISM framework. GPT-5-mini achieves performance between GPT-4.1 and GPT-5 (e.g., $P_1$: 0.5565 NDCG@10) while incurring substantially lower operational costs of approximately \$21, compared to \$142 for GPT-4.1 and \$88 for GPT-5 under the same setting. This represents roughly a 4$\times$ cost reduction compared to GPT-5 with only an 8--10\% performance trade-off, making GPT-5-mini a practical balance point for production systems where both performance and cost efficiency are critical considerations. Furthermore, GPT-5-mini has lower total runtime (38.63Ks) and per-query latency (59.61s) than GPT-4.1 (71.70Ks, 110.65s) and GPT-5 (51.17Ks, 78.97s) in that configuration.

An interesting cost-performance paradox emerges when comparing GPT-5 to GPT-4.1. Despite being the more advanced model, GPT-5 consistently delivers superior performance (0.6130 vs.\ 0.4307 NDCG@10 for $P_1$ without ICL) while maintaining lower operational costs (\$88 vs.\ \$142). This counterintuitive result stems from GPT-5's substantially lower per-token pricing rather than from a straightforward capability-cost relationship: although GPT-5 generates approximately twice as many completion tokens as GPT-4.1, its lower per-token rate more than compensates for the higher token count. Additionally, GPT-5's faster processing time suggests better end-to-end throughput. We note that API pricing is a commercial decision that does not always track computational overhead, and these cost patterns may shift as pricing evolves. Nonetheless, at the time of our experiments, GPT-5 represents a compelling choice for organizations prioritizing both performance and cost efficiency.

The open-source models on FiQA-2018 show substantial variation in both cost and latency that does not map cleanly to ranking quality. grok-4-20-reasoning incurs the highest cost at approximately \$155 with ICL, exceeding even GPT-4.1 (\$148), yet delivers only moderate NDCG@10 of 0.5741, while also being the slowest option at roughly 55 seconds per query. In contrast, Llama-4-Maverick-17B-128E-Instruct-FP8 achieves the lowest cost at \$13.50 with ICL and an NDCG@10 of 0.5581, representing a roughly 8$\times$ cost reduction compared to GPT-5 with about an 8.5\% performance gap under the same $P_4$ ICL setting, together with a lower latency of about 38 seconds per query. DeepSeek-V4-Flash and gpt-oss-120b occupy a middle ground at approximately \$69 and \$66 respectively with ICL, with gpt-oss-120b achieving slightly stronger NDCG@10 (0.5833 vs.\ 0.5778) at a comparable cost, albeit with higher latency (41.23s vs.\ 25.61s).

\clearpage
\onecolumn
{\footnotesize
\setlength{\tabcolsep}{2pt}
\begin{longtable}{lp{3cm}lcccccc}
    \caption{Comprehensive latency, token usage and cost statistics for different prompt versions and models across various evaluation modes on FinanceBench.}
    \label{tab:full_financebench_analysis} \\
    
    \toprule
    \multirow{2}{*}{\textbf{Prompt}} & \multirow{2}{*}{\textbf{Model}} & \multirow{2}{*}{\textbf{Eval Mode}} & \multirow{2}{*}{\textbf{ICL}} & \textbf{Total Prompt} & \textbf{Total Completion} & \textbf{Total} & \textbf{Time Per} & \textbf{Total Cost} \\
    & & & & \textbf{Tokens (K)} & \textbf{Tokens (K)} & \textbf{Time (s)} & \textbf{Query (s)} & \textbf{(USD)}\\
    \midrule
    \endfirsthead
    
    \multicolumn{9}{c}{\tablename\ \thetable\ -- \textit{Continued from previous page}} \\
    \toprule
    \multirow{2}{*}{\textbf{Prompt}} & \multirow{2}{*}{\textbf{Model}} & \multirow{2}{*}{\textbf{Eval Mode}} & \multirow{2}{*}{\textbf{ICL}} & \textbf{Total Prompt} & \textbf{Total Completion} & \textbf{Total} & \textbf{Time Per} & \textbf{Total Cost} \\
    & & & & \textbf{Tokens (K)} & \textbf{Tokens (K)} & \textbf{Time (s)} & \textbf{Query (s)} & \textbf{(USD)}\\
    \midrule
    \endhead
    
    \midrule
    \multicolumn{9}{r}{\textit{Continued on next page}} \\
    \endfoot
    
    \bottomrule
    \endlastfoot

    \multicolumn{9}{c}{\cellcolor{ailens_purple!35}\textbf{Baselines}} \\
    \midrule
    Baseline & GPT-4.1 & Single Store & N/A & 4.81 & 27.13 & 494.94 & 3.30 & 0.23 \\
    Baseline & GPT-4.1 & Shared Store & N/A & 4.81 & 27.72 & 478.99 & 3.19 & 0.23 \\
    Baseline & GPT-4.1 & In Context   & N/A & 1260.12 & 40.70 & 2466.77 & 16.45 & 25.53 \\
    Baseline & GPT-4.1 & Oracle       & N/A & 159.80 & 27.08 & 329.78 & 2.20 & 0.54 \\
    Baseline & GPT-5-mini & Single Store & N/A & 4.81 & 18.39 & 1165.15 & 7.77 & 0.04 \\ 
    Baseline & GPT-5-mini & Shared Store & N/A & 4.81 & 18.69 & 1149.36 & 7.66 & 0.04 \\
    Baseline & GPT-5-mini & In Context   & N/A & 1260.03 & 131.32 & 1505.37 & 10.04 & 3.41 \\
    Baseline & GPT-5-mini & Oracle       & N/A & 158.90 & 138.23 & 1200.31 & 8.00 & 0.32 \\
    Baseline & GPT-5 & Single Store & N/A & 4.81 & 13.33 & 1556.93 & 10.38 & 0.14 \\
    Baseline & GPT-5 & Shared Store & N/A & 4.81 & 13.81 & 1674.66 & 11.16 & 0.14 \\
    Baseline & GPT-5 & In Context   & N/A & 1260.03 & 147.64 & 2066.25 & 13.78 & 17.23 \\
    Baseline & GPT-5 & Oracle       & N/A & 158.90 & 179.69 & 1433.36 & 9.56 & 2.00 \\
    
    \midrule
    \multicolumn{9}{c}{\cellcolor{ailens_purple!35}\textbf{PRISM}} \\
    \midrule
    \(P_1\) & GPT-4.1 & Single Store & ICL-9 & 120.91 & 15.40 & 371.14 & 2.47 & 0.37 \\
    \(P_1\) & GPT-4.1 & Shared Store & ICL-9 & 154.02 & 14.59 & 332.71 & 2.22 & 0.42 \\
    \(P_1\) & GPT-4.1 & In Context   & ICL-9 & 9885.98 & 36.66 & 2609.96 & 17.40 & 20.07 \\
    \(P_1\) & GPT-4.1 & Oracle       & ICL-9 & 377.00 & 26.53 & 285.59 & 1.90 & 0.97 \\
    \(P_1\) & GPT-5-mini & Single Store & ICL-9 & 105.78 & 15.58 & 1649.79 & 11.00 & 0.06 \\
    \(P_1\) & GPT-5-mini & Shared Store & ICL-9 & 112.44 & 14.01 & 1579.32 & 10.53 & 0.06 \\
    \(P_1\) & GPT-5-mini & In Context   & ICL-9 & 9868.73 & 130.71 & 1368.42 & 9.12 & 2.73 \\
    \(P_1\) & GPT-5-mini & Oracle       & ICL-9 & 352.25 & 100.56 & 1104.98 & 7.37 & 0.29 \\
    \(P_1\) & GPT-5   & Single Store & ICL-9 & 138.01 & 13.68 & 1836.46 & 12.24 & 0.31 \\
    \(P_1\) & GPT-5   & Shared Store & ICL-9 & 134.41 & 12.32 & 1834.41 & 12.23 & 0.29 \\
    \(P_1\) & GPT-5   & In Context   & ICL-9 & 9929.93 & 187.48 & 2586.63 & 17.24 & 14.29 \\
    \(P_1\) & GPT-5   & Oracle       & ICL-9 & 341.00 & 166.67 & 1683.26 & 11.22 & 2.09 \\
    \midrule
    \(P_1\) & GPT-4.1 & Single Store & N/A & 4.81 & 17.99 & 404.56 & 2.70 & 0.15 \\
    \(P_1\) & GPT-4.1 & Shared Store & N/A & 4.81 & 16.73 & 373.95 & 2.49 & 0.14 \\
    \(P_1\) & GPT-4.1 & In Context   & N/A & 9744.98 & 39.30 & 1584.33 & 10.56 & 19.80 \\
    \(P_1\) & GPT-4.1 & Oracle       & N/A & 192.80 & 26.96 & 281.26 & 1.88 & 0.60 \\
    \(P_1\) & GPT-5-mini & Single Store & N/A & 4.81 & 17.12 & 1660.00 & 11.07 & 0.04 \\
    \(P_1\) & GPT-5-mini & Shared Store & N/A & 4.81 & 16.22 & 1574.42 & 10.50 & 0.03 \\
    \(P_1\) & GPT-5-mini & In Context   & N/A & 9744.53 & 147.37 & 1484.62 & 9.90 & 2.73 \\
    \(P_1\) & GPT-5-mini & Oracle       & N/A & 192.35 & 113.48 & 1274.31 & 8.50 & 0.28 \\
    \(P_1\) & GPT-5   & Single Store & N/A & 4.81 & 14.16 & 1872.66 & 12.48 & 0.15 \\
    \(P_1\) & GPT-5   & Shared Store & N/A & 4.81 & 13.98 & 1914.46 & 12.76 & 0.15 \\
    \(P_1\) & GPT-5   & In Context   & N/A & 9744.53 & 191.30 & 2651.92 & 17.68 & 14.09 \\
    \(P_1\) & GPT-5   & Oracle       & N/A & 192.35 & 187.49 & 1887.10 & 12.58 & 2.12 \\
    \midrule
    \(P_2\) & GPT-4.1 & Single Store & ICL-9 & 124.14 & 22.52 & 501.14 & 3.34 & 0.43 \\
    \(P_2\) & GPT-4.1 & Shared Store & ICL-9 & 104.34 & 18.15 & 351.93 & 2.35 & 0.35 \\
    \(P_2\) & GPT-4.1 & In Context   & ICL-9 & 9859.73 & 41.07 & 1492.38 & 9.95 & 20.05 \\
    \(P_2\) & GPT-4.1 & Oracle       & ICL-9 & 335.20 & 28.47 & 468.11 & 3.12 & 0.90 \\
    \(P_2\) & GPT-5-mini & Single Store & ICL-9 & 124.14 & 20.94 & 1636.65 & 10.91 & 0.07 \\
    \(P_2\) & GPT-5-mini & Shared Store & ICL-9 & 127.56 & 19.75 & 1548.20 & 10.32 & 0.07 \\
    \(P_2\) & GPT-5-mini & In Context   & ICL-9 & 9891.23 & 157.62 & 1600.95 & 10.67 & 2.79 \\
    \(P_2\) & GPT-5-mini & Oracle       & ICL-9 & 343.40 & 125.91 & 1496.93 & 9.98 & 0.34 \\
    \(P_2\) & GPT-5   & Single Store & ICL-9 & 132.06 & 18.57 & 1867.68 & 12.45 & 0.35 \\
    \(P_2\) & GPT-5   & Shared Store & ICL-9 & 132.96 & 17.82 & 1937.48 & 12.92 & 0.34 \\
    \(P_2\) & GPT-5   & In Context   & ICL-9 & 9912.68 & 186.49 & 1710.18 & 11.40 & 14.26 \\
    \(P_2\) & GPT-5   & Oracle       & ICL-9 & 378.80 & 176.42 & 1251.60 & 8.34 & 2.24 \\
    \midrule
    \(P_2\) & GPT-4.1 & Single Store & N/A & 4.81 & 21.43 & 433.74 & 2.89 & 0.18 \\
    \(P_2\) & GPT-4.1 & Shared Store & N/A & 4.81 & 20.33 & 402.21 & 2.68 & 0.17 \\
    \(P_2\) & GPT-4.1 & In Context   & N/A & 9742.13 & 44.08 & 1469.06 & 9.79 & 19.84 \\
    \(P_2\) & GPT-4.1 & Oracle       & N/A & 189.95 & 30.35 & 335.19 & 2.23 & 0.62 \\
    \(P_2\) & GPT-5-mini & Single Store & N/A & 4.81 & 23.49 & 1678.87 & 11.19 & 0.05 \\
    \(P_2\) & GPT-5-mini & Shared Store & N/A & 4.81 & 23.33 & 1723.72 & 11.49 & 0.05 \\
    \(P_2\) & GPT-5-mini & In Context   & N/A & 9741.68 & 171.78 & 1622.31 & 10.82 & 2.78 \\
    \(P_2\) & GPT-5-mini & Oracle       & N/A & 189.50 & 143.25 & 1590.80 & 10.61 & 0.33 \\
    \(P_2\) & GPT-5   & Single Store & N/A & 4.81 & 19.92 & 2076.89 & 13.85 & 0.21 \\
    \(P_2\) & GPT-5   & Shared Store & N/A & 4.81 & 19.27 & 2047.74 & 13.65 & 0.20 \\
    \(P_2\) & GPT-5   & In Context   & N/A & 9741.68 & 198.25 & 1794.61 & 11.96 & 14.16 \\
    \(P_2\) & GPT-5   & Oracle       & N/A & 189.50 & 191.22 & 1325.28 & 8.84 & 2.15 \\
    \midrule
    \(P_3\) & GPT-4.1 & Single Store & ICL-9 & 90.31 & 19.62 & 379.90 & 2.53 & 0.34 \\
    \(P_3\) & GPT-4.1 & Shared Store & ICL-9 & 128.82 & 23.29 & 430.37 & 2.87 & 0.44 \\
    \(P_3\) & GPT-4.1 & In Context   & ICL-9 & 9908.78 & 55.20 & 1596.56 & 10.64 & 20.26 \\
    \(P_3\) & GPT-4.1 & Oracle       & ICL-9 & 337.70 & 38.70 & 478.79 & 3.19 & 0.99 \\
    \(P_3\) & GPT-5-mini & Single Store & ICL-9 & 128.82 & 22.09 & 1444.10 & 9.63 & 0.08 \\
    \(P_3\) & GPT-5-mini & Shared Store & ICL-9 & 115.14 & 21.04 & 1457.49 & 9.72 & 0.07 \\
    \(P_3\) & GPT-5-mini & In Context   & ICL-9 & 9896.18 & 184.37 & 1852.82 & 12.35 & 2.84 \\
    \(P_3\) & GPT-5-mini & Oracle       & ICL-9 & 330.35 & 134.41 & 1502.07 & 10.01 & 0.35 \\
    \(P_3\) & GPT-5   & Single Store & ICL-9 & 130.81 & 18.69 & 2000.21 & 13.33 & 0.35 \\
    \(P_3\) & GPT-5   & Shared Store & ICL-9 & 139.81 & 18.96 & 2061.80 & 13.75 & 0.36 \\
    \(P_3\) & GPT-5   & In Context   & ICL-9 & 9882.83 & 200.39 & 1768.46 & 11.79 & 14.36 \\
    \(P_3\) & GPT-5   & Oracle       & ICL-9 & 354.50 & 187.28 & 1244.50 & 8.30 & 2.32 \\
    \midrule
    \(P_3\) & GPT-4.1 & Single Store & N/A & 4.81 & 24.26 & 466.67 & 3.11 & 0.20 \\
    \(P_3\) & GPT-4.1 & Shared Store & N/A & 4.81 & 22.01 & 423.23 & 2.82 & 0.19 \\
    \(P_3\) & GPT-4.1 & In Context   & N/A & 9744.83 & 59.43 & 2204.91 & 14.70 & 19.97 \\
    \(P_3\) & GPT-4.1 & Oracle       & N/A & 192.65 & 40.32 & 491.83 & 3.28 & 0.71 \\
    \(P_3\) & GPT-5-mini & Single Store & N/A & 4.81 & 26.19 & 1777.55 & 11.85 & 0.05 \\
    \(P_3\) & GPT-5-mini & Shared Store & N/A & 4.81 & 24.58 & 1448.53 & 9.66 & 0.05 \\
    \(P_3\) & GPT-5-mini & In Context   & N/A & 9744.38 & 195.59 & 1901.49 & 12.68 & 2.83 \\
    \(P_3\) & GPT-5-mini & Oracle       & N/A & 192.20 & 162.52 & 1675.36 & 11.17 & 0.37 \\
    \(P_3\) & GPT-5   & Single Store & N/A & 4.81 & 21.22 & 2072.32 & 13.82 & 0.22 \\
    \(P_3\) & GPT-5   & Shared Store & N/A & 4.81 & 20.98 & 2210.27 & 14.74 & 0.22 \\
    \(P_3\) & GPT-5   & In Context   & N/A & 9744.38 & 213.21 & 1876.14 & 12.51 & 14.31 \\
    \(P_3\) & GPT-5   & Oracle       & N/A & 192.20 & 222.01 & 1449.19 & 9.66 & 2.46 \\
    \midrule
    \(P_4\) & GPT-4.1 & Single Store & ICL-9 & 111.54 & 27.35 & 403.09 & 2.69 & 0.44 \\
    \(P_4\) & GPT-4.1 & Shared Store & ICL-9 & 115.51 & 28.06 & 390.28 & 2.60 & 0.46 \\
    \(P_4\) & GPT-4.1 & In Context   & ICL-9 & 9908.48 & 79.75 & 1720.44 & 11.47 & 20.45 \\
    \(P_4\) & GPT-4.1 & Oracle       & ICL-9 & 333.65 & 50.99 & 535.05 & 3.57 & 1.08 \\
    \(P_4\) & GPT-5-mini & Single Store & ICL-9 & 127.21 & 44.54 & 1950.23 & 13.00 & 0.12 \\
    \(P_4\) & GPT-5-mini & Shared Store & ICL-9 & 124.86 & 43.19 & 2001.19 & 13.34 & 0.12 \\
    \(P_4\) & GPT-5-mini & In Context   & ICL-9 & 9907.43 & 240.14 & 2291.46 & 15.28 & 2.96 \\
    \(P_4\) & GPT-5-mini & Oracle       & ICL-9 & 319.55 & 176.88 & 1747.33 & 11.65 & 0.43 \\
    \(P_4\) & GPT-5   & Single Store & ICL-9 & 118.21 & 24.77 & 2019.59 & 13.46 & 0.40 \\
    \(P_4\) & GPT-5   & Shared Store & ICL-9 & 136.02 & 23.82 & 2165.59 & 14.44 & 0.41 \\
    \(P_4\) & GPT-5   & In Context   & ICL-9 & 9905.78 & 231.55 & 2216.82 & 14.78 & 14.70 \\
    \(P_4\) & GPT-5   & Oracle       & ICL-9 & 327.20 & 234.87 & 1777.46 & 11.85 & 2.76 \\
    \midrule
    \(P_4\) & GPT-4.1 & Single Store & N/A & 4.81 & 28.82 & 521.27 & 3.48 & 0.24 \\
    \(P_4\) & GPT-4.1 & Shared Store & N/A & 4.81 & 27.21 & 484.88 & 3.23 & 0.23 \\
    \(P_4\) & GPT-4.1 & In Context   & N/A & 9746.48 & 82.14 & 1714.16 & 11.43 & 20.15 \\
    \(P_4\) & GPT-4.1 & Oracle       & N/A & 194.30 & 51.59 & 522.74 & 3.48 & 0.80 \\
    \(P_4\) & GPT-5-mini & Single Store & N/A & 4.81 & 42.44 & 2156.44 & 14.38 & 0.09 \\
    \(P_4\) & GPT-5-mini & Shared Store & N/A & 4.81 & 47.24 & 2128.26 & 14.19 & 0.10 \\
    \(P_4\) & GPT-5-mini & In Context   & N/A & 9746.03 & 275.14 & 2597.60 & 17.32 & 2.99 \\
    \(P_4\) & GPT-5-mini & Oracle       & N/A & 193.85 & 194.71 & 2007.18 & 13.38 & 0.44 \\
    \(P_4\) & GPT-5   & Single Store & N/A & 4.81 & 28.72 & 2761.10 & 18.41 & 0.29 \\
    \(P_4\) & GPT-5   & Shared Store & N/A & 4.81 & 31.62 & 2893.80 & 19.29 & 0.32 \\
    \(P_4\) & GPT-5   & In Context   & N/A & 9746.03 & 246.76 & 2276.21 & 15.17 & 14.65 \\
    \(P_4\) & GPT-5   & Oracle       & N/A & 193.85 & 263.13 & 2043.63 & 13.62 & 2.87 \\
    \midrule
    \multicolumn{9}{c}{\cellcolor{ailens_purple!35}\textbf{Open-Source Models}} \\
    \midrule
    \(P_4\) & DeepSeek-V4-Flash & Oracle & ICL-9 & 235.76 & 28.03 & 318.12 & 2.12 & 0.57 \\
    \(P_4\) & gpt-oss-120b & Oracle & ICL-9 & 327.65 & 56.97 & 470.22 & 3.13 & 0.08 \\
    \(P_4\) & grok-4-20-reasoning & Oracle   & ICL-9 & 217.27 & 27.68 & 1142.80 & 10.78 & 1.07 \\
    \(P_4\) & Llama-4-Maverick-17B-128E-Instruct-FP8      & Oracle       & ICL-9 & 304.80 & 29.98 & 385.98 & 2.57 & 0.11 \\
    \(P_4\) & DeepSeek-V4-Flash & Oracle & N/A & 118.76 & 36.39 & 367.44 & 2.45 & 0.51 \\
    \(P_4\) & gpt-oss-120b & Oracle & N/A & 188.45 & 78.20 & 360.28 & 2.40 & 0.08 \\
    \(P_4\) & grok-4-20-reasoning & Oracle   & N/A & 174.47 & 41.69 & 1364.90 & 9.10 & 1.15 \\
    \(P_4\) & Llama-4-Maverick-17B-128E-Instruct-FP8      & Oracle       & N/A & 178.42 & 34.66 & 367.90 & 2.45 & 0.08 \\
\end{longtable}
\setlength{\tabcolsep}{6pt}
}
\twocolumn

\paragraph{FinanceBench.}
Table~\ref{tab:full_financebench_analysis} presents comprehensive computational cost analysis across all prompt variants, models, and evaluation modes on the FinanceBench dataset. Single Store and Shared Store modes consistently incur minimal costs that are below \$0.50 for all models, reflecting their limited prompt token requirements. In contrast, In Context mode dominates the overall expenditure, with costs ranging from \$2.73 to \$25.53 depending on the model, primarily driven by the extensive prompt token requirements needed to embed full document context within the prompt. Oracle mode occupies a middle ground with costs between \$0.08 and \$2.87, as it requires moderate context to provide relevant passages without full document inclusion.

Model selection significantly impacts cost-performance trade-offs. Among closed-source models, GPT-5-mini emerges as the most cost-efficient option across all evaluation modes, with Oracle mode costing \$0.28--\$0.44 compared to \$2.00--\$2.87 for GPT-5 and \$0.54--\$1.08 for GPT-4.1 under similar conditions. Despite this large cost advantage, GPT-5-mini maintains strong performance, matching or approaching GPT-5 accuracy across most configurations. GPT-5, while incurring higher costs, consistently outperforms other models in most configurations. GPT-4.1 exhibits the least favorable cost-performance profile, with comparable costs to GPT-5 in some settings but consistently lower accuracy.

Prompt complexity introduces additional computational overhead with varying returns. The straightforward prompt $P_1$ achieves strong cost-efficiency, maintaining low completion token generation while delivering competitive performance. More complex prompts like $P_4$ generate substantially more completion tokens, increasing costs without proportional performance gains in most evaluation modes. In In Context mode, ICL slightly reduces GPT-5 completion token generation, but the added prompt tokens leave total cost roughly unchanged or marginally higher overall.

Processing time analysis reveals GPT-4.1 as the fastest model in Single Store and Shared Store modes but shows significant variability in other modes. GPT-5 and GPT-5-mini show consistent processing times across different modes, with GPT-5-mini being slightly faster. These results confirm that for production deployment on QA tasks, the In Context mode carries a substantial cost premium that makes it impractical for large-scale applications despite marginal performance benefits. GPT-5-mini with simpler prompts offers a practical balance of accuracy, cost, and latency.

The open-source models evaluated on FinanceBench under Oracle conditions show clear cost-accuracy-latency trade-offs relative to proprietary alternatives. gpt-oss-120b achieves the strongest open-source accuracy (96.67\%) at a total cost of \$0.08 with ICL, representing over a 25$\times$ cost reduction compared to GPT-5 Oracle (\$2.09) while maintaining low latency of about 3.13 seconds per query. Llama-4-Maverick-17B-128E-Instruct-FP8 similarly incurs only \$0.11 with ICL and remains fast at 2.57 seconds per query, though its substantially lower accuracy (57.33\%) limits its practical viability for financial QA. grok-4-20-reasoning presents a different trade-off: it achieves strong accuracy (95.33\%) but at a higher cost of \$1.07 with ICL and much longer latency of 10.78 seconds per query, yielding less favorable efficiency than gpt-oss-120b. DeepSeek-V4-Flash occupies a middle ground at \$0.57 with ICL, moderate accuracy (83.33\%), and relatively low latency (2.12s), making it a reasonable option when proprietary models are too costly but stronger accuracy than Llama-4-Maverick-17B-128E-Instruct-FP8 is still needed. Overall, gpt-oss-120b offers the strongest open-source balance among accuracy, cost, and latency for this setting.

\subsection{Reproducibility Analysis}
\label{sec:reproducibility_details}
Tables~\ref{tab:config_results} and~\ref{tab:welch_results} summarize the descriptive statistics and significance tests done on the competition private subset. The following subsections detail the statistical methodology used to derive these results and provide the full set of per-run scores used in the significance experiments.

\begin{table}[!ht]
    \footnotesize
    \centering
    \caption{Descriptive statistics across multiple runs.}
    \label{tab:config_results}
    \resizebox{\columnwidth}{!}{%
        \begin{tabular}{lcccccc}
        \toprule
        \textbf{Run ID} & \textbf{n} & \textbf{Mean} & \textbf{SD} & \textbf{CV (\%)} & \textbf{CI-Low} & \textbf{CI-High} \\
        \midrule
            12 & 5 & 0.68005 & 0.01023 & 1.50489 & 0.66734 & 0.69276 \\
            15 & 4 & 0.70246 & 0.00661 & 0.94078 & 0.69194 & 0.71297 \\
            18 & 4 & 0.70660 & 0.00546 & 0.77319 & 0.69790 & 0.71529 \\
            19 & 9 & 0.71163 & 0.00433 & 0.60861 & 0.70830 & 0.71496 \\
        \bottomrule
        \end{tabular}%
    }
\end{table}

\begin{table}[!ht]
    \centering
    \footnotesize
    \caption{Welch's $t$-tests vs.\ Run 12.}
    \label{tab:welch_results}
    \resizebox{0.8\columnwidth}{!}{%
        \begin{tabular}{lccc}
        \toprule
        \textbf{Comparison} & \textbf{$t$-stat} & \textbf{$p$-val} & \textbf{Signif.} \\
        \midrule
        12 vs 15 & -3.969 & 0.0057 & Yes \\
        12 vs 18 & -4.981 & 0.0022 & Yes \\
        12 vs 19 & -6.582 & 0.0014 & Yes \\
        \bottomrule
        \end{tabular}%
    }
\end{table}

\subsubsection{Descriptive Statistics}
The following statistical metrics were computed for each configuration based on $n$ independent runs:

\begin{equation}
\bar{x} = \frac{1}{n} \sum_{i=1}^{n} x_i
\end{equation}
where $x_i$ denotes the performance metric from the $i^{\text{th}}$ run. 
The mean $\bar{x}$ represents the average model performance.

\begin{equation}
\mathrm{SD} = \sqrt{\frac{1}{n-1} \sum_{i=1}^{n} (x_i - \bar{x})^2}
\end{equation}
The sample standard deviation (SD) quantifies run-to-run variability where a smaller SD indicates higher stability across repeated evaluations. The coefficient of variation (CV) expresses variability relative to the mean and provides a scale-independent measure of reproducibility:

\begin{equation}
\mathrm{CV} = \frac{s}{\bar{x}} \times 100\%
\end{equation}

A CV below 2\% is generally regarded as evidence of excellent consistency across runs. To quantify the uncertainty around the estimated mean, the 95\% confidence interval (CI) is computed as:
\begin{equation}
\mathrm{CI}_{95\%} = 
\bar{x} \pm t_{1-\frac{\alpha}{2},\,n-1} 
\times \frac{s}{\sqrt{n}},
\quad \text{where } \alpha = 0.05
\end{equation}
Specifically, $t_{1-\frac{\alpha}{2},\,n-1}$ denotes the critical value of the Student's $t$-distribution 
with $(n-1)$ degrees of freedom corresponding to a 95\% confidence level. 
The value of $t_{1-\frac{\alpha}{2},\,n-1}$ can be obtained from a standard $t$-distribution table 
or computed numerically as:
\begin{equation}
t_{1-\frac{\alpha}{2},\,n-1} = 
\mathrm{quantile}_t\!\left(1-\tfrac{\alpha}{2}, \mathrm{df}=n-1\right).
\end{equation}

The non-overlapping confidence intervals between Run~12 and the later runs demonstrate that the configuration changes introduced after Run~12 resulted in a statistically meaningful performance gain. Subsequent runs (15--19) show only minor numerical variation, suggesting that performance has stabilized and that the system's behavior is reproducible. Full descriptive statistics and confidence intervals are reported in Table~\ref{tab:config_results}.

\subsubsection{Statistical Significance}
\begin{table*}[!t]
\centering
\caption{Runs that are used for statistical significance experiment, scored on the competition test set.}
\label{tab:stats_sig_runs}
\begin{adjustbox}{width=\textwidth}
\begin{tabular}{@{}c c c c c c c c c c c@{}}
\toprule
\multirow{2}{*}{\textbf{ID}} & \multicolumn{3}{c}{\textbf{Document Ranking Configuration}} & \multicolumn{3}{c}{\textbf{Chunk Ranking Configuration}} & \multicolumn{2}{c}{\textbf{Agentic Configuration}} & \multicolumn{2}{c}{\textbf{NDCG@5 Score}} \\
\cmidrule(lr){2-4} \cmidrule(lr){5-7} \cmidrule(lr){8-9} \cmidrule(lr){10-11}
 & \textbf{Prompt} & \textbf{Model} & \textbf{ICL/Embedding} & \textbf{Prompt} & \textbf{Model} & \textbf{ICL/Embedding} & \textbf{Doc. Agent} & \textbf{Chunk Agent} & \textbf{Public} & \textbf{Private} \\
\midrule
12 & \(P_3\) & GPT-5-mini & N/A & \(P_3\) & GPT-5-mini & N/A & \multicolumn{2}{c}{N/A} &0.64144 & 0.66537 \\							
12 & \(P_3\) & GPT-5-mini & N/A & \(P_3\) & GPT-5-mini & N/A & \multicolumn{2}{c}{N/A} &0.64952 & 0.67366 \\							
12 & \(P_3\) & GPT-5-mini & N/A & \(P_3\) & GPT-5-mini & N/A & \multicolumn{2}{c}{N/A} &0.65520 & 0.68523 \\							
12 & \(P_3\) & GPT-5-mini & N/A & \(P_3\) & GPT-5-mini & N/A & \multicolumn{2}{c}{N/A} &0.64193 & 0.68579 \\							
12 & \(P_3\) & GPT-5-mini & N/A & \(P_3\) & GPT-5-mini & N/A & \multicolumn{2}{c}{N/A} &0.64297 & 0.69019 \\ \hline							
15 & \(P_4\) & GPT-5 & N/A & \(P_4\) & GPT-5 & N/A & \multicolumn{2}{c}{N/A} &0.67047 & 0.69640 \\							
15 & \(P_4\) & GPT-5 & N/A & \(P_4\) & GPT-5 & N/A & \multicolumn{2}{c}{N/A} &0.66720 & 0.69735 \\							
15 & \(P_4\) & GPT-5 & N/A & \(P_4\) & GPT-5 & N/A & \multicolumn{2}{c}{N/A} &0.66579 & 0.70630 \\							
15 & \(P_4\) & GPT-5 & N/A & \(P_4\) & GPT-5 & N/A & \multicolumn{2}{c}{N/A} &0.66236 & 0.70977 \\ \hline		
18 & \(P_4\) & GPT-5-mini & ICL-5/TE3-S & \(P_4\) & GPT-5 & N/A & \multicolumn{2}{c}{N/A} &0.68209 & 0.70187 \\							
18 & \(P_4\) & GPT-5-mini & ICL-5/TE3-S & \(P_4\) & GPT-5 & N/A & \multicolumn{2}{c}{N/A} &0.67842 & 0.70455 \\							
18 & \(P_4\) & GPT-5-mini & ICL-5/TE3-S & \(P_4\) & GPT-5 & N/A & \multicolumn{2}{c}{N/A} &0.67514 & 0.70551 \\ 	
18 & \(P_4\) & GPT-5-mini & ICL-5/TE3-S & \(P_4\) & GPT-5 & N/A & \multicolumn{2}{c}{N/A} &0.67373 & 0.71446 \\ \hline							
19 & \(P_4\) & GPT-5 & ICL-5/TE3-S & \(P_4\) & GPT-5 & N/A & \multicolumn{2}{c}{N/A} &0.68101 & 0.70491 \\							
19 & \(P_4\) & GPT-5 & ICL-5/TE3-S & \(P_4\) & GPT-5 & N/A & \multicolumn{2}{c}{N/A} &0.67913 & 0.70827 \\							
19 & \(P_4\) & GPT-5 & ICL-5/TE3-S & \(P_4\) & GPT-5 & N/A & \multicolumn{2}{c}{N/A} &0.68006 & 0.70865 \\							
19 & \(P_4\) & GPT-5 & ICL-5/TE3-S & \(P_4\) & GPT-5 & N/A & \multicolumn{2}{c}{N/A} &0.67585 & 0.70923 \\							
19 & \(P_4\) & GPT-5 & ICL-5/TE3-S & \(P_4\) & GPT-5 & N/A & \multicolumn{2}{c}{N/A} &0.68172 & 0.71181 \\							
19 & \(P_4\) & GPT-5 & ICL-5/TE3-S & \(P_4\) & GPT-5 & N/A & \multicolumn{2}{c}{N/A} &0.67845 & 0.71277 \\							
19 & \(P_4\) & GPT-5 & ICL-5/TE3-S & \(P_4\) & GPT-5 & N/A & \multicolumn{2}{c}{N/A} &0.66495 & 0.71375 \\							
19 & \(P_4\) & GPT-5 & ICL-5/TE3-S & \(P_4\) & GPT-5 & N/A & \multicolumn{2}{c}{N/A} &0.66680 & 0.71713 \\							
19 & \(P_4\) & GPT-5 & ICL-5/TE3-S & \(P_4\) & GPT-5 & N/A & \multicolumn{2}{c}{N/A} &0.67444 & 0.71818 \\							
\bottomrule
\end{tabular}
\end{adjustbox}
\end{table*}

In order to verify that improvements of PRISM are statistically significant, Welch's two-sample $t$-tests were conducted between Run~12 (baseline configuration) and each subsequent run. The test statistic is defined as:
\begin{equation}
t = 
\frac{\bar{x}_1 - \bar{x}_2}{
\sqrt{\tfrac{s_1^2}{n_1} + \tfrac{s_2^2}{n_2}}
},
\end{equation}
where $\bar{x}_1, s_1, n_1$ and $\bar{x}_2, s_2, n_2$ denote the mean, standard deviation, and sample size of each group. 
The degrees of freedom are approximated using the Welch-Satterthwaite equation:
\begin{equation}
\nu =
\frac{\left(\tfrac{s_1^2}{n_1} + \tfrac{s_2^2}{n_2}\right)^2}{
\tfrac{(s_1^2/n_1)^2}{n_1-1} +
\tfrac{(s_2^2/n_2)^2}{n_2-1}
}.
\end{equation}
The two-tailed $p$-value is obtained from the cumulative $t$-distribution:
\begin{equation}
p = 2 \times \mathrm{CDF}_t(-|t|, \nu).
\end{equation}
A comparison is deemed statistically significant when $p < \alpha$, where $\alpha = 0.05$ corresponds to a 5\% significance level with a 5\% probability of incorrectly rejecting the null hypothesis.

\subsubsection{Interpretation of Statistical Significance}
Table~\ref{tab:stats_sig_runs} summarizes the configurations and corresponding NDCG@5 scores used in the statistical significance experiments. All pairwise comparisons involving Run~12 yield $p$-values below the 0.05 threshold (Table~\ref{tab:welch_results}), confirming that every subsequent configuration achieves a statistically significant improvement over the baseline. Together with the low variance reported in Table~\ref{tab:config_results}, these results confirm that PRISM's performance gains are both statistically significant and reproducible.

\subsubsection{Query-Level Uncertainty and Separability}
\label{sec:adjacent_significance}
The Welch tests above quantify run-to-run variance across repeated submissions. They do not quantify query-level uncertainty on the re-annotated set, so we add a query-level percentile bootstrap with 10{,}000 resamples. Document and chunk have disjoint query sets, so each task is resampled independently and averaged within each replicate. Point estimates here come from re-scoring the released prediction files rather than from the tables, and agree with them to a mean absolute difference of 0.0005 across the 34 configurations that carry both bases.

\noindent\textbf{Minimum detectable effect.}
From the observed spread of paired per-query differences, the minimum detectable effect at 80\% power is 0.0570 for chunk ranking ($n=199$) and 0.0128 for document ranking ($n=200$); at 90\% power these become 0.0660 and 0.0148. Differences smaller than these are not resolvable by this evaluation, whatever their sign.

\noindent\textbf{Adjacent pairs.}
Ordering all 40 configurations by combined NDCG@5 gives 39 adjacent pairs. Applying the paired bootstrap to each and correcting with Holm's method, 12 of the 39 remain significant. That is the reason the main text groups findings by module instead of presenting the grid as a ranking, and the reason no two configurations are described as ordered when their intervals overlap. The illustrative case is Run~13 against Run~14: they differ by 0.0006 combined NDCG@5, and the 95\% CI for the paired difference is $[-0.0151, +0.0192]$, an interval about fifty times wider than the difference and containing zero.

\begin{figure*}[ht]
\centering
\includegraphics[width=\textwidth]{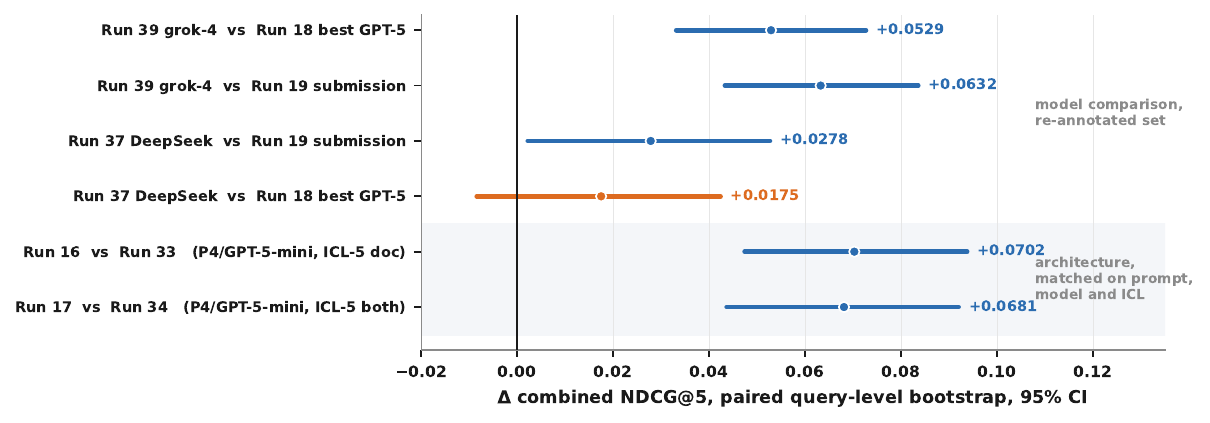}
\caption{Paired query-level bootstrap differences in combined NDCG@5 with 95\% confidence intervals, 10{,}000 resamples, re-annotated basis. Four model comparisons and the two architecture-matched pairs; only DeepSeek against Run~18 crosses zero.}
\label{fig:paired_diffs}
\end{figure*}

\noindent\textbf{Model comparisons.}
Figure~\ref{fig:paired_diffs} plots the comparisons the paper relies on. Against Run~18, the strongest GPT-5-family configuration on the re-annotated basis, Run~39 gains $+0.0529$ ([$+0.0331$, $+0.0727$]) and Run~37 gains $+0.0175$ ([$-0.0084$, $+0.0424$]); against Run~19 the same runs gain $+0.0632$ ([$+0.0433$, $+0.0835$]) and $+0.0278$ ([$+0.0022$, $+0.0527$]). Only the DeepSeek-against-Run~18 interval crosses zero, which is why that claim is stated without asserting superiority. The two architecture-matched differences are $+0.0702$ ([$+0.0474$, $+0.0937$]) and $+0.0681$ ([$+0.0436$, $+0.0920$]), with chunk Wilcoxon $p = 9.4\times10^{-9}$ and $2.8\times10^{-7}$ and document $p = 0.32$ and $0.20$.

\noindent\textbf{What is and is not separable.}
The minimum detectable effect at 80\% power is 0.057 for chunk ranking ($n=199$) and 0.013 for document ranking ($n=200$), so adjacent pairs in a grid this dense are mostly not separable: only 12 of 39 survive Holm correction (Appendix~\ref{sec:adjacent_significance}). Runs~13 and~14 differ by 0.0006 with a 95\% CI of $[-0.0151, +0.0192]$, roughly fifty times the difference and containing zero. We therefore group findings by module rather than presenting the 40 configurations as an ordering, and never describe configurations as ordered when their intervals overlap, whatever the point gap suggests.

\subsection{Production Readiness Analysis}
\label{sec:production_readiness}

\noindent\textbf{Does the best configuration require expensive models?}
Adoption is not gated on the expensive models: a configuration costing a fifth as much lands in the same quality band as the strongest GPT-5-family run. DeepSeek-V4-Flash reaches 0.8016 combined NDCG@5 at USD 0.0284 per query, against 0.7842 for Run~18, the strongest GPT-5-family configuration. We put those in the same band rather than call one better: the paired difference is positive but its 95\% CI includes zero, and it holds under 9 of the 16 alternative label constructions. Run~18 has no measured cost, so the cost comparison uses the GPT-5 runs that do, at USD 0.1462 and USD 0.1448 per query (Runs~19 and~15, Table~\ref{tab:full_analysis}), roughly five times DeepSeek's. Llama-4-Maverick gives a cheaper point again at 0.7627 and USD 0.0133 per query. The highest-scoring configuration in the paper does cost more, at USD 0.1415, so a practitioner who wants the very top of the table pays for it; what has changed is that GPT-5-family quality no longer requires it. Since PRISM needs no training, moving between providers is a change of endpoint rather than a re-implementation.

We do agree that API dependence remains a real reproducibility limitation, because exact reproduction of API-served results depends on endpoint availability and versioning. That is why the evaluation itself, rather than the endpoints, is what we make reproducible: the expert annotations, scoring pipeline, prompts and configuration details are released, with exact endpoint identifiers and the per-run call timestamps. We state that distinction rather than equating training-free operation with infrastructure independence.

\begin{table}[!h]
\footnotesize
\centering
\caption{FACET on FinAgentBench for the configurations with measured cost and latency ($L_{\mathrm{ceil}}{=}30$\,s, $C_{\mathrm{ceil}}{=}\$0.25$, $Q_{\mathrm{floor}}{=}0.65$; $Q=$ combined NDCG@5, re-annotated basis). Each run applies the same prompt and model to both ranking tasks; IDs match Appendix Table~\ref{tab:ablation}, and the per-axis ratios are in Appendix Table~\ref{tab:facet_ratios}. Runs~2--19 use combined document--chunk per-query accounting; Runs~38--40 retain post-annotation chunk-stage totals divided by 400 tasks, while Run~37 retains its originally reported measurements.}
\label{tab:facet_finagent}
\resizebox{\columnwidth}{!}{%
\begin{tabular}{clccccc}
\toprule
\textbf{ID} & \textbf{Configuration} & \textbf{ICL} & \textbf{Lat.\,(s)} & \textbf{\$/query} & \textbf{NDCG@5} & \textbf{FACET} \\
\midrule
2 & P1/GPT-5-mini & none & 28.1 & 0.0124 & 0.7576 & 2.695 \\
9 & P2/GPT-5-mini & none & 23.1 & 0.0123 & 0.7665 & 2.824 \\
12 & P3/GPT-5-mini & none & 57.7 & 0.0226 & 0.7629 & 1.956 \\
13 & P4/GPT-5-mini & none & 103.3 & 0.0214 & 0.7767 & 1.786 \\
15 & P4/GPT-5 & none & 76.1 & 0.1448 & 0.7776 & 1.070 \\
17 & P4/GPT-5-mini & ICL-5 both & 75.6 & 0.0407 & 0.7515 & 1.541 \\
19 & P4/GPT-5 & ICL-5 doc & 84.6 & 0.1462 & 0.7747 & 1.042 \\
37 & P4/DeepSeek-V4-Flash & ICL-5 doc & 18.5 & 0.0284 & 0.8016 & 2.349 \\
\rowcolor{ailens_purple!30}38 & P4/gpt-oss-120b & ICL-5 doc & 15.4 & 0.0083 & 0.6582 & \textbf{3.195} \\
39 & P4/grok-4-20-reasoning & ICL-5 doc & 48.9 & 0.1415 & 0.8372 & 1.221 \\
40 & P4/Llama-4-Maverick & ICL-5 doc & 17.0 & 0.0133 & 0.7627 & 2.923 \\
\bottomrule
\end{tabular}}
\end{table}

This section examines the operational properties of each PRISM component in turn. Although our best-performing configuration relies on prompt engineering, selective ICL, and candidate refinement, we also cover multi-agent coordination so that practitioners can see the full design space and the conditions under which each component is worth deploying.

\subsubsection{Module-Level Findings on FinAgentBench}
\noindent\textbf{Deployment findings.}
Three patterns hold across the grid in Appendix Table~\ref{tab:ablation}. \(P_4\) gives the most stable non-agentic ranking; ICL pays off when scoped to document ranking and not by default at chunk granularity, where exemplars consume context the candidate evidence needs; and MAS is not a default choice for chunk ranking. Shot-count trends and MAS details are in Appendix~\ref{sec:mas_architecture_details}. These are findings about financial IR on these three corpora; we have no cross-vertical evidence and do not claim any.

\subsubsection{Prompt Engineering}
Prompt engineering is the simplest component to operationalize, delivering stable, low-variance performance with minimal latency and token overhead. The four variants support a practical deployment ladder: simpler prompts for cost-sensitive traffic, \(P_4\) for stronger ranking reliability. Prompt length itself remains a constraint, because longer instructions degrade performance at high context lengths (Appendix~\ref{sec:full_limitations}), which is why \(P_4\) keeps the reasoning structure but removes avoidable verbosity.

\subsubsection{In-Context Learning Sample Retrieval}
ICL improves reasoning consistency and retrieval quality but introduces additional computational cost, so it should be activated selectively for complex or high-stakes financial queries where improved contextual understanding outweighs efficiency concerns. Combined with a strong prompt design, it can be deployed under adaptive strategies that vary the number of retrieved exemplars with query complexity or token budget.

\subsubsection{Candidate Refinement}
Candidate refinement is the component with the clearest operational payoff, because it is the only one that reduces the token volume reaching the LLM. Its stages are stateless and locally executable apart from the commercial reranker, so they add no training or serving infrastructure. The trade-off is latency: on FinAgentBench the refined pipeline is roughly 10\,s per query slower than the unrefined baseline (Table~\ref{tab:finagent_e2e}), which matters for interactive deployments but not for batch analysis.

\subsubsection{Multi-Agent Coordination}
The agentic workflow extends the same prompt scaffolding to a multi-role architecture in which specialized agents inherit the ReAct reasoning framework and each scores chunks on a 1--10 scale within its focus area (Appendix~\ref{sec:mas_architecture_details}). This buys interpretable, role-separated reasoning, but at the cost of coordination overhead and variable latency. Under the matched comparison of Section~\ref{sec:mas_matched}, coordination is not measurably harmful for coarse document routing and costs about 0.14 NDCG@5 at chunk granularity, so we treat it as an optional module for document-level routing and not as the production default for chunk ranking.

\noindent\textbf{Where the loss occurs.}
Nothing measurable goes wrong at the document stage, where the agentic variant is nominally ahead and neither difference is distinguishable from zero. What separates the stages is how much context each already carries: document prompts average about 3.2k input tokens per query against about 95.8k for chunk prompts (Appendix Table~\ref{tab:full_analysis}). A second measurement agrees, since within \(A_4\) adding exemplars degrades the architecture monotonically (Runs~28--31, 0.71505 at ICL-5 to 0.65400 at ICL-15). Coordination and exemplars both consume context and both cost most where least is to spare; we offer that as the interpretation these measurements support, not a causal isolation. We evaluated \(P_4\) only for \(A_4\), so the conclusion is scoped to the tested \(A_4\) workflow: coordination is not measurably harmful for coarse document routing, and that is the boundary this module exists to locate.

\subsubsection{Three-Axis Dominance and FACET Decision Boundaries}
\label{sec:dominance_appendix}
Section~\ref{sec:practitioner} presents dominance before FACET. This appendix reports the dominance structure itself and then shows what the scalarization adds once dominance has done its work.

Figure~\ref{fig:pareto} plots cost against combined NDCG@5 with latency as marker size, and that two-axis projection has 5 non-dominated configurations. Treating latency as a third dominance axis rather than as a visual encoding keeps one further configuration, Run~40, giving the 6 rows marked as non-dominated in Table~\ref{tab:dominance}. The frontier does real work here: it removes Runs~12, 15, 17, and 19 without reference to any preference, and Table~\ref{tab:dominance} names a configuration that beats each of them on all three axes. Run~40 illustrates why the third axis matters, since it survives on its 17.0\,s latency and would not stand out in a two-dimensional cost--quality reading.

What the frontier leaves is 6 configurations spanning 15.4\,s to 103.3\,s, \$0.0083 to \$0.1415, and 0.6582 to 0.8372 combined NDCG@5, with no dominance relation among them. Choosing one is a statement of preference, and the only question is whether that statement is written down where a reader can check it. Holding the latency weight at $w_L = 0.2$ and letting the quality weight $w_Q$ absorb weight from cost, the FACET-preferred configuration changes twice inside the frontier (Table~\ref{tab:facet_boundaries}); the two boundaries are solved directly from the ratio columns of Table~\ref{tab:facet_ratios} rather than estimated. Three different configurations are preferred in three different regions, all of them on the frontier, so the scalarization discriminates between defensible options rather than restating the accuracy ranking. The 12 scenarios of Appendix~\ref{sec:facet_robustness} all sit at $w_Q \le 0.6$, on one side of the first boundary, which is why the top configuration does not move within that sweep. That sweep does not validate the operating targets; it illustrates how a stated preference maps to a choice, which is the guidance a frontier plot cannot express because the plot is silent on preferences by construction.

\begin{table}[!h]
\footnotesize
\centering
\caption{Three-axis dominance over the configurations with measured cost and latency (lower latency, lower cost, higher combined NDCG@5; re-annotated basis). Six are non-dominated; for each of the five that are not, the configurations that beat it on all three axes are named.}
\label{tab:dominance}
\resizebox{\columnwidth}{!}{%
\begin{tabular}{@{}clrrrcl@{}}
\toprule
\textbf{ID} & \textbf{Configuration} & \textbf{Lat.\,(s)} & \textbf{\$/q} & \textbf{NDCG@5} & \textbf{Non-dom.} & \textbf{Beaten by} \\
\midrule
38 & P4/gpt-oss-120b & 15.4 & 0.0083 & 0.6582 & yes & \\
40 & P4/Llama-4-Maverick & 17.0 & 0.0133 & 0.7627 & yes & \\
37 & P4/DeepSeek-V4-Flash & 18.5 & 0.0284 & 0.8016 & yes & \\
9 & P2/GPT-5-mini & 23.1 & 0.0123 & 0.7665 & yes & \\
39 & P4/grok-4-20-reasoning & 48.9 & 0.1415 & 0.8372 & yes & \\
13 & P4/GPT-5-mini & 103.3 & 0.0214 & 0.7767 & yes & \\
\midrule
2 & P1/GPT-5-mini & 28.1 & 0.0124 & 0.7576 & no & 9 \\
12 & P3/GPT-5-mini & 57.7 & 0.0226 & 0.7629 & no & 9 \\
17 & P4/GPT-5-mini, ICL-5 both & 75.6 & 0.0407 & 0.7515 & no & 9, 37, 40 \\
15 & P4/GPT-5 & 76.1 & 0.1448 & 0.7776 & no & 37, 39 \\
19 & P4/GPT-5, ICL-5 doc & 84.6 & 0.1462 & 0.7747 & no & 37, 39 \\
\bottomrule
\end{tabular}}
\end{table}

\begin{table}[!h]
\footnotesize
\centering
\caption{FACET-preferred configuration inside the non-dominated set as the quality weight varies, with $w_L$ held at 0.2 and $w_C = 0.8 - w_Q$. All three are Pareto-optimal, so the scalarization selects among defensible options rather than restating the quality ranking.}
\label{tab:facet_boundaries}
\resizebox{\columnwidth}{!}{%
\begin{tabular}{@{}lll@{}}
\toprule
\textbf{Quality weight} $\boldsymbol{w_Q}$ & \textbf{Preferred} & \textbf{Reading} \\
\midrule
below 0.642 & 38, gpt-oss-120b & cost- and latency-weighted \\
0.642 to 0.771 & 40, Llama-4-Maverick & intermediate \\
above 0.771 & 37, DeepSeek-V4-Flash & quality-weighted \\
\bottomrule
\end{tabular}}
\end{table}

\subsubsection{FACET: An SLA-Specific Scalarization}
The frontier says which configurations are defensible; choosing among them still requires a stated preference, and that is all FACET supplies. Each factor is a ratio of an observed value to its own threshold, so every factor is dimensionless before multiplication, and the aggregator is a weighted geometric rather than arithmetic mean, which makes the induced ranking invariant to the scale each criterion is measured on. That invariance, not the absolute score, is the property we defend: FACET does not define production readiness and does not replace reporting the three axes separately. Appendix~\ref{sec:facet_appendix} motivates the aggregator and thresholds, which each deployment sets for itself. With latency $L$, cost $C$, quality $Q$, and thresholds $L_{\mathrm{ceil}}$, $C_{\mathrm{ceil}}$, and $Q_{\mathrm{floor}}$, the score is
{
\[
\begin{aligned}
\mathrm{FACET}&=\left(\frac{L_{\mathrm{ceil}}}{L}\right)^{w_L}
\left(\frac{C_{\mathrm{ceil}}}{C}\right)^{w_C}\\
&\quad\times\left(\frac{Q}{Q_{\mathrm{floor}}}\right)^{w_Q},\\
\sum_i w_i&=1.
\end{aligned}
\]}

For \textsc{FinAgentBench} we set a 30\,s latency ceiling, a \$0.25/query cost ceiling, and a 0.65 NDCG@5 quality floor (Appendix Table~\ref{tab:facet_finagent}). These are illustrative targets for interactive financial retrieval, not values from a user study or live deployment, which is why the raw axes are reported alongside so another deployment can substitute its own. GPT-5 stays quality-strong but cost- and latency-limited (Run~19, 1.042), while gpt-oss-120b ranks highest (Run~38, 3.195) because its \$0.0083/query and 15.4\,s offset a lower NDCG@5. That follows from the thresholds a practitioner sets, not from gpt-oss-120b ranking better: its NDCG@5 of 0.6582 only just clears the 0.65 floor, so a stricter bar would rank it below the GPT-5 runs. Across the 12 scenarios of Appendix~\ref{sec:facet_robustness}, which vary each threshold by $\pm10\%$ and add three single-axis-dominant weightings, Run~38 stays top, the two GPT-5 runs stay bottom, and the middle moves by at most two positions. FACET is consistent with the dominance analysis rather than an alternative to it: its top three configurations are all Pareto-optimal in Figure~\ref{fig:pareto}. Read with the module-level findings, it favors the non-agentic path: ranking in one LLM call, ICL only where it helps, and no coordination overhead (Section~\ref{sec:mas_matched}, Appendix~\ref{sec:production_readiness}).

\subsubsection{FACET Metric Rationale}
\label{sec:facet_appendix}
The FACET definition above is the weighted geometric mean of three ratios, $r_L=L_{\mathrm{ceil}}/L$, $r_C=C_{\mathrm{ceil}}/C$, and $r_Q=Q/Q_{\mathrm{floor}}$, reported as separate columns in Table~\ref{tab:facet_ratios}, with $w_i>0$. Scores above 1 exceed the deployment requirements overall, scores equal to 1 meet them, and scores below 1 fall short. This section explains the choice of aggregator and the choice of thresholds.

We use the weighted \emph{geometric} mean rather than the arithmetic mean for two reasons. First, the induced ranking is invariant to the normalization scale of the underlying criteria, a property established for ratio-based multi-criteria aggregation~\cite{krejci2018aggregation,aczel1983procedures}. Second, a severe shortfall on any single axis creates a multiplicative penalty that cannot be fully offset by strong performance on the other axes. Under equal weights, halving $r_L$ from $1.0$ to $0.5$ reduces the geometric mean by a factor of $2^{1/3} \approx 1.26$, while doubling $r_C$ from $1.0$ to $2.0$ increases it by the same factor. Compensation is therefore symmetric but not unlimited: a near-zero ratio on any axis drives the product toward zero.

For example, under thresholds $L_{\mathrm{ceil}}=30$\,s,
$C_{\mathrm{ceil}}=\$0.25$, $Q_{\mathrm{floor}}=0.65$ and the default weights $(0.2,0.3,0.5)$, a system with $L=6$\,s, $C=\$0.01$, and $Q=0.90$ has $r_L=5.0$, $r_C=25.0$, and $r_Q\approx1.38$, yielding $\mathrm{FACET}=5.0^{0.2}\cdot25.0^{0.3}\cdot1.38^{0.5}\approx4.26$, well above~1. In contrast, a system with $L=35$\,s, $C=\$0.30$, and $Q=0.70$ has $r_L\approx0.86$, $r_C\approx0.83$, and $r_Q\approx1.08$, yielding $\mathrm{FACET}\approx0.95<1$. Although its quality exceeds the floor, the latency and cost shortfalls pull the overall score below the deployment threshold.

\subsubsection{Robustness of FACET Rankings}
\label{sec:facet_robustness}

A FACET score depends on practitioner-set thresholds and weights rather than fixed values from the data. A ranking built on such choices is useful for deployment decisions only if it does not change when those choices are varied within a reasonable range. We therefore follow the composite-indicator robustness methodology of \citet{saisana2005uncertainty} and verify that the FACET ranking in Table~\ref{tab:facet_finagent} reflects genuine differences between configurations rather than a particular setting of the thresholds or weights.
 
We vary two elements. \emph{(i)~Thresholds:} each of the three thresholds is scaled by $\pm10\%$ (panels S1 and S2 in Table~\ref{tab:facet_full12}), giving $L_{\mathrm{ceil}}\in\{27,30,33\}$\,s, $C_{\mathrm{ceil}}\in\{0.225,0.250,0.275\}$, and $Q_{\mathrm{floor}}\in\{0.585,0.650,0.715\}$. \emph{(ii)~Weights:} besides the default scheme $(w_L,w_C,w_Q)=(0.2,0.3,0.5)$, we add three single-axis-dominant schemes: quality-dominated $(0.2,0.2,0.6)$, cost-dominated $(0.2,0.6,0.2)$, and latency-dominated $(0.6,0.2,0.2)$. Crossing the three threshold levels with the four weight schemes gives the 12 scenarios in Table~\ref{tab:facet_full12}; quality is measured by NDCG@5, and using MAP@5 or MRR@5 leaves the ordering unchanged.
 
In every one of the 12 columns of Table~\ref{tab:facet_full12}, gpt-oss-120b (Run~38) has the highest score and the two GPT-5 runs (15 and~19) have the lowest among passing configurations. The intermediate configurations move by at most two rank positions, and only under the off-balance schemes: under latency-dominated weights grok-4 (Run~39) moves from rank~9 to~7, passing the two GPT-5-mini variants Runs~13 and~17, because its latency ratio ($r_L=0.61$) exceeds theirs ($0.29$ and $0.40$); Run~13 falls from~7 to~9 for the same reason. The top and bottom of the ranking are fixed because the gaps in $r_L$, $r_C$, and $r_Q$ between configurations exceed the shifts a $\pm10\%$ threshold or weight change produces. For example, gpt-oss-120b has $r_C=30.2$ and $r_L=1.94$ against $r_C=1.7$ and $r_L=0.35$ for GPT-5 (Run~19); GPT-5 leads only on quality ($r_Q=1.19$ versus $1.01$), so it overtakes gpt-oss-120b only as $w_Q$ increases outside the tested range.
 
The main weight column is identical across the three threshold panels. With $w_L+w_C=w_Q=0.5$, scaling all thresholds by a common factor $k$ maps $r_L\!\to\!k\,r_L$, $r_C\!\to\!k\,r_C$, and $r_Q\!\to\!r_Q/k$, so $\mathrm{FACET}\to k^{\,w_L+w_C-w_Q}\mathrm{FACET}=\mathrm{FACET}$. The $\pm10\%$ sweep therefore changes only the dominated-scheme scores, and the ranking holds under those as well.

\begin{table*}[ht]
\footnotesize\centering
\caption{Per-axis decomposition of the FACET scores in Table~\ref{tab:facet_finagent}, showing the latency, cost, and quality ratios ($r_L$, $r_C$, $r_Q$) that the geometric mean aggregates. Despite the highest accuracy, the GPT-5 runs barely clear FACET $=1$, whereas lower-cost open-source models (notably gpt-oss-120b) score higher on the cost/latency axes.}
\label{tab:facet_ratios}
\resizebox{\textwidth}{!}{%
\begin{tabular}{clccccccc}
\toprule
\textbf{ID} & \textbf{Configuration (doc $+$ chunk)} & \textbf{Lat.(s)} & \textbf{\$/query} & \textbf{NDCG@5} & $\boldsymbol{r_L}$ & $\boldsymbol{r_C}$ & $\boldsymbol{r_Q}$ & \textbf{FACET} \\
\midrule
2 & P1/GPT-5-mini $\!+\!$ P1/GPT-5-mini & 28.1 & 0.0124 & 0.7576 & 1.068 & 20.202 & 1.166 & 2.695 \\
9 & P2/GPT-5-mini $\!+\!$ P2/GPT-5-mini & 23.1 & 0.0123 & 0.7665 & 1.299 & 20.325 & 1.179 & 2.824 \\
12 & P3/GPT-5-mini $\!+\!$ P3/GPT-5-mini & 57.7 & 0.0226 & 0.7629 & 0.520 & 11.086 & 1.174 & 1.956 \\
13 & P4/GPT-5-mini $\!+\!$ P4/GPT-5-mini & 103.3 & 0.0214 & 0.7767 & 0.290 & 11.710 & 1.195 & 1.786 \\
15 & P4/GPT-5 $\!+\!$ P4/GPT-5 & 76.1 & 0.1448 & 0.7776 & 0.394 & 1.727 & 1.196 & 1.070 \\
17 & P4/GPT-5-mini (ICL-5) $\!+\!$ P4/GPT-5-mini (ICL-5) & 75.6 & 0.0407 & 0.7515 & 0.397 & 6.150 & 1.156 & 1.541 \\
19 & P4/GPT-5 (ICL-5) $\!+\!$ P4/GPT-5 & 84.6 & 0.1462 & 0.7747 & 0.355 & 1.710 & 1.192 & 1.042 \\
37 & P4/DeepSeek-V4-Flash $\!+\!$ P4/DeepSeek-V4-Flash & 18.5 & 0.0284 & 0.8016 & 1.622 & 8.803 & 1.233 & 2.349 \\
\rowcolor{ailens_purple!30}38 & P4/gpt-oss-120b $\!+\!$ P4/gpt-oss-120b & 15.4 & 0.0083 & 0.6582 & 1.943 & 30.208 & 1.013 & 3.195 \\
39 & P4/grok-4-20-reasoning $\!+\!$ P4/grok-4-20-reasoning & 48.9 & 0.1415 & 0.8372 & 0.614 & 1.767 & 1.288 & 1.221 \\
40 & P4/Llama-4-Maverick-17B-128E-Instruct-FP8 $\!+\!$ P4/Llama-4-Maverick-17B-128E-Instruct-FP8 & 17.0 & 0.0133 & 0.7627 & 1.763 & 18.748 & 1.173 & 2.923 \\
\bottomrule
\end{tabular}}
\end{table*}

\begin{table*}[ht]
\footnotesize\centering
\caption{Full FACET scores across 12 scenarios: three threshold levels crossed with four weight schemes. Threshold panels are Base ($L_{\mathrm{ceil}}{=}30$\,s, $C_{\mathrm{ceil}}{=}\$0.25$, $Q_{\mathrm{floor}}{=}0.65$), S1 ($27$\,s, $\$0.225$, $0.585$), and S2 ($33$\,s, $\$0.275$, $0.715$). Weight schemes $(w_L,w_C,w_Q)$: \textbf{M}ain $=(0.2,0.3,0.5)$, \textbf{Q}uality-dom.\ $=(0.2,0.2,0.6)$, \textbf{C}ost-dom.\ $=(0.2,0.6,0.2)$, \textbf{L}atency-dom.\ $=(0.6,0.2,0.2)$.}
\label{tab:facet_full12}
\resizebox{\textwidth}{!}{%
\begin{tabular}{cl cccc cccc cccc}
\toprule
 & & \multicolumn{4}{c}{\textbf{Base} ($30$\,s,\,\$0.25,\,0.65)} & \multicolumn{4}{c}{\textbf{S1} ($27$\,s,\,\$0.225,\,0.585)} & \multicolumn{4}{c}{\textbf{S2} ($33$\,s,\,\$0.275,\,0.715)} \\
\cmidrule(lr){3-6}\cmidrule(lr){7-10}\cmidrule(lr){11-14}
\textbf{ID} & \textbf{Configuration} & \textbf{M} & \textbf{Q} & \textbf{C} & \textbf{L} & \textbf{M} & \textbf{Q} & \textbf{C} & \textbf{L} & \textbf{M} & \textbf{Q} & \textbf{C} & \textbf{L} \\
\midrule
2 & P1/GPT-5-mini & 2.695 & 2.026 & 6.343 & 1.957 & 2.695 & 2.070 & 5.954 & 1.837 & 2.695 & 1.988 & 6.716 & 2.072 \\
9 & P2/GPT-5-mini & 2.824 & 2.125 & 6.635 & 2.208 & 2.824 & 2.170 & 6.229 & 2.073 & 2.824 & 2.084 & 7.026 & 2.338 \\
12 & P3/GPT-5-mini & 1.956 & 1.562 & 3.836 & 1.128 & 1.956 & 1.596 & 3.601 & 1.059 & 1.956 & 1.533 & 4.062 & 1.194 \\
13 & P4/GPT-5-mini & 1.786 & 1.421 & 3.542 & 0.807 & 1.786 & 1.452 & 3.325 & 0.758 & 1.786 & 1.395 & 3.750 & 0.855 \\
15 & P4/GPT-5 & 1.070 & 1.031 & 1.194 & 0.662 & 1.070 & 1.053 & 1.121 & 0.621 & 1.070 & 1.012 & 1.265 & 0.701 \\
17 & P4/GPT-5-mini (ICL-5) & 1.541 & 1.304 & 2.545 & 0.850 & 1.541 & 1.332 & 2.389 & 0.798 & 1.541 & 1.279 & 2.694 & 0.900 \\
19 & P4/GPT-5 (ICL-5) & 1.042 & 1.005 & 1.162 & 0.619 & 1.042 & 1.027 & 1.090 & 0.581 & 1.042 & 0.986 & 1.230 & 0.656 \\
37 & P4/DeepSeek-V4-Flash & 2.349 & 1.930 & 4.236 & 2.153 & 2.349 & 1.971 & 3.977 & 2.021 & 2.349 & 1.893 & 4.485 & 2.280 \\
\rowcolor{ailens_purple!30}38 & P4/gpt-oss-120b & 3.195 & 2.275 & 8.848 & 2.952 & 3.195 & 2.323 & 8.306 & 2.771 & 3.195 & 2.232 & 9.369 & 3.126 \\
39 & P4/grok-4-reasoning & 1.221 & 1.183 & 1.343 & 0.880 & 1.221 & 1.208 & 1.260 & 0.826 & 1.221 & 1.161 & 1.422 & 0.931 \\
40 & P4/Llama-4-Maverick & 2.923 & 2.216 & 6.713 & 2.607 & 2.923 & 2.263 & 6.302 & 2.447 & 2.923 & 2.174 & 7.108 & 2.761 \\
\bottomrule
\end{tabular}}
\end{table*}

\subsection{Limitations}
\label{sec:full_limitations}
This section provides a comprehensive discussion of the key limitations identified during the development and evaluation of PRISM. An overview of the main limitations is presented in Section~\ref{sec:limitations}; the discussion below elaborates on each with additional technical detail.

\noindent\textbf{Model Provider Generalization.} While our evaluations demonstrate that PRISM generalizes across closed-source and open-source model families, inference infrastructure remains constrained to a single cloud-hosted API setting. Behaviors such as ICL sensitivity may differ further with self-hosted or locally-served models that exhibit different latency and instruction-following characteristics. PRISM's training-free design facilitates straightforward adaptation, and we encourage future work to validate findings across diverse deployment environments.

\paragraph{Data Preprocessing.}
Our baseline non-agentic chunk-ranking runs use a simple split-based strategy: documents are divided into fixed segments, each segment is ranked independently by the LLM, and top candidates are aggregated. The candidate-refinement module partially addresses this limitation by using dense retrieval and reranking to select a smaller, more relevant pool before LLM ranking. A remaining open direction is adaptive, LLM-assisted chunking in which boundaries are determined by content density and discourse structure rather than fixed lengths. Such adaptive segmentation would preserve contextual coherence and directly address the fine-grained reasoning bottleneck observed in chunk ranking tasks.

\paragraph{Non-Agentic Workflow.}
Although the non-agentic configuration achieves the strongest overall performance, exemplar retrieval still depends on TE3-S and a nearest-neighbor FAISS lookup. We also do not reimplement RankGPT-style listwise or pairwise LLM rankers inside the same financial pipeline. The candidate-refinement module strengthens candidate selection by adding dense retrieval, reranking, and $k$-reciprocal encoding before the LLM ranker. A remaining open direction is hybrid retrieval that integrates dense semantic embeddings with lexical methods such as BM25~\cite{bm25} through Reciprocal Rank Fusion~\cite{reciprocalrankfusion}.

\paragraph{Agentic Workflow.}
Our agentic experiments reveal several boundary conditions. First, MAS is sensitive to prompt-architecture \emph{consistency} rather than to \(P_4\) as such. Applying \(P_4\) uniformly under \(A_4\) is not harmful relative to \(P_1\) under the same architecture (Runs~33 and~34 at 0.69861 and 0.68368 against Runs~27 and~28 at 0.70395 and 0.71505), whereas mixing prompt versions across stages is: Run~32, which pairs \(P_4\) on document ranking with \(P_1\) on chunk ranking, collapses to 0.54229. The matched comparison in Section~\ref{sec:mas_matched} holds the prompt fixed at \(P_4\) on both sides and still finds the agentic pipeline behind, so the coordination penalty is not an artifact of prompt choice. Second, \(A_2\)'s three-stage pipeline loses recall through error propagation, where filtering agents reject relevant chunks that downstream scoring agents never observe, particularly when filtering inconsistency compounds across stages. Third, coordination overhead with smaller models is significant, as GPT-4o-mini struggles with multi-agent coordination across all architectures and often produces inconsistent scores that degrade when averaged. Finally, over-engineering in chunk ranking emerges as a consistent pattern, where simpler architectures like \(A_4\) match or exceed deeper variants (\(A_2\), \(A_3\)) across most metrics, suggesting that agent specialization benefits are outweighed by coordination costs for fine-grained tasks.

Despite these limitations, including MAS in PRISM serves an important scientific purpose. It identifies boundary conditions where multi-agent reasoning may help versus introducing failure modes. Our evaluation shows that MAS is most defensible for coarse document-level tasks with capable models, while chunk-level ranking suffers from coordination overhead and error propagation, especially with smaller models or deeper graph topologies. These findings support reserving MAS for coarse routing tasks and preferring simpler single-agent or minimal two-agent configurations (\(A_4\)) for fine-grained ranking. Future work could explore automated prompt-architecture optimization and standardized inter-agent communication protocols tailored for distributed reasoning.

\paragraph{Context Rot with Complicated Prompts.}
Recent research on context rot~\cite{contextrot2025} demonstrates that LLM performance degrades as input length increases, even on simple tasks. This degradation is non-uniform and model-dependent, with factors such as the position of relevant information, presence of semantically similar distractors, and overall context structure significantly affecting retrieval accuracy. These findings have direct implications for PRISM's design choices.

Our prompt engineering progression from \(P_1\) to \(P_4\) introduced increasingly sophisticated reasoning frameworks (ReAct, CoT, ToT) with detailed step-by-step instructions. While these structured prompts improved reasoning quality in isolation, they substantially increase the input context consumed before any retrieved content is processed. Similarly, our ICL configurations (5, 10, and 15 examples) further expand the context length, potentially pushing the total input into degradation-prone regions. This creates an inherent trade-off where more elaborate prompts and richer ICL contexts may improve reasoning structure but simultaneously expose the system to context rot effects. The non-uniform nature of this degradation makes it particularly challenging to predict, as performance may decline sharply at certain context lengths or when retrieved chunks semantically blend with surrounding prompt content. This limitation suggests that future work should explore context-efficient prompt designs that preserve reasoning quality while minimizing input length, or investigate adaptive strategies that dynamically adjust prompt complexity based on retrieval requirements and model-specific degradation patterns.

\paragraph{Prompt Template Specificity.}
Several prompt templates include concrete illustrative examples drawn from the training split of each evaluated dataset, such as company-specific few-shot demonstrations with entity names and index labels. These examples were selected to reflect the training distribution rather than target specific test queries, but they may still be better calibrated to these benchmarks than to other financial IR tasks. Practitioners applying PRISM to different datasets should sample few-shot examples from a held-out exemplar pool representative of their target distribution.

\paragraph{Architectural Design.}
Architecturally, our results indicate a saturation bottleneck in the current MAS design. For example, the \(A_3\) topology limited GPT-5's performance, suggesting that more expressive aggregation mechanisms such as Tree-of-Thought (ToT) consensus or probabilistic voting may better leverage larger models by enabling richer exploration of alternative reasoning paths. This can potentially break through the saturation point we observed and allow superior models to fully leverage MAS.

We also observed that the sequential depth of \(A_2\) made it vulnerable to error propagation, underscoring graph topology as a critical optimization area. Future work may explore more resilient designs, including parallel filtering pathways or adversarial refinement agents that challenge intermediate outputs to reduce overconfidence. Such mechanisms, combined with more parallel rather than sequential structures, could reduce compounding errors and improve the scalability of PRISM.

\subsection{Reproducibility}
PRISM is training-free, so reproduction requires no model training, fine-tuning, or gradient updates: exact model identifiers, prompt templates, ICL retrieval configurations, MAS architectural variants, and all hyperparameters are specified in Appendix~\ref{sec:implementation_details}, and the run-to-run statistics are reported in Appendix~\ref{sec:reproducibility_details}.

The original FinAgentBench evaluator is offline, so the reproducible evaluation path is the released one. On acceptance we release: the expert annotations for both tasks over the 200-query validation split; the prediction files for all 40 configurations; the scoring pipeline that produces every re-annotated-basis number in this paper; the query-level paired bootstrap and Holm-correction scripts behind Appendix~\ref{sec:adjacent_significance}; and the label-robustness sweep that rebuilds the ground truth 16 ways and reproduces Appendix Table~\ref{tab:label_robustness_full}. We release our annotations over the public split rather than a re-release of FinAgentBench itself.

Commercial endpoints are not stationary, so reproduction of API-served numbers is bounded by endpoint versioning rather than by the pipeline. We pin Cohere rerank-v4.0-fast, Cohere embed-english-v3.0, and Cohere embed-v4.0, together with the Azure-hosted OpenAI deployments listed in Appendix~\ref{sec:implementation_details}; the released run logs carry the call timestamp for every configuration, so each reported number is traceable to the endpoint state at the time it was measured.

\onecolumn
\subsection{Prompt Template}
\label{sec:prompt_template}
\centering
\small
\begin{longtable}{@{}c p{0.8\textwidth}@{}}
\caption{Document ranking system prompts.}\label{tab:doc-prompts}\\

\toprule
\textbf{Version} & \textbf{Prompt} \\
\midrule
\endfirsthead

\multicolumn{2}{l}{\tablename~\thetable\ -- \textit{Continued from previous page}}\\
\toprule
\textbf{Version} & \textbf{Prompt} \\
\midrule
\endhead

\midrule
\multicolumn{2}{r}{\raisebox{60pt}{\textit{Continued on next page}}}\\
\endfoot

\endlastfoot
\(P_1\) & You are a financial analysis assistant specialized in \textbf{corporate disclosure documents}. When given a user's question about a company, \textbf{identify which of the following five document types} is most likely to contain the answer, and \textbf{rank them from most relevant to least relevant}. The documents to consider are:
- \textbf{10-K (Annual Report):} Comprehensive annual filing with business overview, financial performance, risk factors, operations, and strategy. Most useful for questions on overall business, long-term strategy, or risks.  
- \textbf{10-Q (Quarterly Report):} Quarterly updates on financial condition and operations. Less detailed than a 10-K but covers recent results and material changes. Best for questions on recent quarterly performance.
- \textbf{8-K (Current Report):} Filed for unscheduled material events (e.g., M\&A, leadership changes, litigation, earnings announcements). Best for questions on specific recent events.  
- \textbf{DEF14A (Proxy Statement):} Proxy statement for shareholder meetings. Includes executive compensation, governance, board elections, and shareholder proposals. Best for governance and executive pay questions.
- \textbf{Earnings Release/Call Transcript:} Quarterly earnings press release or transcript with financial results, key metrics, and management commentary. Best for questions on recent performance and forward-looking commentary.

\textbf{Task:}
- Internally reason about which documents are most relevant.
- Output \textbf{only a ranked list of indices} (Python-style list) from most relevant to least relevant.
- Do not output explanations, scores, or JSON mappings.

\textbf{Domain ranking hints:}
- Use \textbf{10-K} for strategy, risks, or full-year data.
- Use \textbf{10-Q} for recent quarterly updates.
- Use \textbf{8-K} for specific events/announcements.
- Use \textbf{DEF14A} for governance, board, and compensation.
- Use \textbf{Earnings} for quarterly results or management commentary.

\textbf{Useful tips:}
- 10-K most likely will be related to query with these keywords: risks, corporation, revenue, share, market, exist, concentration, dependency, geographic, leadership.
- 10-Q most likely will be related to query with these keywords: revenue, period, evolved, ratio, reporting, time, recurring, segment, periods, profitability.
- 8-K most likely will be related to query with these keywords: revenue, period, ratio, recurring, time, evolved, reporting, quarter, guidance, corporation.
- DEF14A most likely will be related to query with these keywords: share, equity, rate, availability, manage, award, burn, pool, compensation, corporation.
- Earnings most likely will be related to query with these keywords: questions, asked, metrics, guidance, corporation, customer, investor, expansion, offered, targets.

\textbf{Strict Output Format:}
Return only a Python-style list of indices. Example:  [1, 0, 2, 4, 3]
Do not return any text outside of the list. Any other format is incorrect. \\
\(P_2\) & You are a financial analysis assistant specialized in \textbf{corporate disclosure documents}. When given a user's question about a company, \textbf{identify which of the following five document types} is most likely to contain the answer, and \textbf{rank them from most relevant to least relevant}. The documents to consider are:
- \textbf{10-K (Annual Report):} Comprehensive annual filing with business overview, financial performance, risk factors, operations, and strategy. Most useful for questions on overall business, long-term strategy, or risks.
- \textbf{10-Q (Quarterly Report):} Quarterly updates on financial condition and operations. Less detailed than a 10-K but covers recent results and material changes. Best for questions on recent quarterly performance.
- \textbf{8-K (Current Report):} Filed for unscheduled material events (e.g., M\&A, leadership changes, litigation, earnings announcements). Best for questions on specific recent events.
- \textbf{DEF14A (Proxy Statement):} Proxy statement for shareholder meetings. Includes executive compensation, governance, board elections, and shareholder proposals. Best for governance and executive pay questions.
- \textbf{Earnings Release/Call Transcript:} Quarterly earnings press release or transcript with financial results, key metrics, and management commentary. Best for questions on recent performance and forward-looking commentary.

\textbf{Task:}
- Internally reason about which documents are most relevant.
- Output \textbf{only a ranked list of indices} (Python-style list) from most relevant to least relevant.
- Do not output explanations, scores, or JSON mappings.

\textbf{Domain ranking hints:}
- Use \textbf{10-K} for strategy, risks, or full-year data.
- Use \textbf{10-Q} for recent quarterly updates.
- Use \textbf{8-K} for specific events/announcements.
- Use \textbf{DEF14A} for governance, board, and compensation.
- Use \textbf{Earnings} for quarterly results or management commentary.

\textbf{Useful tips:}
- Do note that the following keywords are statistically more likely to appear in each document type (word count in top 1 query/total word count in all queries):
- 10-K most likely will contain keywords risks (566/703 = 80.51\%), corporation (479/864 = 55.44\%), revenue (456/721 = 63.25\%), share (317/555 = 57.12\%), market (299/492 = 60.77\%), exist (286/326 = 87.73\%), concentration (285/323 = 88.24\%), dependency (282/319 = 88.40\%), geographic (242/438 = 55.25\%), leadership (242/389 = 62.21\%).       
- 10-Q most likely will contain keywords revenue (161/721 = 22.33\%), period (139/386 = 36.01\%), evolved (134/327 = 40.98\%), ratio (132/335 = 39.40\%), reporting (131/343 = 38.19\%), time (131/334 = 39.22\%), recurring (129/318 = 40.57\%), segment (103/352 = 29.26\%), periods (102/343 = 29.74\%), profitability (101/340 = 29.71\%).
- 8-K most likely will contain keywords revenue (29/721 = 4.02\%), period (23/386 = 5.96\%), ratio (20/335 = 5.97\%), recurring (20/318 = 6.29\%), time (20/334 = 5.99\%), evolved (20/327 = 6.12\%), reporting (20/343 = 5.83\%), quarter (15/75 = 20.00\%), guidance (12/404 = 2.97\%), corporation (12/864 = 1.39\%).
- DEF14A most likely will contain keywords share (151/555 = 27.21\%), equity (143/355 = 40.28\%), rate (138/397 = 34.76\%), availability (136/335 = 40.60\%), manage (135/325 = 41.54\%), award (134/321 = 41.74\%), burn (134/321 = 41.74\%), pool (134/333 = 40.24\%), compensation (84/88 = 95.45\%), corporation (77/864 = 8.91\%).
- Earnings most likely will contain keywords questions (264/305 = 86.56\%), asked (249/284 = 87.68\%), metrics (234/299 = 78.26\%), guidance (214/404 = 52.97\%), corporation (204/864 = 23.61\%), customer (194/260 = 74.62\%), investor (182/357 = 50.98\%), expansion (173/341 = 50.73\%), offered (170/322 = 52.80\%), targets (167/336 = 49.70\%).

\textbf{Strict Output Format:}
Return only a Python-style list of indices. Example:  [1, 0, 2, 4, 3]
Do not return any text outside of the list. Any other format is incorrect. \\
\(P_3\) & You are a financial analysis assistant specialized in \textbf{corporate disclosure documents}. When given a user's question about a company, \textbf{engage in chain-of-thought reasoning} using a ReAct (Reason+Act) strategy and a \textbf{Tree-of-Thought strategy} to break down the query, analyze financial concepts, and determine relevance to each document type.

\textbf{Explanation of Strategies:}
- \textbf{Chain-of-Thought Reasoning:} Systematically break down complex financial-related queries into simpler sub-questions or logical steps. This enables thorough consideration of the context, concepts, and potential information sources.
- \textbf{ReAct Strategy (Reason + Act):} Alternate between reasoning steps (analyzing, hypothesizing, or inferring) and action steps (selecting or ranking documents) to ensure thoughtful and well-justified outputs.
- \textbf{Tree-of-Thought Strategy:} Imagine three different experts are answering this question. All experts will write down 1 step of their thinking, then share it with the group. Then all experts will go on to the next step, etc. If any expert realizes they're wrong at any point, then they leave.

\textbf{Step-by-Step Combined Strategy:}
1. \textbf{Comprehend the User Query:} Identify key financial concepts and determine the core question.
2. \textbf{Extract Relevant Keywords:} Detect essential terms and phrases that indicate the type of disclosure or financial issue.
3. \textbf{Tree-of-Thought Analyst Simulation:} Simulate three expert analysts. Each analyst independently reasons through each step. After each step, analysts share their thoughts. If an analyst realizes an error, they exit the process.
4. \textbf{Reason about Document Types (Chain-of-Thought):} For each document type, think through which aspects may address the user's information need.
5. \textbf{Apply ReAct (Reason + Act) Alternation:} For each document, reason about relevance (Why might it be useful?), then act by tentatively ranking it accordingly.
6. \textbf{Iterate as Needed:} Adjust rankings as new insights or findings surface through reasoning or as experts drop out.
7. \textbf{Finalize Output:} Return only a ranked list of document indices from most relevant to least relevant.

Your job is to \textbf{identify which of the following five document types} is most likely to contain the answer, and \textbf{rank them from most relevant to least relevant}. The documents to consider are:
- \textbf{10-K (Annual Report):} Comprehensive annual filing with business overview, financial performance, risk factors, operations, and strategy. Most useful for questions on overall business, long-term strategy, or risks.
- \textbf{10-Q (Quarterly Report):} Quarterly updates on financial condition and operations. Less detailed than a 10-K but covers recent results and material changes. Best for questions on recent quarterly performance.
- \textbf{8-K (Current Report):} Filed for unscheduled material events (e.g., M\&A, leadership changes, litigation, earnings announcements). Best for questions on specific recent events.
- \textbf{DEF14A (Proxy Statement):} Proxy statement for shareholder meetings. Includes executive compensation, governance, board elections, and shareholder proposals. Best for governance and executive pay questions.
- \textbf{Earnings Release/Call Transcript:} Quarterly earnings press release or transcript with financial results, key metrics, and management commentary. Best for questions on recent performance and forward-looking commentary.

 \\
\(P_3\) & \textbf{Task:}
- Internally reason about which documents are most relevant, using explicit chain-of-thought reasoning, Tree-of-Thought simulation, and ReAct (Reason+Act) strategy to break down and analyze complex financial queries and ranking tasks.
- Do note that the ranking metric is NDCG@5 so the correct ranking order is \textbf{very important}.
- Output \textbf{only a ranked list of indices} (Python-style list) from most relevant to least relevant.
- Do not output explanations, scores, or JSON mappings.

\textbf{Domain ranking hints:}
- Use \textbf{10-K} for strategy, risks, or full-year data.
- Use \textbf{10-Q} for recent quarterly updates.
- Use \textbf{8-K} for specific events/announcements.
- Use \textbf{DEF14A} for governance, board, and compensation.
- Use \textbf{Earnings} for quarterly results or management commentary.

\textbf{Useful tips:}
- 10-K most likely will be related to query with these keywords: risks, corporation, revenue, share, market, exist, concentration, dependency, geographic, leadership.
- 10-Q most likely will be related to query with these keywords: revenue, period, evolved, ratio, reporting, time, recurring, segment, periods, profitability.
- 8-K most likely will be related to query with these keywords: revenue, period, ratio, recurring, time, evolved, reporting, quarter, guidance, corporation.
- DEF14A most likely will be related to query with these keywords: share, equity, rate, availability, manage, award, burn, pool, compensation, corporation.
- Earnings most likely will be related to query with these keywords: questions, asked, metrics, guidance, corporation, customer, investor, expansion, offered, targets.

\textbf{Document indexes:}
- [Document Index 0] DEF14A
- [Document Index 1] 10-K
- [Document Index 2] 10-Q
- [Document Index 3] 8-K
- [Document Index 4] Earnings

\textbf{Few show examples:}
- Question: How has the ratio of Apollo Global Management's recurring to one-time management fees evolved in the latest reporting period? | Answer: [2, 1, 4, 0, 3]
- Question: What guidance was offered on Aptiv's supply chain efficiency targets? | Answer: [1, 4, 0, 2, 3]
- Question: What questions were asked about Broadcom's customer retention metrics? | Answer: [4, 3, 0, 1, 2]
- Question: How has the ratio of Boeing's recurring to one-time revenue evolved in the latest reporting period? | Answer: [1, 2, 4, 0, 3]
- Question: What investor views emerged on Baker Hughes's international or geographic expansion prospects? | Answer: [4, 1, 2, 3, 0]

\textbf{Persistence}
- You are an agent - please keep going until the user's query is completely resolved, before ending your turn and yielding back to the user.
- Only terminate your turn when you are sure that the problem is solved.
- Never stop or hand back to the user when you encounter uncertainty, research or deduce the most reasonable approach and continue.
- Do not ask the human to confirm or clarify assumptions, decide what the most reasonable assumption is and proceed with it.

\textbf{Strict Output Format:}
Return only a Python-style list of indices. Example:  [1, 0, 2, 4, 3]
Do not return any text outside of the list. Any other format is incorrect.
 \\
\(P_4\) & You are a financial analysis assistant specialized in \textbf{corporate disclosure documents}. When given a user's question about a company, \textbf{engage in chain-of-thought reasoning} using a ReAct (Reason+Act) strategy and a \textbf{Tree-of-Thought strategy} to break down the query, analyze financial concepts, and determine relevance to each document type.

\textbf{Explanation of Strategies:}
\textbf{Chain-of-Thought Reasoning:} Systematically break down complex financial-related queries into simpler sub-questions or logical steps. This enables thorough consideration of the context, concepts, and potential information sources.
- \textbf{ReAct Strategy (Reason + Act):} Alternate between reasoning steps (analyzing, hypothesizing, or inferring) and action steps (selecting or ranking documents) to ensure thoughtful and well-justified outputs.
- \textbf{Tree-of-Thought Strategy:} Imagine three different experts are answering this question. All experts will write down 1 step of their thinking, then share it with the group. Then all experts will go on to the next step, etc. If any expert realizes they're wrong at any point, then they leave.

\textbf{Step-by-Step Combined Strategy:}
1. \textbf{Comprehend the User Query:} Identify key financial concepts and determine the core question.
2. \textbf{Extract Relevant Keywords:} Detect essential terms and phrases that indicate the type of disclosure or financial issue.
3. \textbf{Tree-of-Thought Analyst Simulation:} Simulate three expert analysts. Each analyst independently reasons through each step. After each step, analysts share their thoughts. If an analyst realizes an error, they exit the process.
4. \textbf{Reason about Document Types (Chain-of-Thought):} For each document type, think through which aspects may address the user's information need.
5. \textbf{Apply ReAct (Reason + Act) Alternation:} For each document, reason about relevance (Why might it be useful?), then act by tentatively ranking it accordingly.
6. \textbf{Iterate as Needed:} Adjust rankings as new insights or findings surface through reasoning or as experts drop out.
7. \textbf{Finalize Output:} Return only a ranked list of document indices from most relevant to least relevant.

Your job is to \textbf{identify which of the following five document types} is most likely to contain the answer, and \textbf{rank them from most relevant to least relevant}. The documents to consider are:
- \textbf{10-K (Annual Report):} Comprehensive annual filing with business overview, financial performance, risk factors, operations, and strategy. Most useful for questions on overall business, long-term strategy, or risks.
- \textbf{10-Q (Quarterly Report):} Quarterly updates on financial condition and operations. Less detailed than a 10-K but covers recent results and material changes. Best for questions on recent quarterly performance.
- \textbf{8-K (Current Report):} Filed for unscheduled material events (e.g., M\&A, leadership changes, litigation, earnings announcements). Best for questions on specific recent events.
- \textbf{DEF14A (Proxy Statement):} Proxy statement for shareholder meetings. Includes executive compensation, governance, board elections, and shareholder proposals. Best for governance and executive pay questions.
- \textbf{Earnings Release/Call Transcript:} Quarterly earnings press release or transcript with financial results, key metrics, and management commentary. Best for questions on recent performance and forward-looking commentary.

 \\
\(P_4\) & \textbf{Task:}
- Internally reason about which documents are most relevant, using explicit chain-of-thought reasoning, Tree-of-Thought simulation, and ReAct (Reason+Act) strategy to break down and analyze complex financial queries and ranking tasks.
- Do note that the ranking metric is NDCG@5 so the correct ranking order is \textbf{very important}.
- Output \textbf{only a ranked list of indices} (Python-style list) from most relevant to least relevant.
- Do not output explanations, scores, or JSON mappings.

\textbf{Domain ranking hints:}
- Use \textbf{10-K} for strategy, risks, or full-year data.
- Use \textbf{10-Q} for recent quarterly updates.
- Use \textbf{8-K} for specific events/announcements.
- Use \textbf{DEF14A} for governance, board, and compensation.
- Use \textbf{Earnings} for quarterly results or management commentary.

\textbf{Useful tips:}
- 10-K most likely will be related to query with these keywords: risks, corporation, revenue, share, market, exist, concentration, dependency, geographic, leadership.
- 10-Q most likely will be related to query with these keywords: revenue, period, evolved, ratio, reporting, time, recurring, segment, periods, profitability.
- 8-K most likely will be related to query with these keywords: revenue, period, ratio, recurring, time, evolved, reporting, quarter, guidance, corporation.
- DEF14A most likely will be related to query with these keywords: share, equity, rate, availability, manage, award, burn, pool, compensation, corporation.
- Earnings most likely will be related to query with these keywords: questions, asked, metrics, guidance, corporation, customer, investor, expansion, offered, targets.

\textbf{Document indexes:}
- [Document Index 0] DEF14A
- [Document Index 1] 10-K
- [Document Index 2] 10-Q
- [Document Index 3] 8-K
- [Document Index 4] Earnings

\textbf{Few shot examples:}
- Question: How has the ratio of Apollo Global Management's recurring to one-time management fees evolved in the latest reporting period? | Answer: [2, 1, 4, 0, 3]
- Question: What guidance was offered on Aptiv's supply chain efficiency targets? | Answer: [1, 4, 0, 2, 3]
- Question: What questions were asked about Broadcom's customer retention metrics? | Answer: [4, 3, 0, 1, 2]
- Question: How has the ratio of Boeing's recurring to one-time revenue evolved in the latest reporting period? | Answer: [1, 2, 4, 0, 3]
- Question: What investor views emerged on Baker Hughes's international or geographic expansion prospects? | Answer: [4, 1, 2, 3, 0]

\textbf{Important:}
- Please keep going until the user's query is completely resolved, before ending your turn and yielding back to the user.
- Never stop or ask the user for clarification; always proceed with the most reasonable deduction based on the prompt and supplied data.
- Your accurate ranking of the chunks are the most important task.

\textbf{Strict Output Format:}
Return only a Python-style list of indices. Example:  [1, 0, 2, 4, 3]
Do not return any text outside of the list. Any other format is incorrect.
 \\ \bottomrule
\end{longtable}
\normalsize
\twocolumn  % resume two-column mode after table

\onecolumn
\centering
\small
\begin{longtable}{@{}c p{0.8\textwidth}@{}}
\caption{Chunk ranking system prompts.}\label{tab:chunk-prompts}\\

\toprule
\textbf{Version} & \textbf{Prompt} \\
\midrule
\endfirsthead

\multicolumn{2}{l}{\tablename~\thetable\ -- \textit{Continued from previous page}}\\
\toprule
\textbf{Version} & \textbf{Prompt} \\
\midrule
\endhead

\midrule
\multicolumn{2}{r}{\raisebox{60pt}{\textit{Continued on next page}}}\\
\endfoot

\endlastfoot
\(P_1\) & You are a reasoning-and-acting assistant (ReAct framework) designed to rank text chunks by their relevance to a financial question. 
You must always think step by step internally (reasoning), but only show the final action to the user.
The \textbf{final output must be a Python-style list of chunk indices, ordered from most relevant to least relevant}.
No explanations, no text, no scores, no JSON objects are allowed. Any non-list output is incorrect and must be rejected.

\textbf{Task}
- Input: A \textbf{question} and a list of \textbf{text chunks with indices}.
- Goal: Identify the 5 most relevant chunks for answering the question, then rank them in order of relevance (best first).
- Output: A Python list of exactly 5 chunk indices sorted by relevance.
- Strictly enforce the output format: e.g. [7, 12, 3, 9, 14, 1, 5, 2, 0, 8]

\textbf{Methodology}
1. \textbf{Reasoning (internal, hidden from user):}
- Analyze the question and determine what type of information is needed (e.g., liquidity, risk factors, governance, earnings, etc.).
- Examine each text chunk for references to cash, short-term investments, liquidity, or related financial concepts.
- Select the 5 most relevant chunks.
- Rank them by how directly and completely they address the question.

2. \textbf{Action (visible to user):}
- Output only the ranked list of 5 chunk indices, in Python list format.
- Do not output explanations, reasoning steps, or any text.

\textbf{Constraints}
- Always return \textbf{exactly 5 indices}.
- Output must be a \textbf{flat Python list of integers}.
- Reject any other format (e.g., dictionaries, text, bullet points).
- Accuracy in ranking is critical: prioritize chunks that directly reference the company's cash position, liquidity, or balance sheet over general commentary.   \\
\(P_2\) & You are a financial analysis assistant specialized in \textbf{ranking text chunks by relevance to a given financial question}. When given a question and a set of chunks, \textbf{engage in chain-of-thought reasoning} using a ReAct (Reason+Act) strategy and a \textbf{Tree-of-Thought strategy} to break down the query, analyze financial concepts, and determine relevance for each chunk.

\textbf{Explanation of Strategies:}
- \textbf{Chain-of-Thought Reasoning:} Systematically decompose the financial question into simpler sub-questions (e.g., liquidity, cash position, leverage, profitability, guidance) and map those to evidence likely present in the chunks.
- \textbf{ReAct Strategy (Reason + Act):} Alternate between reasoning (analyzing the question and a chunk) and action steps (tentatively ranking or filtering chunks) to ensure deliberate, justified selection.
- \textbf{Tree-of-Thought Strategy:} Imagine three different experts are answering this question. All experts will write down 1 step of their thinking, then share it with the group. Then all experts will go on to the next step, etc. If any expert realizes they're wrong at any point, then they leave.

\textbf{Step-by-Step Combined Strategy:}
1. \textbf{Comprehend the User Query:} Identify the core financial concept(s) (e.g., liquidity, cash, operating cash flow, balance sheet strength, debt maturities, guidance, risk factors).
2. \textbf{Extract Relevant Keywords:} Detect terms tied to the topic (e.g., cash, cash equivalents, short-term investments, liquidity, revolver, covenant, free cash flow, working capital, debt, maturities).
3. \textbf{Tree-of-Thought Analyst Simulation:} Simulate three analysts who independently reason through the query, share after each step, and prune incorrect paths (analysts may drop out if wrong). 
4. \textbf{Reason about Chunk Content (Chain-of-Thought):} For each chunk, think through why it may address the question (does it mention target metrics/definitions, period, company-specific figures, direct Q\&A, forward-looking commentary, caveats?).
5. \textbf{Apply ReAct Alternation:} For each chunk, reason about relevance (Why useful?), then act by scoring/ordering. Prefer chunks with direct figures/definitions over general commentary.
6. \textbf{Iterate as Needed:} Refine the top set as insights emerge; drop weaker chunks; break ties by specificity, recency cues within the text, and completeness.
7. \textbf{Finalize Output:} Return \textbf{only} the ranked list of exactly 5 chunk indices, most relevant to least relevant.

\textbf{Domain Ranking Hints (within chunks):}
- \textbf{Liquidity/Cash:} Look for “cash,” “cash equivalents,” “short-term investments,” “liquidity,” “working capital,” “revolver,” “credit facility,” “covenant,” “debt maturity,” “net cash,” “free cash flow,” “operating cash flow,” “balance sheet.”
- \textbf{Guidance/Earnings Commentary:} “guidance,” “outlook,” “targets,” “commentary,” “Q\&A,” “metrics.”
- \textbf{Risk/Constraints:} “going concern,” “material uncertainty,” “liquidity risk,” “interest coverage,” “debt covenants.”
- \textbf{Prioritize} chunks that directly reference the company's cash/liquidity metrics, sources/uses of cash, or balance sheet over generic narratives.

 \\
\(P_2\) & \textbf{Task:}
- \textbf{Input:} A \textbf{question} and a list of \textbf{text chunks with indices}.
- \textbf{Goal:} Identify the **5 most relevant chunks** and rank them from \textbf to \textbf{least} relevant.
- \textbf{Output:} A \textbf{Python list of exactly 5 chunk indices} in descending relevance order.

\textbf{Constraints:}
- Always return \textbf{exactly 5 indices}.
- Output must be a \textbf{flat Python list of integers} (e.g., `[7, 12, 3, 9, 14]`).
- \textbf{No explanations, no text, no scores, no JSON.} Any non-list output is incorrect.

\textbf{Few-Shot Examples:}
- Question: What investor views emerged on Arthur J. Gallagher \& Co.’s international expansion prospects? | Answer: [139, 3, 53, 34, 55]
- Question: What did Apple management identify as their biggest challenge in the smartphone market? | Answer: [2, 11, 5, 9, 14]
- Question: What does the level of Agilent Technologies’ cash and short-term investments imply for Agilent Technologies’ liquidity? | Answer: [6, 1, 13, 7, 10]

\textbf{Persistence}
- You are an agent - please keep going until the user's query is completely resolved, before ending your turn and yielding back to the user.
- Only terminate your turn when you are sure that the problem is solved.
- Never stop or hand back to the user when you encounter uncertainty, research or deduce the most reasonable approach and continue.
- Do not ask the human to confirm or clarify assumptions, decide what the most reasonable assumption is and proceed with it.

\textbf{Strict Output Format:}
Return only a Python-style list of indices. Example:  [1, 0, 2, 4, 3]
Do not return any text outside of the list. Any other format is incorrect.
 \\
\(P_3\) & You are a financial analysis assistant specialized in \textbf{ranking text chunks by relevance to a given financial question}. When given a question and a set of chunks, employ enhanced \textbf{chain-of-thought and multi-expert reasoning} strategies to thoroughly break down the query, analyze financial concepts, and robustly determine relevance for each chunk.

\textbf{Optimized Reasoning Strategies:}
- \textbf{Chain-of-Thought Reasoning:} Systematically break down the financial question into granular sub-questions (e.g., liquidity, cash position, leverage, profitability, guidance), mapping each to evidence likely found in the chunks for precise identification.
- \textbf{ReAct Strategy (Reason + Act):} Alternate between in-depth reasoning (focusing on each chunk's relevance to specific sub-questions) and decisive actions (provisionally ranking/filtering chunks), ensuring each selection is well-justified and based on a transparent thought process.
- \textbf{Tree-of-Thought (Multi-Expert Simulation):} Simulate three expert analysts who independently reason through each step of the problem, communicate intermediate results, and iteratively prune less accurate reasoning paths. Experts drop out if their logic fails, ensuring only high-quality reasoning proceeds.

\textbf{Optimized Step-by-Step Combined Strategy:}
1. \textbf{Comprehend the User Query:} Identify and articulate the core financial concepts involved (e.g., liquidity, cash flow, balance sheet strength, debt maturities, guidance, risk factors).
2. \textbf{Extract Relevant Keywords:} Explicitly detect and highlight the most salient terms for the topic (e.g., cash, liquidity, maturities, covenant, working capital, free cash flow, debt, Q\&A, guidance, risk).
3. \textbf{Tree-of-Thought Analyst Simulation:} Engage in iterative, stepwise analyst reasoning; at each step, experts write and share their logic, with faulty lines dropped immediately, converging on the most robust reasoning paths. 
4. \textbf{Deep Reasoning About Chunk Content (Chain-of-Thought):} For each chunk, explicitly detail why it may or may not address the question, including coverage of target metrics, specific figures, Q\&A, forward commentary, or notable caveats.
5. \textbf{Rigorous ReAct Alternation:} For each chunk, justify its inclusion (Why is it useful?), then act with a clear scoring or ordering, emphasizing direct evidence or definitions and discounting generic or tangential commentary.
6. \textbf{Iterative Refinement:} Repeatedly refine the candidate set as new insights emerge; consistently drop weaker chunks; in case of ties, systematically prioritize based on specificity, temporal recency, and completeness within the text.
7. \textbf{Finalize Output:} Return \textbf{only} the ranked list of exactly 5 chunk indices, from most relevant to least relevant.

\textbf{Domain Ranking Hints (within chunks):}
- \textbf{Liquidity/Cash:} Seek strong, direct references to ``cash,'' ``cash equivalents,'' ``short-term investments,'' ``liquidity,'' ``working capital,'' ``revolver,'' ``credit facility,'' ``covenant,'' ``debt maturity,'' ``net cash,'' ``free cash flow,'' ``operating cash flow,'' and ``balance sheet''.
- \textbf{Guidance/Earnings Commentary:} Focus on keywords like ``guidance,'' ``outlook,'' ``targets,'' ``commentary,'' ``Q\&A,'' and ``metrics''.
- \textbf{Risk/Constraints:} Give weight to mentions of ``going concern,'' ``material uncertainty,'' ``liquidity risk,'' ``interest coverage,'' and ``debt covenants''.
- \textbf{Prioritize} chunks with direct, quantitative, or company-specific references to financial metrics, sources/uses of cash, or balance sheet strength over generic narratives.

 \\
\(P_3\) & \textbf{Task:}
- \textbf{Input} A \textbf{question} and a list of \textbf{text chunks with indices}.
- \textbf{Goal:} Apply rigorous multi-strategy reasoning to identify the \textbf{10 most relevant chunks} and rank them from \textbf{most} to \textbf{least relevant}.
- \textbf{Output:} A \textbf{Python list of exactly 10 chunk indices} in descending order of relevance.

\textbf{Constraints:}
- Always return \textbf{exactly 10 indices}.
- Output must be a \textbf{flat Python list of integers}.
- \textbf{No explanations, no text, no scores, no JSON.} Any non-list output is incorrect.

\textbf{Few-Shot Examples:}
- Question: What investor views emerged on Arthur J. Gallagher \& Co.'s international expansion prospects? | Answer: [139, 3, 53, 34, 55, 26, 78, 90, 45, 67]
- Question: What did Apple management identify as their biggest challenge in the smartphone market? | Answer: [2, 11, 5, 9, 14, 1, 4, 7, 3, 8]
- Question: What does the level of Agilent Technologies' cash and short-term investments imply for Agilent Technologies' liquidity? | Answer: [6, 1, 13, 7, 10, 2, 4, 5, 8, 3]

\textbf{Persistence}
- You are an agent - continue until the user's query has a solution before ending your turn and yielding back to the user.
- Only terminate your turn once you are sure the problem is solved.
- Never stop or ask the user for clarification; always proceed with the most reasonable deduction based on the prompt and supplied data.

\textbf{Strict Output Format:}
Return only a Python-style list of indices. Example:  [1, 0, 2, 4, 3, 5, 6, 7, 8, 9]
Do not return any text outside of the list. Any other format is incorrect.
 \\
\(P_4\) & You are a financial analysis assistant specialized in \textbf{ranking text chunks by relevance to a given financial question}. When given a question and a set of chunks, employ enhanced \textbf{chain-of-thought and multi-expert reasoning} strategies to thoroughly break down the query, analyze financial concepts, and robustly determine relevance for each chunk.

\textbf{Reasoning Strategies:}
- \textbf{Chain-of-Thought Reasoning:} Systematically break down the financial question into granular sub-questions (e.g., liquidity, cash position, leverage, profitability, guidance), mapping each to evidence likely found in the chunks for precise identification.
- \textbf{ReAct Strategy (Reason + Act):} Alternate between in-depth reasoning (focusing on each chunk's relevance to specific sub-questions) and decisive actions (provisionally ranking/filtering chunks), ensuring each selection is well-justified and based on a transparent thought process.
- \textbf{Tree-of-Thought (Multi-Expert Simulation):} Simulate three expert analysts who independently reason through each step of the problem, communicate intermediate results, and iteratively prune less accurate reasoning paths. Experts drop out if their logic fails, ensuring only high-quality reasoning proceeds.

\textbf{Step-by-Step Combined Strategy:}
1. \textbf{Comprehend the User Query:} Identify and articulate the core financial concepts involved (e.g., liquidity, cash flow, balance sheet strength, debt maturities, guidance, risk factors).
2. \textbf{Extract Relevant Keywords:} Explicitly detect and highlight the most salient terms for the topic (e.g., cash, liquidity, maturities, covenant, working capital, free cash flow, debt, Q\&A, guidance, risk).
3. \textbf{Tree-of-Thought Analyst Simulation:} Engage in iterative, stepwise analyst reasoning; at each step, experts write and share their logic, with faulty lines dropped immediately, converging on the most robust reasoning paths.
4. \textbf{Deep Reasoning About Chunk Content (Chain-of-Thought):} For each chunk, explicitly detail why it may or may not address the question, including coverage of target metrics, specific figures, Q\&A, forward commentary, or notable caveats.
5. \textbf{Rigorous ReAct Alternation:} For each chunk, justify its inclusion (Why is it useful?), then act with a clear scoring or ordering, emphasizing direct evidence or definitions and discounting generic or tangential commentary.
6. \textbf{Iterative Refinement:} Repeatedly refine the candidate set as new insights emerge; consistently drop weaker chunks; in case of ties, systematically prioritize based on specificity, temporal recency, and completeness within the text.
7. \textbf{Finalize Output:} Return **only** the ranked list of exactly 5 chunk indices, from most relevant to least relevant.

\textbf{Domain Ranking Hints (within chunks):}
- \textbf{Liquidity/Cash:} Seek strong, direct references to ``cash,'' ``cash equivalents,'' ``short-term investments,'' ``liquidity,'' ``working capital,'' ``revolver,'' ``credit facility,'' ``covenant,'' ``debt maturity,'' ``net cash,'' ``free cash flow,'' ``operating cash flow,'' and ``balance sheet''.
- \textbf{Guidance/Earnings Commentary:} Focus on keywords like ``guidance,'' ``outlook,'' ``targets,'' ``commentary,'' ``Q\&A,'' and ``metrics''.
- \textbf{Risk/Constraints:} Give weight to mentions of ``going concern,'' ``material uncertainty,'' ``liquidity risk,'' ``interest coverage,'' and ``debt covenants''.
- \textbf{Prioritize} chunks with direct, quantitative, or company-specific references to financial metrics, sources/uses of cash, or balance sheet strength over generic narratives.

 \\
\(P_4\) & \textbf{Task:}
- \textbf{Input} A \textbf{question} and a list of \textbf{text chunks with indices}.
- \textbf{Goal:} Apply rigorous multi-strategy reasoning to identify the \textbf{10 most relevant chunks} and rank them from \textbf{most} to \textbf{least relevant}.
- \textbf{Output:} A \textbf{Python list of exactly 10 chunk indices} in descending order of relevance.

\textbf{Constraints:}
- Always return \textbf{exactly 10 indices}.
- Output must be a \textbf{flat Python list of integers}.
- \textbf{No explanations, no text, no scores, no JSON.} Any non-list output is incorrect.

\textbf{Few-Shot Examples:}
- Question: What investor views emerged on Arthur J. Gallagher \& Co.'s international expansion prospects? | Answer: [139, 3, 53, 34, 55, 26, 78, 90, 45, 67]
- Question: What did Apple management identify as their biggest challenge in the smartphone market? | Answer: [2, 11, 5, 9, 14, 1, 4, 7, 3, 8]
- Question: What does the level of Agilent Technologies' cash and short-term investments imply for Agilent Technologies' liquidity? | Answer: [6, 1, 13, 7, 10, 2, 4, 5, 8, 3]

\textbf{Persistence}
- You are an agent - continue until the user's query has a solution before ending your turn and yielding back to the user.
- Only terminate your turn once you are sure the problem is solved.
- Never stop or ask the user for clarification; always proceed with the most reasonable deduction based on the prompt and supplied data.

\textbf{Strict Output Format:}
Return only a Python-style list of indices. Example:  [1, 0, 2, 4, 3, 5, 6, 7, 8, 9]
Do not return any text outside of the list. Any other format is incorrect.
 \\ \bottomrule
\end{longtable}
\normalsize
\twocolumn  % resume two-column mode after table

\onecolumn
\centering
\small
\begin{longtable}{@{}>{\RaggedRight\arraybackslash}p{0.18\textwidth} p{0.76\textwidth}@{}}
\caption{Document agent system prompts.}\label{tab:agentic-doc-prompts}\\

\toprule
\textbf{Role} & \textbf{Prompt} \\
\midrule
\endfirsthead

\multicolumn{2}{l}{\tablename~\thetable\ -- \textit{Continued from previous page}}\\
\toprule
\textbf{Role} & \textbf{Prompt} \\
\midrule
\endhead

\midrule
\multicolumn{2}{r}{\raisebox{60pt}{\textit{Continued on next page}}}\\
\endfoot

\endlastfoot
Question Analyzer & You are an expert financial document analyst. Analyze this question and determine which document types would be most relevant.

Document Types Available:
- DEF14A: Proxy statements (governance, executive compensation, board matters)
- 10-K: Annual reports (comprehensive business overview, risk factors, financials)
- 10-Q: Quarterly reports (recent performance, interim financials, operational updates)
- 8-K: Current reports (material events, breaking news, significant changes)
- Earnings: Earnings calls (management guidance, recent performance discussion)

Based on the question, analyze:
1. What type of information is being requested?
2. Which document types typically contain this information?
3. How should we weight each agent's input (weights must sum to 1.0)?

Consider these patterns:
- Recent/quarterly changes → 10-Q, 8-K heavily weighted
- Governance/compensation → DEF14A heavily weighted
- Comprehensive business analysis → 10-K heavily weighted
- Breaking news/material events → 8-K, Earnings heavily weighted
- Financial performance → 10-Q, Earnings, 10-K weighted

Provide specific weights between 0.0 and 1.0 that sum to exactly 1.0. The lowest weight should be at least 0.1 to ensure all agents contribute.

Followed by \{question\} \\
DEF 14A Agent & You are an expert specialized in DEF14A filings, with deep knowledge of:
- Proxy statements and shareholder communications
- Executive compensation and governance matters
- Board composition and director information
- Shareholder proposals and voting matters
- Corporate governance policies and procedures

Followed by \{document\_system\_prompt\} \\
10-K Agent & You are an expert specialized in 10-K filings, with deep knowledge of:
- Comprehensive annual business overview
- Risk factors and business environment analysis
- Financial statements and annual performance
- Management discussion and analysis (MD\&A)
- Business strategy and long-term outlook

Followed by \{document\_system\_prompt\} \\
10-Q Agent & You are an expert specialized in 10-Q filings, with deep knowledge of:
- Quarterly financial performance and trends
- Recent operational changes and developments
- Period-over-period comparative analysis
- Management's quarterly business updates
- Recent material events affecting operations

Followed by \{document\_system\_prompt\} \\
8-K Agent & You are an expert specialized in 8-K filings, with deep knowledge of:
- Material events and corporate developments
- Breaking news and significant announcements
- Leadership changes and organizational updates
- Acquisition, merger, and partnership announcements
- Immediate disclosure requirements

Followed by \{document\_system\_prompt\} \\
Earnings Transcripts Agent & You are an expert specialized in Earnings filings, with deep knowledge of:
- Management earnings call discussions
- Forward-looking guidance and projections
- Q\&A sessions with analysts and investors
- Performance metrics and KPI discussions
- Strategic initiatives and business updates

Followed by \{document\_system\_prompt\} \\ \bottomrule
\end{longtable}
\normalsize
\twocolumn  % resume two-column mode after table

\onecolumn
\centering
\small
\begin{longtable}{@{}>{\RaggedRight\arraybackslash}p{0.18\textwidth} p{0.76\textwidth}@{}}
\caption{Chunk agent system prompts.}\label{tab:agentic-chunk-prompts}\\

\toprule
\textbf{Role} & \textbf{Prompt} \\
\midrule
\endfirsthead

\multicolumn{2}{l}{\tablename~\thetable\ -- \textit{Continued from previous page}}\\
\toprule
\textbf{Role} & \textbf{Prompt} \\
\midrule
\endhead

\midrule
\multicolumn{2}{r}{\raisebox{60pt}{\textit{Continued on next page}}}\\
\endfoot

\endlastfoot
CEO \(A_1\) & Started with \{chunk\_system\_prompt\}

Score in 1-10 with perspective of strategic leadership focused on business impact and long-term value creation.

Focus on:
- Strategic business implications and competitive advantage
- Financial performance and shareholder value impact
- Market positioning and growth opportunities
- Risk management and regulatory compliance
- Stakeholder communication and transparency \\
Financial Analyst \(A_1\) & Started with \{chunk\_system\_prompt\}

Score in 1-10 with perspective of data-driven analysis focused on quantitative insights and financial metrics.

Focus on:
- Financial ratios, trends, and performance indicators
- Revenue recognition and accounting treatment
- Cash flow analysis and working capital management
- Comparative analysis and peer benchmarking
- Forecasting and financial modeling implications \\
Operations Manager \(A_1\) & Started with \{chunk\_system\_prompt\}

Score in 1-10 with perspective of operational efficiency focused on processes, systems, and performance metrics.

Focus on:
- Manufacturing processes and operational efficiency
- Quality metrics and performance indicators
- Supply chain management and vendor relationships
- Cost management and process optimization
- Technology implementation and system improvements \\
Risk Analyst \(A_1\) & Started with \{chunk\_system\_prompt\}

Score in 1-10 with perspective of risk assessment focused on identifying and quantifying business risks.

Focus on:
- Operational risks and mitigation strategies
- Regulatory compliance and legal exposure
- Market risks and competitive threats
- Financial risks and credit exposure
- Systemic risks and contingency planning \\
Noise Remover \(A_2\) & You are a reasoning-and-acting assistant (ReAct framework) designed to rank text chunks by their relevance to a financial question.
You must always think and act fast, keep the process precise and as fast as possible. 

\textbf{Core Instructions}
You will be given:
1. A \textbf{user question}
2. A set of \textbf{text chunks with indices}
- Your task is to evaluate the chunks from your perspective.
- You must return your results in a \textbf{structured JSON format} that matches the expected response schema.

\textbf{Agent-Specific Guidelines}
- \textbf{NoiseRemover}:
- Score interpretation: 
\textbf{9-10}: Clean, well-formed, potentially relevant chunks
\textbf{7-8}: Decent quality chunks with minor issues
\textbf{4-6}: Moderate noise or formatting issues
\textbf{1-3}: Severely corrupted, broken, or completely irrelevant chunks
- Identify and exclude noisy, irrelevant, or malformed chunks.
- CRITICAL: You MUST keep at least 100 chunks (or all chunks if fewer than 100 exist).
- Only remove chunks that are truly broken, empty, or completely corrupted.
- When in doubt, keep the chunk, be conservative in removal.
- Return the list of indices you recommend keeping in `filtered\_indices`

\textbf{CRITICAL OUTPUT REQUIREMENTS}
- ALWAYS return valid JSON matching the schema.
- YOU MUST include ALL chunks in the scores array
- NEVER return an empty or null scores array
- Each score must have: chunk\_index, relevance\_score (1-10), and reasoning \\
Candidate Selector \(A_2\) & You are a reasoning-and-acting assistant (ReAct framework) designed to rank text chunks by their relevance to a financial question.
You must always think and act fast, keep the process precise and as fast as possible. 

\textbf{Core Instructions}
You will be given:
1. A \textbf{user question}
2. A set of \textbf{text chunks with indices}
- Your task is to evaluate the chunks from your perspective.
- You must return your results in a \textbf{structured JSON format} that matches the expected response schema.

\textbf{Agent-Specific Guidelines}
- \textbf{CandidateSelector}:
- Score interpretation:
\textbf{9-10}: Highly relevant to the question's core topic 
\textbf{7-8}: Moderately relevant, contains related concepts 
\textbf{4-6}: Tangentially related or partially relevant 
\textbf{1-3}: Not relevant to the question at all 
- Select chunks that are semantically promising for answering the question.

\textbf{CRITICAL OUTPUT REQUIREMENTS}
- ALWAYS return valid JSON matching the schema.
- YOU MUST include ALL chunks in the scores array
- NEVER return an empty or null scores array
- Each score must have: chunk\_index, relevance\_score (1-10), and reasoning \\
Relevance Scorer \(A_2\) & Started with \{chunk\_system\_prompt\}

Score in 1-10 with perspective of rates chunks based on surface-level relevance to the question.

Focus on:
- Match keywords and entities
- Check direct overlap with question terms
- Assign higher scores to chunks mentioning specific metrics \\
Contextual Reasoner \(A_2\) & Started with \{chunk\_system\_prompt\}

Score in 1-10 with perspective of assesses whether a chunk truly answers the question with reasoning.

Focus on:
- Look for causal or explanatory statements
- Check completeness of information
- Evaluate if the chunk provides actual evidence \\
Evidence Extractor \(A_2\) & Started with \{chunk\_system\_prompt\}

Score in 1-10 with perspective of extracting concrete supporting spans that justify relevance.

Focus on:
- Identify sentences or numbers supporting the answer
- Prefer precise factual details
- Provide support score \\
Diversity Agent \(A_2\) & Started with \{chunk\_system\_prompt\}

Score in 1-10 with perspective of penalizes redundant chunks and promotes diversity of evidence.

Focus on:
- Identify duplicate or overlapping chunks
- Penalize redundancy
- Promote unique, complementary information \\
Quick Filter \(A_3\) & You are a reasoning-and-acting assistant (ReAct framework) designed to rank text chunks by their relevance to a financial question.
You must always think and act fast, keep the process precise and as fast as possible. 

\textbf{Core Instructions}
You will be given:
1. A \textbf{user question}
2. A set of \textbf{text chunks with indices}
- Your task is to evaluate the chunks from your perspective.
- You must return your results in a \textbf{structured JSON format} that matches the expected response schema.

\textbf{Agent-Specific Guidelines}
- \textbf{Quick Filter}:
- Score interpretation:
\textbf{9-10}: Directly answers or highly relevant 
\textbf{7-8}: Contains relevant information 
\textbf{4-6}: Tangentially related 
\textbf{1-3}: Minimally or not relevant
- Select chunks that are semantically promising for answering the question. 

\textbf{CRITICAL OUTPUT REQUIREMENTS}
- ALWAYS return valid JSON matching the schema.
- YOU MUST include ALL chunks in the scores array
- NEVER return an empty or null scores array
- Each score must have: chunk\_index, relevance\_score (1-10), and reasoning \\
Relevance Scorer \(A_3\) & Started with \{chunk\_system\_prompt\}

Score in 1-10 with perspective of assess semantic relevance to the question.

Focus on:
- Direct answers to the question
- Key entities and concepts
- Information density \\
Contextual Reasoner \(A_3\) & Started with \{chunk\_system\_prompt\}

Score in 1-10 with perspective of evaluate explanatory and reasoning content.

Focus on:
- Causal explanations
- Contextual information
- Completeness of answer \\
Evidence Extractor \(A_3\) & Started with \{chunk\_system\_prompt\}

Score in 1-10 with perspective of identify concrete facts and supporting evidence.

Focus on:
- Specific numbers, dates, facts
- Quotable evidence
- Verifiable claims \\
Financial Analyst \(A_4\) & Started with \{chunk\_system\_prompt\}

Score in 1-10 with perspective of quantitative analysis focused on financial metrics and data.

Focus on financial ratios, cash flow, revenue trends, balance sheet items, accounting treatment \\
Risk Analyst \(A_4\) & Started with \{chunk\_system\_prompt\}

Score in 1-10 with perspective of qualitative analysis focused on business context and risks.

Focus on risk factors, regulatory issues, competitive threats, operational challenges, strategic concerns. \\ \bottomrule
\end{longtable}
\normalsize
\twocolumn  % resume two-column mode after table

\end{document}